%% file: paper.tex
\documentclass[journal,onecolumn]{IEEEtran}

\usepackage{amsmath,amsfonts,amssymb,paralist,latexsym,epsfig,color,rotating,times,float,subcaption,algorithm,verbatim,enumitem,bm,graphicx,natbib,listings,url}
\usepackage[normalem]{ulem}
\usepackage{epstopdf}
\usepackage{tikz,pgfplots}
\usetikzlibrary{arrows,shapes,positioning,snakes}
\usetikzlibrary{decorations.markings,decorations.pathmorphing,decorations.pathreplacing}
\usetikzlibrary{calc,patterns,shapes.geometric,topaths,fit,positioning}

\tikzset{
    block/.style={rectangle, draw, line width=0.5mm, black, text width=4.5em, text centered,
                 minimum height=1em},
               line/.style={draw, -latex}}

      \tikzset{
    block3/.style={rectangle, draw, line width=0.5mm, black, text width=7.5em, text centered,
                 minimum height=1em},
               line/.style={draw, -latex}}

\tikzset{
    block2/.style={rectangle, draw, line width=0.5mm, text centered,
                 minimum height=2em},
               line/.style={draw, -latex}}

\tikzset{
    blockd/.style={rectangle, draw, line width=0.2mm, black, text width=10.8em,  text centered,
                 minimum height=2em},
    line/.style={draw, -latex}}

\tikzset{
    blockb/.style={rectangle, draw, line width=0.2mm, black,  text centered,
                 minimum height=2em},
    line/.style={draw, -latex}}
\tikzset{
    blockff/.style={rectangle, draw, line width=0.2mm, black,  text centered,text width=6em,
                 minimum height=2em},
    line/.style={draw, -latex}}
\tikzset{
    blockfg/.style={rectangle, draw, line width=0.2mm, black,  text centered,text width=9.5em,
                 minimum height=2em},
    line/.style={draw, -latex}}

\tikzset{
    blockss/.style={rectangle, draw, line width=0.3mm, black, text width=2.5em, text centered,
                 minimum height=1em},
               line/.style={draw, -latex}}
             \tikzset{
    blockssd/.style={rectangle, draw, line width=0.3mm, black, text width=7.5em, text centered,
                 minimum height=1em},
               line/.style={draw, -latex}}

\tikzset{
    blocka/.style={rectangle, draw, line width=0.3mm, black, text width=3.3em, text centered,
                 minimum height=1em},
               line/.style={draw, -latex}}

             \tikzset{
   blockf/.style={rectangle, draw, line width=0.3mm, black, text width=13.2em, text centered,
                 minimum height=1em},
               line/.style={draw, -latex}}

\newcommand{\blue}[1] {}
\include{mycommands}

\begin{document}

\title{Langevin Dynamics for Adaptive Inverse Reinforcement Learning of Stochastic Gradient Algorithms}

\author{Vikram Krishnamurthy,  George Yin\thanks{{Vikram Krishnamurthy is with the School of 
 Electrical \& Computer
Engineering,  Cornell University, NY 14853, USA. 
vikramk@cornell.edu.  G. Yin is with the Department of Mathematics, University of Connecticut, \\ Storrs, CT 06269-1009, USA. gyin@uconn.edu}}

\today
}


\maketitle

\begin{abstract}
  Inverse reinforcement learning (IRL) aims to  estimate the reward function of optimizing agents by observing their response (estimates or actions).
  This paper  considers IRL  when noisy estimates of the gradient of
a reward  function  generated by
  multiple stochastic gradient agents are observed.
  We  present a generalized Langevin dynamics algorithm to estimate the reward function $\Reward(\th)$; specifically, the resulting Langevin algorithm asymptotically generates samples from the  distribution
  proportional to $\exp(\Reward(\th))$. The proposed adaptive IRL algorithms use  kernel-based passive learning schemes. We also construct multi-kernel passive Langevin algorithms for IRL which are suitable for high dimensional data. The performance of the proposed IRL algorithms are illustrated on examples in adaptive Bayesian learning, logistic regression (high dimensional problem) and constrained Markov decision processes.  We prove weak convergence of the proposed IRL algorithms using martingale averaging methods.
We also analyze the tracking performance of the IRL algorithms in non-stationary environments where the utility function $\Reward(\th)$ jump changes over time as a slow Markov chain.


\end{abstract}


\section{Introduction} \label{sec:intro}
 Inverse reinforcement learning (IRL) aims to estimate the reward function of optimizing agents by observing their actions (estimates). Classical IRL is off-line: given a data set of actions chosen according to the optimal policy of a Markov decision process,  \cite{NR00} formulated a set of inequalities that  the reward function must satisfy.  In comparison, this paper constructs and analyzes real time  IRL algorithms by observing optimizing agents that are performing real time reinforcement learning (RL).
The problem we consider is this: Suppose we  observe  estimates of multiple (randomly initialized) stochastic gradient algorithms (reinforcement learners) that  aim to maximize a (possibly non-concave) expected reward. {\em How to  design another stochastic gradient algorithm \textit{(inverse learner)} to estimate the expected reward function?}

\subsection{RL and IRL Algorithms}
To discuss  the main ideas, we first describe the point of view of multiple  agents performing reinforcement learning (RL).  These agents act sequentially to perform RL by using   stochastic gradient algorithms  to maximize a reward function.
Let $\tslow = 1,2\ldots$ index agents that perform RL sequentially. The sequential protocol is as follows.  The
agents  aim to maximize a possibly non-concave   reward $\Reward(\th) = \E\{\reward_\dtime(\th)\}$ where $\th \in \reals^\thdim$.
Each agent $\tslow$  runs a stochastic  gradient algorithm over the time horizon $ \dtime \in \{ \stoptime_{\tslow} , \stoptime_\tslow+1,\ldots, \stoptime_{\tslow+1}-1\}$:
\begin{equation}
  \label{eq:rl}
 \begin{split}
   \th_{\dtime+1} &= \th_\dtime + \step \,\nabla_\th \reward_\dtime(\th_\dtime), \quad \dtime = \stoptime_{\tslow} , \stoptime_\tslow+1,\ldots, \stoptime_{\tslow+1}-1 \\
 &  \text{ initialized independently by }   \th_{\stoptime_\tslow} \sim \belief(\cdot).
 \end{split}
\end{equation}
Here $\nabla_\th \reward_\dtime(\th_\dtime)$ denotes the sample path gradient evaluated at $\th_\dtime$, and
$\stoptime_\tslow$, $\tslow=1,2\ldots,$ denote stopping times measurable wrt the $\sigma$-algebra
generated by
$\{\th_{\stoptime_\tslow}, \nabla_\th \reward_\dtime(\th_\dtime),\dtime=\stoptime_\tslow, \stoptime_\tslow+1,\ldots\} $. The initial estimate  $\th_{\stoptime_\tslow} $ for agent $n$ is sampled independently
from  probability density function $\belief$ defined on   $\reals^\thdim$.  Finally,  $\step$ is a small positive constant step size.

Next  we consider the point of view of an observer that  performs \textit{inverse reinforcement learning (IRL)} to estimate the reward function $\Reward(\th)$. The observer (inverse learner)  knows initialization density $\belief(\cdot)$ and only has access to    the estimates $\{\th_\dtime\}$ generated by RL algorithm  (\ref{eq:rl}). The observer  reconstructs
  the gradient  $\nabla_\th \reward_\dtime(\th_\dtime)$  as
$\hat{\nabla}_\th \reward_\dtime(\th_\dtime) = (\th_{\dtime} - \th_{\dtime+1})/\stepa$ for some positive step size $\stepa$.
The main idea of this paper is to propose and analyze  the following IRL  algorithm (which is a passive Langevin dynamics  algorithm) deployed by the observer:
\begin{equation}
  \boxed{  \eth_{\dtime+1} = \eth_\dtime +  \stepa  \bigl[ \kerneln\big(\frac{\th_\dtime-\eth_\dtime}{\kernelstep}\bigr)\,
 \frac{\temperature}{2}\, \hnabla_\th \reward_\dtime(\th_\dtime) + \nabla_\eth \belief(\eth_\dtime)\bigr] \,  \belief(\eth_\dtime)+  \sqrt{\stepa}\, \belief(\eth_\dtime)\, \noise_\dtime,
    \quad \dtime =1,2,\ldots},
\label{eq:irl}
\end{equation}
initialized by  $\eth_0 \in \reals^\thdim$.
Here $\stepa$ and $\kernelstep$ are small positive constant step sizes,  $\{\noise_k, \dtime \geq 0\}$ is an i.i.d. sequence
of  standard $\thdim$-variate Gaussian random variables, and $\temperature = \step/\stepa $ is a fixed constant. 
Note that we have  expressed  (\ref{eq:irl}) in terms of $\nabla_\th \reward_k(\th_k)$
(rather than $\hat{\nabla}_\th \reward_\th(\th_k)$) since we have absorbed the ratio of step sizes into  the scale factor $\temperature$.

The key construct in (\ref{eq:irl})  is the kernel function
   $\kernel(\cdot)$. This kernel function is chosen by the observer such that $\kernel(\cdot)$ decreases monotonically to zero as any component of the argument increases to infinity,
\beq \label{eq:kernel_properties} \kernel(\th) \geq 0, \quad \kernel(\th) = \kernel(-\th), \quad  \int_{\reals^\thdim} \kernel(\th) d\th = 1.
\eeq
An example is to choose the kernel as a multivariate normal  $\normal(0,\sigma^2I_{\thdim})$ density with $\sigma = \kernelstep $, i.e.,
$$ \kerneln\bigl(\frac{\th}{\kernelstep}\bigr) = (2 \pi)^{-\thdim/2}  \kernelstep^{-N} \exp \bigl (- \frac{\|\th\|^2}{2 \kernelstep^2}\bigr),
$$ which is essentially  like
a Dirac delta centered at 0 as $\kernelstep \rightarrow 0$.
Our main result stated informally is as follows; see Theorem~\ref{thm:weak-conv} in
Sec.\ref{sec:weak} for the formal statement.

\

\noindent {\bf Informal Statement of Result.}\label{result:informal}
{\it
Based on the estimates $\{\th_\dtime\}$ generated by RL algorithm  \eqref{eq:rl}, the
 IRL algorithm  \eqref{eq:irl} asymptotically generates samples
$\{\eth_\dtime\}$
from the Gibbs measure
\beq \stat(\eth)   \propto  \exp \bigl(  \temperature \Reward(\eth ) \bigr) , \quad \eth \in \reals^\thdim, \quad
\text{ where } \temperature =  \step/\stepa .\label{eq:stationary1} \eeq
}

To explain the above result, let   $\hat{\stat}$ denote the empirical density function constructed from samples $\{\eth_\dtime\}$ generated by IRL algorithm (\ref{eq:irl}). Then
clearly\footnote{Since the IRL algorithm  does not know the step size $\step$ of the RL, it can only estimate $\Reward(\cdot)$ up to a proportionality constant $\temperature$. In classical Langevin dynamics  $\temperature $ denotes an inverse temperature parameter.}   $\log \hat{\stat}(\eth) \propto \Reward(\eth)$. Thus  IRL algorithm~(\ref{eq:irl}) serves as a \textit{non-parametric method} for  stochastically exploring and reconstructing  reward $\Reward$, given the estimates $\{\th_\dtime\}$ of  RL algorithm (\ref{eq:rl}).
Hence  based on  the estimates $\{\th_\dtime\}$ generated by  RL algorithm (\ref{eq:rl}), IRL algorithm~(\ref{eq:irl})  serves as a randomized sampling method for exploring the reward $\Reward(\eth)$ by  simulating random samples from it. Finally, in adaptive Bayesian learning discussed in  Sec.\ref{sec:numerical},  the RL agents maximize  $\log \Reward(\eth)$ using gradient algorithm (\ref{eq:rl});  then IRL algorithm (\ref{eq:irl}) directly yields samples from $\temperature \Reward(\eth)$.

\subsection{Context and Discussion}

{\color{black}{
The stochastic gradient RL algorithm (\ref{eq:rl}) together with non-parametric
passive Langevin IRL algorithm (\ref{eq:irl}) constitute our main setup. Figure \ref{fig:schematic} displays our framework.

\begin{figure}[h]
\begin{tikzpicture}[node distance =5.25cm and 2cm, auto]
   \node [blockff] (l1) {Stochastic Gradient Learner $\{\th_k\}$};
 \node [blockff,right of=l1] (l2) {Passive IRL Langevin\\ Dynamics~$\{\eth_\dtime\}$};
 \node [blockff,left of=l1] (l0) {Time evolving Utility $\Reward(\cdot)$};
  \node [right of=l2,node distance=2.5cm] (estimatedu)[draw=none]{};
 \draw[->](l0) -- node[above,pos=0.5]{noisy} node[below,pos=0.5]{measurement} (l1);
 \draw[->](l1) -- node[above,pos=0.5]{ $\{\nabla_\th \reward_\dtime(\th_\dtime)\}$} (l2);
 \draw[->](l2) --  node[above,pos=0.5]{ $\hat{\Reward}(\cdot)$} (estimatedu);
 \draw[dashed,->](l2) --  node[anchor=south] {} ++(0,-1.5) -| (l1);
 \node[text width=4cm] at (3,-1.2) {active (Sec.\ref{sec:activeirl})};
\end{tikzpicture}
\caption{\blue{\small Schematic of proposed adaptive IRL framework. Multiple agents (learners)
  compute noisy gradient estimates $ \nabla_\th \reward_\dtime(\th_\dtime)$ of
  a possibly time evolving reward $\Reward(\cdot)$.
  By observing these gradient estimates, the IRL Langevin dynamics algorithm
 (\ref{eq:irl})  generates samples $\eth_k \sim \exp(\temperature \Reward(\eth))$. So reward
  $\Reward(\cdot)$ can be estimated from the log of the empirical distribution of $\{\eth_k\}$.  The IRL algorithm (\ref{eq:irl}) is \textit{passive}:  its estimate $\eth_k$
  plays no role in determining the point $\th_\dtime$ where the learner evaluates gradients $ \nabla_\th \reward_\dtime(\th_\dtime)$.
Sec.\ref{sec:alternate} presents several additional IRL algorithms including a variance reduction algorithm and a non-reversible diffusion.
 Sec.\ref{sec:activeirl} presents an
  active IRL where the IRL requests the learner to provide a gradient at $\eth_k$, but the learner provides the noisy mis-specified gradient estimate
  $\nabla_\th \reward_\dtime(\eth_\dtime+ \obsnoise_\dtime)$ where $\{v_k\}$ is a noise sequence. Finally, Sec.\ref{sec:markov} analyzes the tracking properties of the IRL algorithm when the reward $\Reward(\cdot)$ evolves in time according to an unknown Markov chain.
}}
\label{fig:schematic}
\end{figure} } }

\blue{More abstractly, the IRL problem we address is this: given a sequence of noisy sample path  gradients $\{\nabla_\th \reward_\dtime(\th_k)\}$, how to estimate the expected reward $\Reward(\cdot)$? The IRL algorithm (\ref{eq:irl}) builds on classical stochastic gradient algorithms in 3 steps. First, it is {\em passive}: it does not specify where the RL agents compute gradient estimates.  The gradient estimates are evaluated by RL agents at points $\th_k$, whereas the IRL algorithm requires  gradients at $\eth_k$. To incorporate these mis-specified gradients, the passive algorithm uses the kernel $\kernel(\cdot)$.
  Second, a classical passive stochastic gradient algorithm only estimates a local  maximum of $\Reward(\cdot)$; in comparison we are interested in non-parametric reconstruction  (estimation) of the entire reward $\Reward(\cdot)$. Therefore,  we use  a passive {\em Langevin dynamics} based algorithm. Finally, we are interested in {\em tracking} (estimating) time evolving reward functions $\Reward(\cdot)$. Therefore we use a {\em constant step size}, passive Langevin dynamics IRL algorithm.}

To give additional insight we now discuss the context, useful generalizations of IRL algorithm (\ref{eq:irl}),  and related works in the literature.

\begin{enumerate}[wide,labelwidth=!, labelindent=0pt,label=(\roman*)]
\item {\em Multiple agents}.  The multiple agent RL algorithm  (\ref{eq:rl}) is natural in non-convex stochastic optimization problems. Starting from various randomly chosen  initial conditions  $\th_{\stoptime_\tslow} \sim \belief(\cdot)$, the  agents evaluate the gradients $\nabla_\th\reward_\dtime(\th_\dtime)$
  at various points $\th_\dtime$ to estimate the global maximizer. Since 
  the initializations $\{\th_{\stoptime_\tslow}\}$ is a sequence of
  independent random variables, 
  the RL agents can also act in parallel
(instead of sequentially).
 Given this sequence of gradients $\{\nabla_\th\reward_\dtime(\th_\dtime)\}$, the aim of this paper is to construct  IRL algorithms to estimate $\Reward(\th)$.
\blue{As an example, motivated by
 stochastic control involving information theoretic measures \citep{GRW14} detailed in Sec.\ref{sec:bayesian}, suppose
 multiple RL agents run stochastic gradient algorithms to estimate the minimum
 of a non-convex Kullback Leibler (KL) divergence.
By observing these gradient estimates, the passive IRL algorithm (\ref{eq:irl})  reconstructs  the KL divergence.}

{\color{black}{
\item  {\em Passive IRL}. The IRL algorithm (\ref{eq:irl}) is a Langevin dynamics based gradient algorithm
  with  injected noise $\{\noise_\dtime\}$. It is a {\em passive} learning algorithm since the gradients are not evaluated at $\eth_\dtime$ by the inverse learner; instead the gradients are evaluated at the random points  $\th_\dtime$ chosen by the RL algorithm. This passive framework  is natural in an IRL. The inverse learner passively observes the RL algorithm and aims to estimate its utility.

  \blue{We emphasize  that the passive Langevin  IRL algorithm \eqref{eq:rl}  estimates
    the utility function $\Reward(\th)$; see (\ref{eq:stationary1}). This is unlike classical passive stochastic gradient algorithms that
    estimate a local stationary point of the utility. To the best of our knowledge, such passive Langevin dynamics algorithms have not been proposed or analyzed - yet such algorithms arise naturally in estimating the utility by observing the estimates from a stochastic gradient algorithm.}

The kernel $\kernel(\cdot)$ in (\ref{eq:irl}) effectively weights the usefulness of the gradient $\nabla_\th\reward_\dtime(\th_\dtime)$ compared to the required gradient
$\nabla_\eth\reward_\dtime(\eth_\dtime)$.
If $\th_\dtime$ and $\eth_\dtime$ are far apart, then kernel $\kernel((\th_\dtime-\eth_\dtime)/\kernelstep)$  will be  small. Then only a small proportion of the gradient estimate $\nabla_\th(\reward_\dtime(\th_\dtime)$ is added to the IRL iteration. On the other hand, if $
\eth_\dtime = \th_\dtime$, 
(\ref{eq:irl}) becomes a standard Langevin dynamics type algorithm. We refer to \cite{Rev77,HN87,NPT89,YY96} for the analysis of  passive stochastic  gradient algorithms.  The key difference compared to these works is that
we are dealing with a passive Langevin dynamics algorithm, i.e., there is an extra injected noise  term involving $\noise_\dtime$.

\item  {\em Intuition behind passive Langevin IRL algorithm (\ref{eq:irl})}. To discuss the intuition behind (\ref{eq:irl}), we first
  discuss the classical Langevin dynamics and also a more general reversible diffusion.
    The classical Langevin dynamics algorithm with fixed step size $\mu>0$ and deterministic reward $\Reward(\th)$    is of the form
    \begin{equation}
      \label{eq:classical_lang}
      \th_{\dtime+1} = \th_{\dtime} + \stepa\,  \hnabla \Reward(\th_\dtime)
   + \sqrt{\stepa}\,  \sqrt{\frac{2}{\temperature}}\, \noise_\dtime,
    \quad \dtime =1,2,\ldots
  \end{equation}
 Indeed \eqref{eq:classical_lang}  is the Euler-Maruyama time discretization of the continuous time diffusion
  process
  \begin{equation}
    \label{eq:3}
    d\th(t) = \nabla_\th R(\th) + \sqrt{\frac{2}{\temperature}}\, d\bm(t)
  \end{equation}
    which has stationary measure $\pdf(\th)$ given by  (\ref{eq:stationary1}).
More generally,  assuming $\sigma(\cdot)$ is differentiable, \citet{ST99} studied  reversible diffusions of the form
    \begin{equation}
      \label{eq:tweedie}
     d\th(t) =  \biggl[\frac{\temperature}{2} \sigma(\th)\, \nabla_\th \Reward(\th)\, dt +  \nabla_\th \sigma(\th)\, dt +  d\bm(t) \biggr]\,\sigma(\th),
   \end{equation}
   whose  Euler-Maruyama time discretization  yields
   \begin{equation}
     \label{eq:tweedie_discrete}
     \th_{\dtime+1} = \th_\dtime +  \stepa  \bigl[
 \frac{\temperature}{2}\, \hnabla_\th \Reward_\dtime(\th_\dtime) + \nabla_\th \sigma(\th_\dtime)\bigr] \,  \sigma(\th_\dtime)+  \sqrt{\stepa}\, \sigma(\th_\dtime)\, \noise_\dtime,  \quad \dtime =1,2,\ldots
   \end{equation}
    It is easily verified that  reversible diffusion \eqref{eq:tweedie} has the same   Gibbs stationary measure $\pdf(\th)$ in~(\ref{eq:stationary1}).

    The IRL algorithm \eqref{eq:irl} substantially generalizes
      \eqref{eq:tweedie_discrete} in three ways: First, the gradient is at a mis-specified point $\th_k$ compared to $\eth_k$; hence we use the kernel
      $\kernel$ as discussed in point (ii) above.
      Second, unlike \eqref{eq:tweedie_discrete} which uses  $\nabla_\th \Reward(\th)$, IRL
      algorithm (\ref{eq:irl}) only has the (noisy) gradient estimate $\nabla_\th \reward_\dtime(\th)$. Finally, we choose $\sigma(\th)$ as $\belief(\th)$, namely the initialization density specified in (\ref{eq:rl}), to ensure that the stationary measure is as specified in \eqref{eq:stationary1} as explained at the end of
      Sec.\ref{sec:informal}.

The intuition behind the weak convergence of the passive Langevin IRL algorithm (\ref{eq:irl}) is explained in Sec.\ref{sec:informal}.
It is shown there  via stochastic averaging arguments  as the kernel converges to a Dirac-delta, the IRL algorithm (\ref{eq:irl})  converges to the reversible diffusion process \eqref{eq:tweedie} with
stationary measure given by (\ref{eq:stationary1}).

  \item {\em  IRL for Markov Decision Process.}
  Several types of RL based policy gradient algorithms in the Markov decision process
(MDP) literature \citep{BT96,SB98} fit our framework.
As a motivation, we now briefly discuss IRL for an infinite horizon average cost\footnote{As mentioned in Sec.\ref{sec:mdp}, our   IRL algorithms also apply to the simpler discounted cost MDP case.} MDP; details are  discussed in
Sec.\ref{sec:mdp}. Let  $\{\state_n\}$ denote a finite state Markov chain with
controlled transition probabilities $\tp_{ij}(\action) = \prob[\state_{n+1} = j | \state_n = i, \action_n = \action]$ where action $\action_n$ is chosen from policy $\policy_\th$ parametrized by $\th$ as  $\action_n = \policy_\th(\state_n)$.
Solving an average cost MDP (assuming it is unichain \citep{Put94}) involves computing  the optimal parameter $\th^*= \sup\{\th: \Reward(\th)\}$ where  the cumulative reward is
\begin{equation}
   \label{Jo}
\Reward(\th) = \lim_{\finaltime\to\infty} \inf \frac{1}{\finaltime} \esp_{\th} \Big[\sum_{n=1}^\finaltime
\mreward(\state_n,\action_n) \mid \state_0 = x\Big], \action_n = \policy_\th(\state_n)
\end{equation}
Suppose now that a forward learner runs a policy gradient RL  algorithm  that evaluates
estimates $\nabla_\th r_k(\th)$ of $\nabla_\th \Reward(\th)$ in order to estimate $\th^*$.
Given these gradient estimates, how can an IRL algorithm estimate $\Reward(\th)$?

In Sec.\ref{sec:mdp}, motivated by widely used fairness constraints
  in wireless communications, we consider more general  average cost
  \textit{constrained} MDPs (CMDPs), see \cite{Alt99,NK10,BJ10}. In CMDPs \citep{Alt99}, the optimal policy is randomized.  Since the optimal policy is randomized, classical stochastic dynamic programming or Q-learning cannot be used to solve CMDPs as they yield deterministic policies. One can construct a Lagrangian dynamic programming formulation \citep{Alt99} and Lagrangian Q-learning algorithms \citep{DK07}.
  Sec.\ref{sec:mdp}  considers the case where the RL agents deploy a  policy gradient  algorithm. By observing these gradient estimates, our IRL algorithm reconstructs the Lagrangian of the CMDP.

  Notice that our non-parametric setup is different to classical IRL in \cite{NR00}
where the inverse learner has access to actions from the optimal policy, knows the controlled transition probabilities, and formulates a set of linear inequalities that the reward function
$\mreward(\state,\action)$ satisfies.
Our IRL framework only has access to gradient estimates $\nabla_\th \reward_\dtime(\th)$ evaluated at random points $\th$, and does not require knowledge of the parameters of the CMDP.
Also  our IRL framework is adaptive (see point (vii) below): the IRL algorithm
(\ref{eq:irl}) can track a time evolving $\Reward(\th)$ due to  the transition probabilities or rewards  of the MDP evolving
over time (and unknown to the inverse learner).}}


\item  {\em Multi-kernel IRL}.  IRL algorithm (\ref{eq:irl}) requires the gradient $\nabla_\th \reward(\th_\dtime)$
and knowing the density $\belief(\cdot)$. In Sec.\ref{sec:highdim} we will discuss a two-time scale multi-kernel IRL algorithm, namely (\ref{eq:mcmcirl}),  that does not require knowledge of the density $\belief(\cdot)$. All that is required
is a sequence of samples $\{\nabla_\th \reward_\dtime(\th_i),i=1\,\ldots,\numparticles\}$ when the IRL estimate is $\eth_\dtime$.  The multi-kernel  IRL algorithm (\ref{eq:mcmcirl}) incorporates variance reduction and is suitable for high dimensional inference. In Sec.\ref{sec:alternate}, we also discuss several other variations of
IRL algorithm (\ref{eq:irl}) including a mis-specified active IRL algorithm where the gradient is evaluated at a point $\th_\dtime$ that is a corrupted value of $\eth_\dtime$.

\item  {\em Global Optimization vs IRL}. Langevin dynamics based gradient algorithms have been studied  as a means for achieving global minimization for non-convex stochastic
optimization problems, see for example \cite{GM91}. The papers \cite{TTV16,RRT17} give a comprehensive study of convergence of the Langevin dynamics stochastic gradient algorithm in a non-asymptotic setting.
Also \cite{WT11} studies Bayesian learning, namely, sampling from the posterior using stochastic gradient Langevin dynamics.

Langevin dynamics for global optimization  considers the limit as $\temperature \rightarrow \infty$.
In comparison, the IRL algorithms in this  paper consider the case of fixed $\temperature = \step/\stepa$, since we are interested in sampling from the reward $\Reward(\cdot)$.   Also, we consider  passive Langevin dynamics algorithms in the context of IRL. Thus the IRL algorithm (\ref{eq:irl}) is non-standard in two ways. First, as mentioned above, it has a kernel to facilitate passive  learning. Second, the IRL algorithm (\ref{eq:irl}) incorporates
the initialization probability $\belief(\cdot)$ which appears in the RL algorithm (\ref{eq:rl}). Thus
(\ref{eq:irl}) is  a non-standard generalized Langevin dynamics algorithm (which still has  reversible diffusion dynamics).

\item  {\color{black}{\em Constant step size Adaptive IRL for Time Evolving Utility}.  An important feature of the IRL algorithm (\ref{eq:irl}) is the  constant step size $\stepa$  (as opposed to a decreasing step size).  This facilities  estimating (adaptively tracking) rewards  that evolve over time. Sec.\ref{sec:markov}  gives a formal weak convergence analysis of the asymptotic tracking capability of the IRL algorithm (\ref{eq:irl}) when the reward $\Reward(\cdot)$  jump changes over time according to an  unknown Markov chain. The analysis is very different to classical
  tracking analysis of stochastic  gradient algorithms \citep{BMP90} where the underlying parameter  (called hyper-parameter) evolves continuously over time.

  In Sec.\ref{sec:markov}, three cases of adaptive IRL are analyzed: (i)  the reward jump changes on a slower time scale than the dynamics of the Langevin IRL  algorithm (ii) the reward jump changes on the same time scale as the Langevin IRL algorithm
  (iii) The reward jump changes on a faster time scale compared to the Langevin IRL algorithm.
  The most interesting (and difficult) case considered in Sec.\ref{sec:markov}  is when the reward changes at the same rate as the IRL algorithm. Then  stochastic averaging theory yields a Markov switched diffusion limit as the asymptotic behavior of the IRL algorithm.
This is in stark contrast to classical averaging theory of stochastic gradient algorithms which yields a deterministic ordinary differential equation \citep{KY03,BMP90}.
  Due to the constant step size, the appropriate notion of convergence is weak convergence \citep{KY03,EK86,Bil99}.  The Markovian hyper-parameter tracking analysis generalizes our earlier work \cite{YKI04,YIK09} in stochastic gradient algorithms to the current case of  passive Langevin dynamics with a kernel.}

\item  {\em Estimating utility functions}. Estimating a utility function given the response of agents is studied under the area of revealed preferences in microeconomics.
 Afriat's theorem \citep{Afr67,Die12,Var12} in
revealed preferences uses the response of a linearly constrained optimizing agent to construct a set of linear inequalities that are necessary and sufficient for an agent to be an utility maximizer; and gives a set valued estimate
of the class of utility functions that rationalize the agents behavior.
Different to revealed preferences, the current paper uses noisy gradients to recover the utility function and that too in real time via a constant step size Langevin diffusion algorithm.

\blue{We already mentioned classical IRL \citep{NR00,AN04} which aims to estimate an unknown deterministic reward function  of an agent by observing the optimal actions of the agent in a Markov decision process setting. 
  \cite{ZMB08} uses the principle of maximum entropy for achieving IRL of optimal agents.
More abstractly, IRL falls under the area of {\em imitation learning}  \citep{HE16,OPN18} which is the process of learning from demonstration.
Our IRL approach can be considered as imitation learning from mis-specified noisy gradients
evaluated at random points in Euclidean space.
  We perform {\em adaptive} (real time) IRL:  given  samples from a stochastic gradient algorithm (possibly a forward RL algorithm), we propose a Langevin dynamics algorithm to estimate the utility function. Our  real-time IRL algorithm  facilitates  adaptive  IRL, i.e.,   estimating (tracking) time evolving  utility functions. In Sec.\ref{sec:markov}, we analyze the tracking properties of such  non-stationary IRL algorithms  when the utility function jump changes according to an unknown Markov process.}

\item  {\em Interpretation as a numerical integration algorithm}. Finally, it is helpful to view IRL algorithm (\ref{eq:irl}) as a numerical integration algorithm when the integrand (gradients to be integrated) are presented at  random points and the integrand terms are corrupted by noise (noisy gradients).
One possible offline approach  is to discretize $\reals^\thdim$  and numerically build up an estimate of the integral at the discretized points by rounding off the evaluated integrands terms to the nearest discretized point. However, such an approach suffers from the curse of dimensionality: one needs
$O(2^\thdim)$ points to construct the integral with a specified level of tolerance.
In comparison, the passive IRL algorithm (\ref{eq:irl}) provides a principled real time approach for generating samples from the integral, as depicted by main result~(\ref{eq:stationary1}).

\item \blue{Although our main motivation for {\em passive} Langevin dynamics stems from IRL, namely estimating a utility function, we mention  the interesting recent paper by \citet{KHH20}  which shows that classical Langevin dynamics   yields more robust RL algorithms compared to classic stochastic gradient.\footnote{We thank an
    anonymous reviewer for bringing this paper to our attention.}  In analogy to \cite{KHH20}, in future work it is worthwhile exploring if our passive Langevin dynamics algorithm can be viewed as a robust version of classical passive stochastic gradient algorithms.}

\end{enumerate}

\subsection{Organization}
The rest of the paper is organized as follows:
\begin{compactenum} \item
  Sec.\ref{sec:context} discusses the IRL algorithm (\ref{eq:irl}), related works in the literature
  and gives an informal  proof of convergence based on averaging theory arguments.  Also the following IRL algorithms are discussed: \begin{compactenum} \item
    A two time scale multi-kernel IRL algorithm  with variance reduction. This IRL algorithm is illustrated in a high dimensional example.

  \item An active IRL algorithm with mis-specified gradient. That is, given the current estimate $\eth_\dtime$, the  IRL is given  a gradient estimate at
    $\nabla_\th \reward_\dtime(\eth_\dtime+ \obsnoise_\dtime)$ where $\obsnoise_\dtime$ is a noise process, and the mis-specified point $\eth_\dtime+ \obsnoise_\dtime$ is known to the IRL algorithm.
  \item A non-reversible diffusion IRL  where a skew symmetric matrix yields a larger spectral gap and therefore faster convergence to the stationary distribution
    (at the expense of increased computational cost).
  \end{compactenum}

\item Sec.\ref{sec:numerical} gives three classes of numerical examples that illustrate our proposed  IRL algorithms:
  \begin{compactenum} \item Learning the KL divergence given noisy gradients. Also IRL for   Adaptive Bayesian learning is discussed.
  \item IRL on  a logistic regression classifier involving the adult {\tt a9a} dataset; this is a large dimensional example with $\thdim = 124$ and requires careful use  of the proposed multi-kernel IRL algorithm.
  \item  IRL for  reconstructing the cumulative reward of an finite horizon constrained Markov decision process (CMDP). Such  CMDPs are non-convex in the action probabilities and have optimal polices that are randomized. We demonstrate how the Langevin-based  IRL can learn a from a policy gradient RL algorithm.
  \end{compactenum}

\item Sec.\ref{sec:weak}  gives a complete weak convergence proof of IRL algorithm (\ref{eq:irl}) using martingale averaging methods.
  Sec.\ref{sec:proofmcmcirl} gives a formal proof of convergence of the multi-kernel algorithm~(\ref{eq:mcmcirl}).
  \item  Sec.\ref{sec:markov}  provides a formal weak convergence analysis of the asymptotic tracking capability of the IRL algorithm (\ref{eq:irl}) when the utility function jump changes according to a slow (but unknown) Markov chain.
  \item Finally, the appendix gives  Matlab source codes for the three numerical examples presented in the paper. So the  numerical results of this paper  are fully reproducible.
\end{compactenum}

\section{Informal Proof and Alternative IRL Algorithms} \label{sec:context}

The RL algorithm (\ref{eq:rl}) together with IRL algorithm (\ref{eq:irl}) constitute our main setup.
In this section, we first start with  an informal  proof of convergence of (\ref{eq:irl}) based on stochastic averaging theory; the formal proof is in Sec.\ref{sec:weak}. The informal proof provided below is useful  because it
gives additional insight into the design of related  IRL algorithms. We then discuss several  related IRL algorithms including a novel multi-kernel version with variance reduction.

\subsection{Informal  Proof of Main Result
\eqref{eq:stationary1}}    \label{sec:informal}
Since
the IRL algorithm (\ref{eq:irl}) uses a constant step size, the  appropriate notion of  convergence  is  weak convergence.
Weak convergence  (for example, \citep{EK86}) is a function space generalization of convergence in distribution; function space because we  prove convergence of the entire trajectory (stochastic process) rather than simply the estimate at a fixed time (random variable).

\blue{A few words about our proof approach. Until the mid 1970s, convergence  proofs of stochastic gradient algorithms assumed martingale difference type of uncorrelated noises. The so-called ordinary differential equation (ODE) approach  was proposed by \cite{Lju77} for correlated
noises and decreasing step size, yielding with probability one convergence. This was subsequently generalized by Kushner and coworkers (see for example \cite{Kus84}) to weak convergence analysis of constant step size algorithms. The assumptions required in this paper  are weaker and
  the results  more general than that used in classical mean square error analysis
because  we are
dealing with
suitably scaled sequences of the iterates that are treated
as stochastic processes rather than random variables. Our
approach captures the dynamic evolution of the
algorithm. As a consequence, using weak convergence
methods we can also analyze the tracking properties of the
IRL  algorithms when the parameters are time varying (see Sec.\ref{sec:markov}).}

As is typically done in weak convergence analysis,
we first represent the sequence of estimates $\{\eth_k\}$  generated by the IRL algorithm as a continuous-time random process. This is done by constructing the continuous-time trajectory  via piecewise
constant interpolation as follows:
For
 $t   \in [0,\horizon]$,
define the continuous-time piecewise constant interpolated processes parametrized by the step size
$\stepa$ as
\beq
\eth^\stepa(t) = \eth_\dtime , \; \text{ for } \ t\in [\stepa \dtime, \stepa \dtime+ \stepa). \label{eq:interpolatedp} \eeq

Sec.\ref{sec:weak} gives the detailed weak convergence  proof using the martingale problem formulation of Strook and Varadhan \citep{EK86}.

Our informal proof of the main result  (\ref{eq:stationary1})  proceeds in two steps:

\underline{\em Step I}.
We  first {\it fix} the kernel step size  $\kernelstep$ and apply stochastic averaging theory arguments: this says that at the slow time scale, we can replace the fast variables by their expected value.
For small step sizes $\step$ and $\stepa=\step/\temperature$, there are three time scales in IRL algorithm (\ref{eq:irl}):
\begin{compactenum} \item
$\{\th_\dtime\}$ evolves slowly on intervals $ \dtime \in \{\stoptime_\tslow, \stoptime_{\tslow+1}-1\}$, and $\{\eth_\dtime\}$ evolves slowly versus $\dtime$.
\item
 We assume that the run-time of the RL algorithm (\ref{eq:rl}) for each agent $\tslow$  is bounded by some finite constant, i.e.,
 $\stoptime_{\tslow+1} - \stoptime_\tslow < M $ for some constant $M$.
 So $\{\th_{\stoptime_{\tslow}}\} \sim \belief$
 is a fast variable compared to
 $\{\eth_\dtime\}$.
\item Finally  the noisy gradient process $\{\nabla_\th \reward_\dtime(\cdot)\}$ evolves at each time $\dtime$ and is  a faster  variable than  $\{\th_{\stoptime_{\tslow}}\}$ which is updated at stopping times $\stoptime_\tslow$.
\end{compactenum}
 With the above time scale separation,
there are two levels of averaging involved. First  averaging the noisy gradient $\nabla_\th \reward_\dtime(\th_\dtime)$ yields $\nabla_\th\Reward(\th)$. Next averaging $\{\th_{\stoptime_{\tslow}}\} $ yields $\th \sim \belief$. Thus applying
 averaging theory to IRL algorithm (\ref{eq:irl}) yields the following averaged system:
 \begin{multline}  \bar{\eth}_{\dtime+1} = \bar{\eth}_\dtime +  \stepa \, \E_{\th \sim \belief} \Big[ \kerneln\big(\frac{\th-\bar{\eth}_\dtime}{\kernelstep}\bigr)\,
 \frac{\temperature}{2}\, \hnabla_\th \Reward(\th) + \nabla_\eth \belief(\bar{\eth}_\dtime)\Big] \,  \belief(\bar{\eth}_\dtime)+  \sqrt{\stepa}\, \belief(\bar{\eth}_\dtime)\, \noise_\dtime \\
  = \bar{\eth}_\dtime +  \stepa \,\int_{\reals^\thdim} \, \kerneln\big(\frac{\th-\bar{\eth}_\dtime}{\kernelstep}\bigr)\,
 \frac{\temperature}{2}\, \belief(\bar{\eth}_\dtime) \hnabla_\th \Reward(\th) \belief(\th) d\th + \belief(\bar{\eth}_\dtime)\, \nabla_\eth \belief(\bar{\eth}_\dtime)   +  \sqrt{\stepa}\, \belief(\bar{\eth}_\dtime)\, \noise_\dtime . \label{eq:discavg}
 \end{multline}
Given the sequence $\{\bar{\eth}_\dtime\}$, define the interpolated continuous time process  $ \bar{\eth}^\stepa$  as in (\ref{eq:interpolatedp}).  Then
 as $\stepa$ goes to zero,
$ \bar{\eth}^\stepa$ converges weakly to the  solution of the stochastic differential equation
\begin{equation} \begin{split}
      d\eth(t) &=   \int_{\reals^\thdim}\kerneln\bigl(\frac{\th - \eth}{\kernelstep}\bigr)\, \biggl[\frac{\temperature}{2} \belief(\eth)\, \nabla_\th \Reward(\th)\, dt \biggr] \,\belief(\th)  \,d\th + \belief(\eth)\, \nabla_\eth \belief(\eth)\, dt +  \belief(\eth)\,  d\bm(t) , \\
      \eth(0) & = \eth_0,
        \end{split} \label{eq:2levelaverage}
      \end{equation}
      where $\bm(t)$ is standard Brownian motion. Put differently,
the Euler-Maruyama time discretization of (\ref{eq:2levelaverage}) yields (\ref{eq:discavg}).
To summarize (\ref{eq:2levelaverage}) is the continuous-time averaged dynamics of  IRL algorithm~(\ref{eq:irl}). This is formalized
in Sec.\ref{sec:weak}.


\underline{\em Step II}. Next, we set the kernel step size $\kernelstep \rightarrow 0$.
  Then $\kernel(\cdot)$ mimics a Dirac delta function and so  the asymptotic dynamics of (\ref{eq:2levelaverage}) become the  diffusion
     \beq
     d\eth(t) =  \biggl[\frac{\temperature}{2} \belief(\eth)\, \nabla_\eth \Reward(\eth)\, dt +  \nabla_\eth \belief(\eth)\, dt +  d\bm(t) \biggr]\,\belief(\eth),  \quad \eth(0) = \eth_0 \label{eq:gld}
     \eeq
     %
     Finally,  (\ref{eq:gld}) is a reversible diffusion and its  stationary measure  is the Gibbs measure $\stat(\eth)$ defined in~(\ref{eq:stationary1}).  Showing this is  straightforward:\footnote{Note  \citet[Eq.34]{ST99}  has a typographic  error in specifying the determinant.}
Recall \citep{KS91} that for a generic diffusion process denoted as $d\state(t) = \statem(\state) dt + \snoisecov(\state) d\bm(t)$, the stationary distribution $\stat$ satisfies
\beq  \forward \stat = \frac{1}{2} \Tr [ \nabla^2 ( \kalmancov \stat)] - \div ( \statem \stat) = 0 , \qquad  \text{ where }
\kalmancov = \snoisecov \snoisecov^\p \label{eq:forward} \eeq
and  $\forward$ is the  forward operator.
From (\ref{eq:gld}),
$\statem(\eth) = [\frac{\temperature}{2} \belief(\eth)\, \nabla_\eth \Reward(\eth)  +  \nabla_\eth \belief(\eth)] \belief(\eth)$, $\snoisecov = \belief(\eth) I$. Then it is  verified by elementary calculus  that $\stat(\eth) \propto  \exp \bigl(  \temperature \Reward(\eth ) \bigr) $ satisfies (\ref{eq:forward}).

To summarize, we have shown informally that IRL algorithm (\ref{eq:irl})  generates  samples from (\ref{eq:stationary1}).
Sec.\ref{sec:weak} gives the formal weak convergence  proof.

(v)
{\em  Why not use classical Langevin dynamics?}
The passive version of the  classical Langevin dynamics  algorithm reads:
\begin{equation}
   \eth_{\dtime+1} = \eth_\dtime + \stepa\, \kerneln\big(\frac{\th_\dtime-\eth_\dtime}{\kernelstep}\bigr)\,  \hnabla \reward_\dtime(\th_\dtime)
   + \sqrt{\stepa}\,  \sqrt{\frac{2}{\temperature}}\, \noise_\dtime,
    \quad \dtime =1,2,\ldots
\label{eq:irlstandard}
\end{equation}
 where $\th_\dtime$ are computed by RL (\ref{eq:rl}).
Then averaging theory (as $\stepa\rightarrow 0$ and then $\kernelstep \rightarrow 0$)
 yields the following asymptotic dynamics  (where $\bm(t)$ denotes standard Brownian motion)
 \beq
 d\eth(t) =  \nabla_\eth \Reward(\eth)\, \belief(\eth) dt + \sqrt{\frac{2}{\temperature}} d\bm(t)   ,  \quad \eth(0) = \eth_0 \label{eq:hard}
 \eeq
 Then  the stationary distribution of (\ref{eq:hard}) is proportional to
 $\exp(\temperature \int \nabla_\eth \Reward(\eth) \belief(\eth) d\eth )$. Unfortunately, this is difficult to relate to $\Reward(\eth)$ and therefore less useful.
  In comparison, the generalized Langevin algorithm (\ref{eq:irl})   yields samples from stationary distribution proportional to $\exp(\temperature\Reward(\eth))$
  from which $\Reward(\eth)$ is easily estimated (as discussed below (\ref{eq:stationary1})).
 This is the reason why we  will use the passive generalized Langevin dynamics (\ref{eq:irl}) for IRL instead of the passive classical  Langevin dynamics (\ref{eq:irlstandard}).

 \subsection{Alternative IRL Algorithms} \label{sec:alternate}
 IRL algorithm (\ref{eq:irl}) is the vanilla  IRL algorithm considered in this paper and its formal
 proof of convergence is given in Sec.\ref{sec:weak}.
 In this section we discuss  several variations of  IRL algorithm  (\ref{eq:irl}). The algorithms discussed below include a passive version of the classical Langevin dynamics, a two-time scale multi-kernel MCMC based IRL algorithm (for variance reduction) and finally,  a non-reversible diffusion algorithm. The construction of these algorithms are based on the informal proof discussed above.

 \subsubsection{Passive Langevin Dynamics Algorithms for IRL}
 IRL algorithm (\ref{eq:irl}) can be viewed as a passive modification of the generalized Langevin dynamics proposed in \cite{ST99}. Since  generalized Langevin dynamics includes  classical Langevin dynamics as a special case, it stands to reason that we can construct a passive version of the classical Langevin dynamics algorithm.  Indeed, instead of (\ref{eq:irl}),  the following   passive Langevin dynamics can be used for IRL
 (initialized by $\eth_0 \in \reals^\thdim$):
     \begin{equation}
 \boxed{   \eth_{\dtime+1} = \eth_\dtime +  \stepa  \, \kerneln\big(\frac{\th_\dtime-\eth_\dtime}{\kernelstep}\bigr)\,
 \frac{\temperature}{2\, \belief(\eth_\dtime)}\, \hnabla_\th \reward_\dtime(\th_\dtime) +  \sqrt{\stepa}\,  \noise_\dtime,
    \quad \dtime =1,2,\ldots }
\label{eq:irl2}
\end{equation}
Note that this algorithm is different to (\ref{eq:irlstandard}) due to the term $\belief(\eth_k)$ in the denominator,
 which makes a  crucial difference. Indeed, unlike  (\ref{eq:irlstandard}), algorithm  (\ref{eq:irl2}) generates samples from (\ref{eq:stationary1}), as we now explain:
By stochastic averaging theory arguments as $\stepa$ goes to zero,
the interpolated processes $  \eth^\stepa$ converges weakly to  (where $\bm(t)$ below is standard Brownian motion)
\begin{equation} \begin{split}
    d\eth(t) &=   \int_{\reals^\thdim}\kerneln\bigl(\frac{\th - \eth}{\kernelstep}\bigr)\, \biggl[\frac{\temperature}{2\,\belief(\eth)} \, \nabla_\th \Reward(\th)\, dt \biggr] \,\belief(\th)  \,d\th +   d\bm(t) , \quad \eth(0) = \eth_0
        \end{split}
      \end{equation}
          Again   as $\kernelstep \rightarrow 0$, $\kernel(\cdot)$ mimics a Dirac delta function and so  the
     $\belief(\cdot)$ in the numerator and denominator cancel out. Therefore
the      asymptotic dynamics become the  reversible diffusion
     \beq
     d\eth(t) = \frac{\temperature}{2}\, \nabla_\eth \Reward(\eth)\, dt  +  d\bm(t),  \quad \eth(0) = \eth_0 \label{eq:ld2}
     \eeq
     Note that (\ref{eq:ld2}) is the classical Langevin diffusion and has stationary distribution
     $\stat$ specified by (\ref{eq:stationary1}). So algorithm  (\ref{eq:irl2}) asymptotically  generates samples from (\ref{eq:stationary1}).

Finally, we note that Algorithm  (\ref{eq:irl2}) can be viewed  as a special case of IRL algorithm (\ref{eq:irl}) since its  limit dynamics (\ref{eq:ld2})
     is a special case of the limit dynamics (\ref{eq:gld}) with $\belief(\cdot) = 1$.

     \subsubsection{Variance Reduction for High Dimensional IRL} \label{sec:highdim}
     For large dimensional problems (e.g., $\thdim=124$ in the numerical example of Sec.\ref{sec:numerical}), the passive IRL algorithm (\ref{eq:irl}) can take a very large number of iterations to converge to its stationary distribution.
This is because with high probability, the kernel $\kernel(\th_k,\eth_k)$ will be close to zero and so updates of $\eth_k$  will occur very rarely.

There is strong motivation to introduce variance reduction in the algorithm. Below we propose a
two time step, multi-kernel variance reduction IRL algorithm motivated by importance sampling.
Apart from the ability to deal with high dimensional problems, the algorithm also does not require
 knowledge of the initialization probability density $\belief(\cdot)$.

Suppose the IRL operates at a slower time scale than the RL algorithm.
At each time $k$ (on the slow time scale), by observing the RL  algorithm, the IRL obtains a  pool of samples
of the
gradients $  \nabla_\th \reward_\dtime(\th_{\dtime,i}) $ evaluated  at a large number of points $\th_{\dtime,i}$,
$i=1,2,\ldots,\numparticles$ (here $i$ denotes the fast time scale).
As previously, each sample $\th_{\dtime,i}$ is chosen randomly from $\belief(\cdot)$.
     Given these sampled derivatives, we propose the following multi-kernel IRL algorithm:
\beq
\boxed{\begin{split}
  \eth_{\dtime+1} &= \eth_\dtime + \stepa \,\frac{\temperature}{2} \, \frac{\sum_{i=1}^\numparticles   \pdf(\eth_\dtime|\th_{\dtime,i}) \hnabla_\th \reward_\dtime(\th_{\dtime,i}) }{\sum_{l=1}^\numparticles \pdf(\eth_\dtime|\th_{\dtime,l})}
  + \sqrt{\stepa}  \noise_\dtime , \quad \th_{\dtime,i} \sim \belief(\cdot)
     \end{split}} \label{eq:mcmcirl}
     \eeq
      In (\ref{eq:mcmcirl}),  we choose the conditional probability density function $\pdf(\th|\eth)$ as follows:
  \beql{eq:cond-al}\pdf(\eth|\th) = \pdf_\obsnoise(\th - \eth)\quad \text{ where } \pdf_\obsnoise(\cdot) = \normal(0,\sigma^2I_\thdim).\eeq
    For notational convenience, for each $\eth$, denote the normalized weights in (\ref{eq:mcmcirl})  as
\beq  \weight_{\dtime,i}(\eth)  =   \frac{\pdf(\eth|\th_{\dtime,i}) }{\sum_{l=1}^\numparticles 
\pdf (\eth| \theta_{k,l})}
\quad i = 1,\ldots, \numparticles  \eeq
     Then these
     $\numparticles $ normalized weights  qualify  as
   symmetric kernels in the sense of (\ref{eq:kernel_properties}). Thus IRL algorithm (\ref{eq:mcmcirl}) can be viewed as  a multi-kernel  passive stochastic approximation
   algorithm. Note that the algorithm does not require knowledge of $\belief(\cdot)$.

Since for each $\dtime$,  the samples $\{\th_{\dtime,i}, i=1,\ldots,\numparticles\}$ are generated i.i.d. random variables,
it is well known from self-normalized importance sampling \cite{CMR05} that as  $\numparticles\rightarrow \infty$, then for fixed $\eth$,
\beq   \sum_{i=1}^\numparticles \weight_{\dtime,i}(\al) \,
\hnabla_\th \reward_\dtime(\th_{\dtime,i}) \rightarrow \E\{\nabla_\th \reward_\dtime(\th) | \eth \}
\quad \text{ w.p.1,}
   \label{eq:sir}
   \eeq
   provided $\E| \pdf(\th|\eth) \,\nabla \reward_\th(\th)| < \infty$.
Similar results can also be established more generally if
$\{\th_{\dtime,i}, i=1,\ldots,\numparticles\}$ is a geometrically  ergodic Markov process with stationary distribution $\belief(\cdot)$.

{\em    Remark}: Clearly  the conditional expectation $ \E\{\nabla_\th \reward_\dtime(\th) | \eth_\dtime \}$ always has smaller variance than
$ \nabla_\th \reward_\dtime(\th)$; therefore variance reduction is achieved in IRL algorithm
(\ref{eq:mcmcirl}).
In sequential Markov chain Monte Carlo  (particle filters), to avoid degeneracy, one resamples from the pool of ``particles''  $\{\th_i,i=1\ldots,\numparticles\}$ according to the probabilities (normalized weights) $\weight_i$.  For large $\numparticles$, the resulting resampled particles have a density $\pdf(\th|\eth_k)$.
However,   we are only interested in computing an estimate of the gradient (and not in propagating particles  over time). So we use the estimate  $\sum_i \gamma_{\dtime,i}  \nabla_\th \reward_\dtime(\th_{\dtime,i})$ in (\ref{eq:mcmcirl}); this always has a smaller variance than resampling and then estimating the gradient; see \citet[Sec.12.6]{Ros13} for an elementary proof.

Why not use the popular MCMC tool of  sequential importance sampling with resampling?   Such a process resamples from the pool of particles and  pastes together components of $\th_i$ from other more viable candidates $\th_j$. As a result, $\numparticles$ composite vectors are obtained, which are more viable. However, since our IRL framework  is passive, this is of no  use since we cannot obtain  the gradient for these $\numparticles$ composite vectors. Recall that in our passive framework, the IRL
has no control over where the gradients $\nabla_\th \reward_k(\th)$ are evaluated.

{\em Informal Analysis of IRL algorithm (\ref{eq:mcmcirl})}.
By stochastic averaging theory arguments as $\stepa$ goes to zero, the interpolated process
  $\eth^\stepa$ from IRL algorithm (\ref{eq:mcmcirl}) converges weakly to
  \beq \label{eq:avgmcmc}
  d \eth(t) = \int_{\reals^\thdim} \frac{\temperature}{2}\,\nabla_\th \Reward(\th) \, \pdf\big(\th| \eth(t)\big) \, d\th \, dt + d\bm(t)  , \qquad \eth(0) = \eth_0
  \eeq
  where $\bm(t)$ is standard Brownian motion.
  Notice that even though $\th_i$ are sampled from the density $\belief(\cdot)$, the above averaging is w.r.t. the conditional density $\pdf(\th|\alpha)$ because  of
  (\ref{eq:sir}).
  For small variance $\sigma^2$, by virtue of the classical  Bernstein von-Mises theorem \citep{Vaa00},
    the conditional density  $\pdf(\th| \eth)$ in
  (\ref{eq:avgmcmc}) acts as a Dirac delta yielding the classical Langevin diffusion
  \beq \label{eq:classicaldiffusion}
  d \eth(t) =\frac{\temperature}{2}\, \nabla_\eth \Reward(\eth(t)) \, dt + d\bm(t)  \eeq
  Therefore  algorithm  (\ref{eq:mcmcirl})  generates samples from distribution (\ref{eq:stationary1}). The formal proof is in Sec.\ref{sec:proofmcmcirl}.

  \subsubsection{Active IRL with Mis-specified Gradient} \label{sec:activeirl}

Thus far we have considered the case where  the RL algorithm provides estimates $\nabla_\th \reward_\dtime(\th_\dtime)$  at randomly chosen points
 independent of the IRL estimate $\eth_\dtime$. In other words, the IRL is passive and has no role in determining where the RL algorithm evaluates gradients.

We now consider
 a modification where  the RL algorithm gives a noisy version of the gradient evaluated
 at a stochastically perturbed value of $\eth_\dtime$.  That is, when the IRL   estimate is $\eth_\dtime$, it  requests the RL algorithm to provide a  gradient estimate
 $\nabla_\th \reward_\dtime(\eth_\dtime)$. But the RL algorithm evaluates the gradient at a mis specified point
 $\th_\dtime = \eth_\dtime + \obsnoise_\dtime$, namely,   $\nabla_\th \reward_\dtime(\eth_\dtime+ \obsnoise_\dtime)$. Here $\obsnoise_\dtime \sim \normal(0,\sigma^2 I_{\thdim})$ is an i.i.d. sequence.  The RL algorithm then provides the IRL algorithm with $\th_\dtime$ and $\nabla_\th \reward_\dtime(\th_\dtime)$. So,  instead of $\th_\dtime$ being independent of $\eth_\dtime$, now  $\th_\dtime$ is conditionally dependent on $\eth_\dtime$ as
 \beq \pdf(\th_\dtime|\eth_\dtime) = \frac{1}{(2 \pi)^\thdim\,\sigma^{\thdim}} \exp( - \frac{1}{2 \sigma^2} \|\th_k - \eth_k\|^2), \qquad \th_k,\eth_k \in \reals^\thdim
 \label{eq:conditional}
 \eeq
 In other words, the IRL now actively specifies where to evaluate the gradient; however, the RL algorithm evaluates a noisy gradient  and that too at  a stochastically perturbed (mis-specified) point~$\th_k$.

 The active  IRL algorithm we propose is  as follows:
 \beq \label{eq:activeirl}
 \begin{split}
   \eth_{k+1} &= \eth_k + \stepa\, \frac{1}{\kernelstep^\thdim}\, \kernel(\frac{\th_k-\eth_k}{\kernelstep})
   \frac{\temperature}{2\, \pdf(\th_k|\eth_k)}\, \nabla_{\th} \reward_k(\th_k) + \sqrt{\stepa} \noise_k , \quad
 \text{ where }   \th_k = \eth_k + \obsnoise_k
\end{split}
\eeq
The  proof of convergence again follows using averaging theory arguments. Since $\{\th_k\} \sim \pdf(\th|\eth_k)$ is the fast signal and $\{\eth_k\}$ is the slow signal, the averaged system  is
$$ d\eth(t) =  \int_{\reals^\thdim} \kerneln\bigl(\frac{\th - \eth}{\kernelstep}\bigr)\ \frac{\temperature}{2\,\pdf(\th|\eth(t))}\,\nabla_\th \Reward(\th)\, {\pdf(\th|\eth(t))} \,d\th\, dt + d\bm(t)  $$
So the $\pdf(\th|\eth(t))$ cancel out in the numerator and denominator.  As $\kernelstep \rightarrow 0$,
the kernel acts as
a  Dirac delta thereby yielding the classical Langevin diffusion (\ref{eq:classicaldiffusion}).

{\em Remark}: The active IRL algorithm (\ref{eq:activeirl}) can be viewed as an idealization of the multi-kernel IRL algorithm (\ref{eq:mcmcirl}). The multi-kernel algorithm  constructs weights to  approximate  sample from the conditional distribution $\pdf(\th|\eth)$. In comparison, the active IRL has direct measurements from this conditional density. So the active IRL can be viewed as an upper bound to the performance of the multi-kernel IRL
Another motivation is inertia. Given the dynamics of the RL algorithm, it may not be possible to the RL to abruptly  jump to evaluate a gradient at $\eth_k$, at best the RL can only evaluate a gradient at a point $\eth_k + \obsnoise_k$. A third motivation stems from mis-specification:  if the IRL represents a machine (robot) learning from a human, it is difficult to specify to the human exactly what policy
$\eth_\dtime$  to    perform. Then $\th_k = \eth_k + \obsnoise_k$ can be viewed as an approximation to this  mis-specification.

     \subsubsection{Non-reversible Diffusion for IRL} So  far we have defined four  different  passive Langevin dynamics algorithms for IRL, namely (\ref{eq:irl}),
 (\ref{eq:irl2}),  (\ref{eq:mcmcirl}), and (\ref{eq:activeirl}).
These algorithms
yield  reversible diffusion processes that asymptotically sample from the stationary distribution
(\ref{eq:stationary1}).
  It is well  known \citep{HHS93,HHS05,Pav14} that adding a skew symmetric matrix to the gradient always improves the convergence rate of Langevin dynamics to its stationary distribution.
  That is for any $\thdim\times \thdim$ dimensional  skew symmetric matrix $\skew = -\skew^\p$, the non-reversible diffusion process
  \beq
   d\eth(t) = \frac{\temperature}{2} \, (I_\thdim + S) \, \nabla_\eth \Reward(\eth)  dt  +   d \bm(t) , \quad \eth(0) = \eth_0
 \label{eq:skew1}
\eeq
has a larger spectral gap and therefore converges to stationary distribution $\belief(\eth)$ faster than (\ref{eq:gld}).  The resulting IRL algorithm obtained by a  Euler-Maruyama time discretization of (\ref{eq:skew1})  and then introducing a kernel $\kernel(\cdot)$ is
  \begin{equation}
 \boxed{  \eth_{\dtime+1} = \eth_\dtime +  \stepa  \, \kerneln\big(\frac{\th_\dtime-\eth_\dtime}{\kernelstep}\bigr)\,
 \frac{\temperature\,(I_\thdim+ \skew)}{2\, \belief(\eth_\dtime)}\, \hnabla_\th \reward_\dtime(\th_\dtime) +  \sqrt{\stepa}\,  \noise_\dtime,
    \quad \dtime =1,2,\ldots }
\label{eq:irl3}
\end{equation}
initialized by  $\eth_0 \in \reals^\thdim$.
Again  a stochastic  averaging theory argument  shows that IRL algorithm (\ref{eq:irl3}) converges weakly to the
non-reversible diffusion (\ref{eq:skew1}).
In numerical examples, we  found empirically that the convergence of (\ref{eq:irl3})  is  faster than (\ref{eq:irl}) or~(\ref{eq:irl2}). However,
the faster convergence comes at the expense of  an order of magnitude increased computational cost. The  computational cost of IRL algorithm  (\ref{eq:irl3}) is $O(\thdim^2)$ at each iteration due to  multiplication with  skew symmetric matrix $\skew$. In comparison  the computational costs of IRL
algorithms (\ref{eq:irl}) and (\ref{eq:irl2}) are each $O(\thdim)$.

\section{Numerical Examples} \label{sec:numerical}
This section presents three examples to illustrate the performance of the proposed IRL algorithms.

\subsection{Example 1. IRL  for Bayesian KL divergence and Posterior Reconstruction} \label{sec:bayesian}

 This section illustrates the performance of our proposed IRL algorithms in reconstructing the Kullback Leibler (KL) divergence and multi-modal posterior distribution. Our formulation is  a stochastic  generalization of  adaptive Bayesian learning in \cite{WT11} as  explained below.

{\color{black}{{\bf Motivation.} Exploring and estimating the KL divergence of a multimodal posterior distribution is important in Bayesian inference \citep{RC13},  maximum likelihood estimation, and also stochastic control with KL divergence cost \citep{GRW14}.
To motivate the problem,
suppose random variable  $\th$ has prior probability density $ \pdf(\th)$.
Let $\thtrue$ denote a fixed (true) value of $\th$ which is unknown to the optimizing agents and inverse learner.
Given a sequence of observations
$\obs_{1:\horizon} = (\obs_1,\ldots,\obs_\horizon)$,  generated from distribution $\pdf(\obs_{1:\horizon}|\thtrue)$, the KL divergence of the posterior distribution is
\beq \label{eq:kldiv}
 J(\thtrue,\th)=
\E_\thtrue\{ \log \pdf(\thtrue| \obs_{1:\horizon})  -
 \log \pdf(\th| \obs_{1:\horizon})
\}  = \int
\log \frac{\pdf(\thtrue|\obs_{1:\horizon})}
{\pdf(\th|\obs_{1:\horizon})}
\, \pdf(\obs_{1:\horizon}|\thtrue) d\obs_{1:\horizon}
\eeq
It is well known  (via Jensen's inequality) that the global minimizer $\th^*$ of  $J(\thtrue,\th)$ is
$\thtrue$. Therefore minimizing the KL divergence yields a consistent estimator
of $\thtrue$. Moreover, under mild stationary conditions,   when the prior is non-informative (and so possibly improper), the Shannon-McMillan-Breiman theorem \citep{Bar85} implies that
the global minimizer of the  KL divergence converges with probability 1 to the maximum likelihood estimate as $\horizon \rightarrow \infty$. So there is strong motivation to explore and estimate the KL divergence.

Typically the KL divergence $J(\thtrue,\th)$ is non-convex in $\th$. So we are in the non-convex optimization setup of (\ref{eq:rl})  where multiple agents seek to estimate the global
minimizer of the KL divergence.}}

\subsubsection{Model Parameters}

We consider a stochastic optimization problem where a RL system chooses actions
$\action_k$
from randomized policy $\pdf(\th|\obs_{1:\horizon})$. In order to learn the optimal policy, the RL system aims to
estimate the global minimizer
$ \th^* = \argmin_\th J(\thtrue,\th) $; see for example \cite{GRW14} for motivation of KL divergence minimization in stochastic control.
Then by observing the  gradient estimates of the RL agents, we will use our proposed passive
IRL algorithms to reconstruct the KL divergence.

Ignoring the constant term  $\pdf(\thtrue| \obs_{1:\horizon}) $ in (\ref{eq:kldiv}),
minimizing $J(\thtrue,\th)$ wrt $\th$ is equivalent to maximizing the relative
entropy $\Reward(\th) = \E_{\thtrue}\{ \log \pdf(\th|\obs_{1:\horizon})\}$.
So  multiple RL agents aim to 
solve the following  non-concave stochastic maximization problem: Find
\beq
\th^* = \argmax_\th  \Reward(\th), \quad \text{ where } \Reward(\th)  = \E_{\thtrue}\{ \log \pdf(\th|\obs_{1:\horizon})\}
\label{eq:relative}
\eeq
In our numerical example we choose  $\th =[\th(1),\th(2)]^\p\in \reals^2$  and $\thtrue$ is  the true parameter value which is  unknown to the learner. The prior is $\pdf(\th) = \normal(0,\Sigma)$ where $\Sigma=\diag[10,2]$. The observations $\obs_\dtime$ are independent and generated from the multi-modal mixture likelihood
$$ \obs_\dtime \sim \pdf(\obs|\thtrue) = \frac{1}{2} \normal(\thtrue(1), 2) + \frac{1}{2} \normal(\thtrue(1) + \thtrue(2), 2) $$
Since $\obs_1,\ldots,\obs_\horizon$ are independent and identically distributed,
 the objective $\Reward(\th)$ in (\ref{eq:relative})  is
 \beq  \Reward(\th) =  \E_\thtrue\{ \log \pdf(\th) +  \horizon \, \log \pdf(\obs|\th) \} + \text{ constant indpt of $\th$}
 \label{eq:jtheta}
\eeq
For  true parameter value  $\thtrue=[0, 1]^\p$, it can be verified that
the
objective  $\Reward(\th)$ is non-concave and has two maxima at
$\th = [0, 1]^\p$ and $\th=[1, -1]^\p$.

\subsubsection{Classical Langevin Dynamics}
\begin{figure}[p] \centering
  \begin{subfigure}{.45\textwidth}
    \includegraphics[scale=0.5]{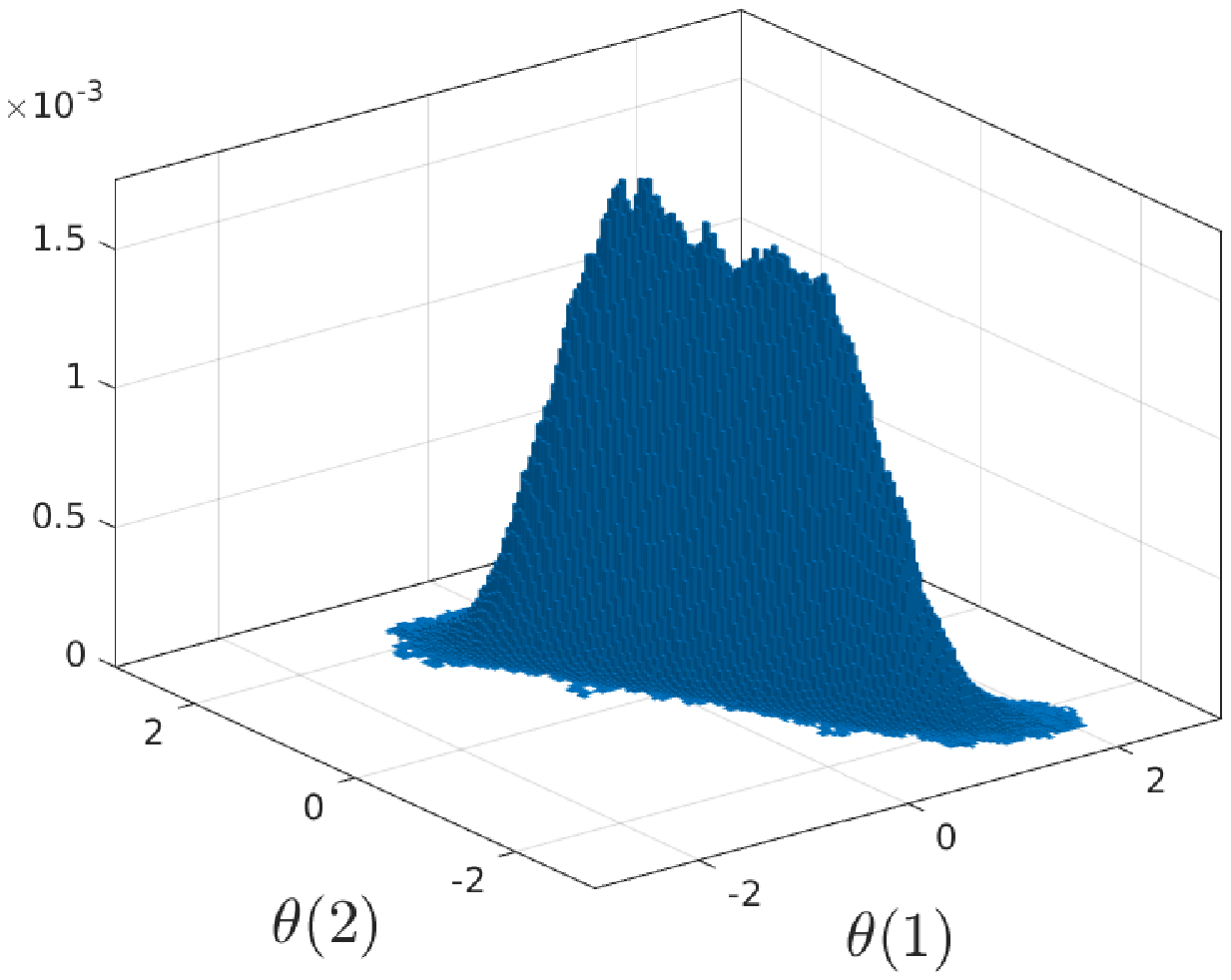}
  \end{subfigure}
  \begin{subfigure}{.45\textwidth}
    \includegraphics[scale=0.5]{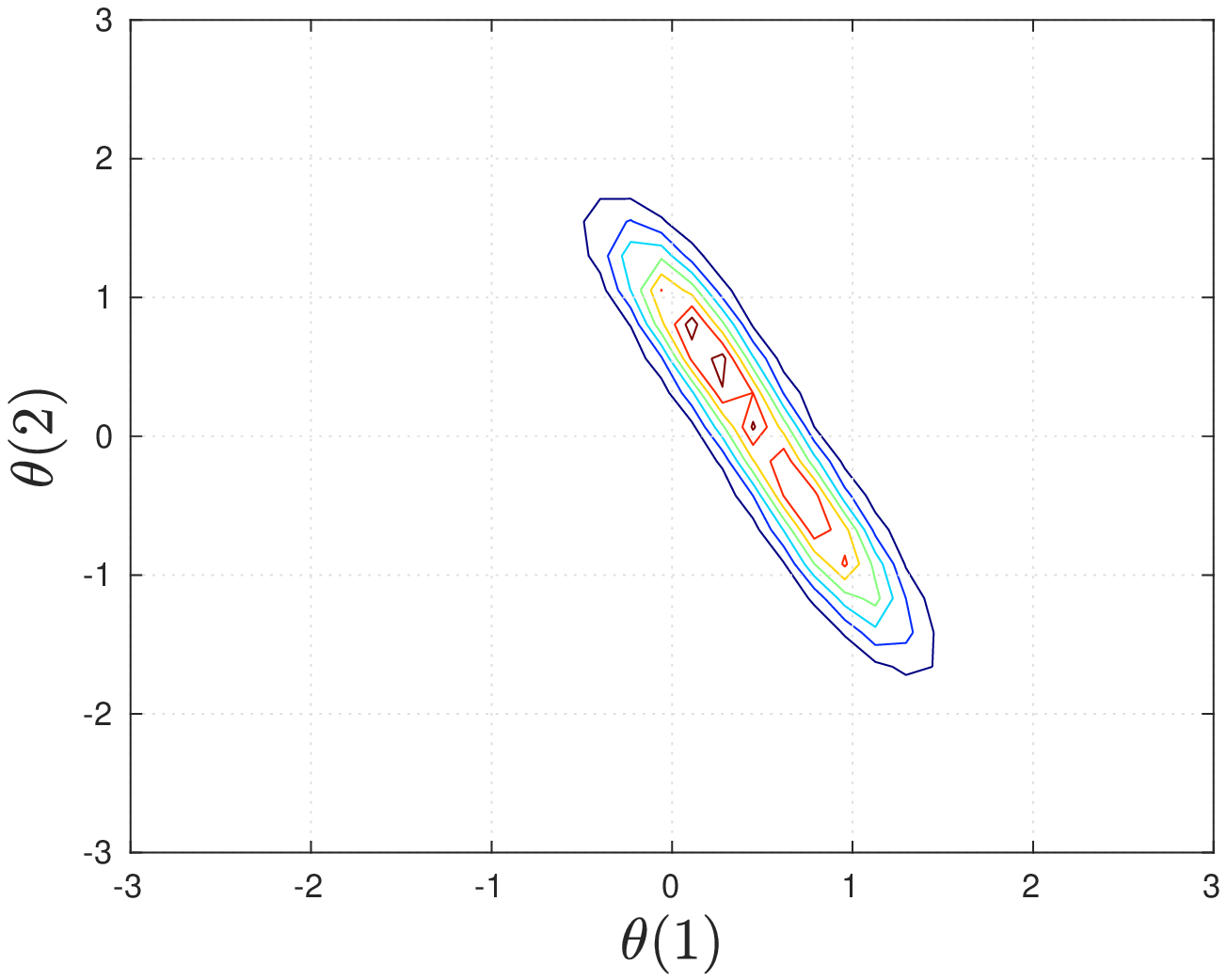}
  \end{subfigure}
  \caption{Classical Langevin dynamics (ground truth)}
  \label{fig:classicalLang}
\end{figure}

To  benchmark the performance of our passive  IRL algorithms (discussed below),
we  ran the classical Langevin dynamics algorithm:
\begin{equation} \label{eq:classical_langevin}
   \th_{\dtime+1} = \th_\dtime + \stepa\,  \frac{\temperature}{2}  \nabla_\th \reward_\dtime(\th_\dtime)
   + \sqrt{\stepa}\,   \noise_\dtime,
   \quad \dtime =1,2,\ldots,
\end{equation}
Note that the classical Langevin dynamics \eqref{eq:classical_langevin}
evaluates the gradient estimate $\nabla_\th \reward_\dtime(\th_\dtime)$  unlike our passive IRL algorithm which has no control of where the gradient is evaluated.
Figure \ref{fig:classicalLang}  displays
both the empirical histogram and a contour plot of the estimate $\Reward(\th)$ generated by classical Langevin dynamics. The classical Langevin dynamics can be viewed as an upper bound for the performance of our passive IRL algorithm; since our passive algorithm cannot specify where the gradients are evaluated.

\begin{figure}[p] \centering
  \begin{subfigure}{.45\textwidth}
     \includegraphics[scale=0.5]{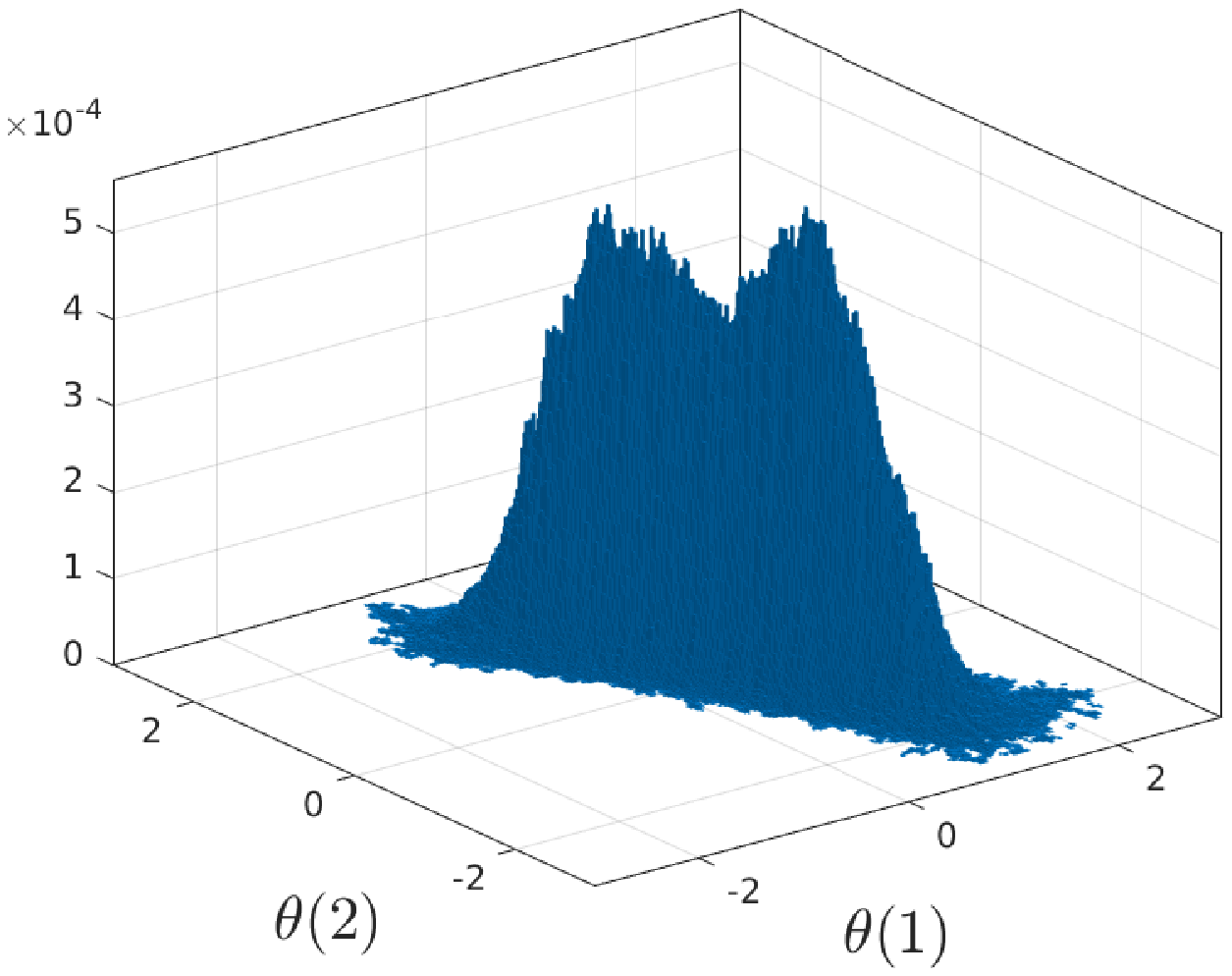}
  \end{subfigure}
  \begin{subfigure}{.45\textwidth}
        \includegraphics[scale=0.5]{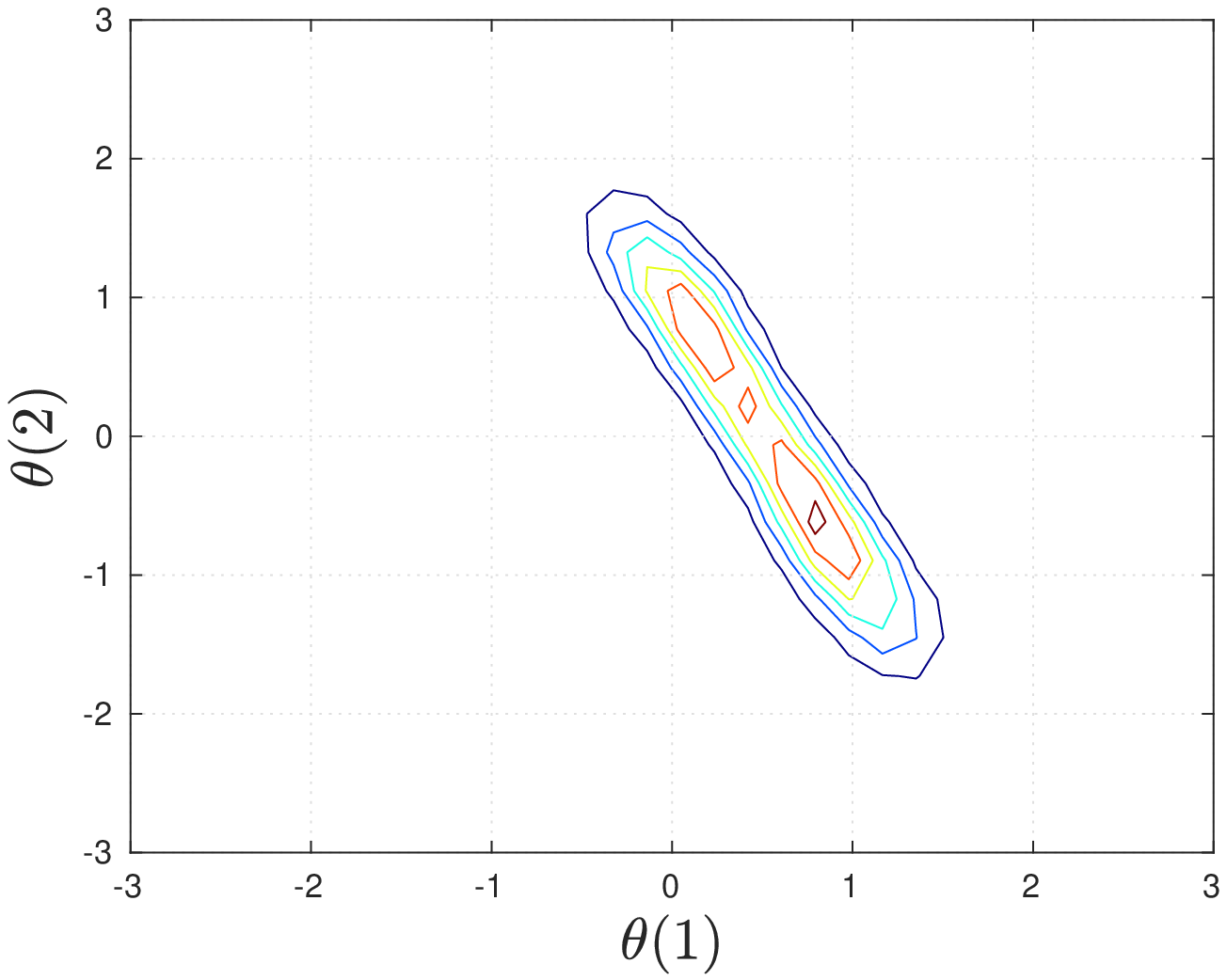}
  \end{subfigure}
  \caption{IRL Algorithm (\ref{eq:irl})}
  \label{fig:mh}
\end{figure}

\begin{figure}[p]
   \begin{subfigure}{.45\textwidth}
    \includegraphics[scale=0.5]{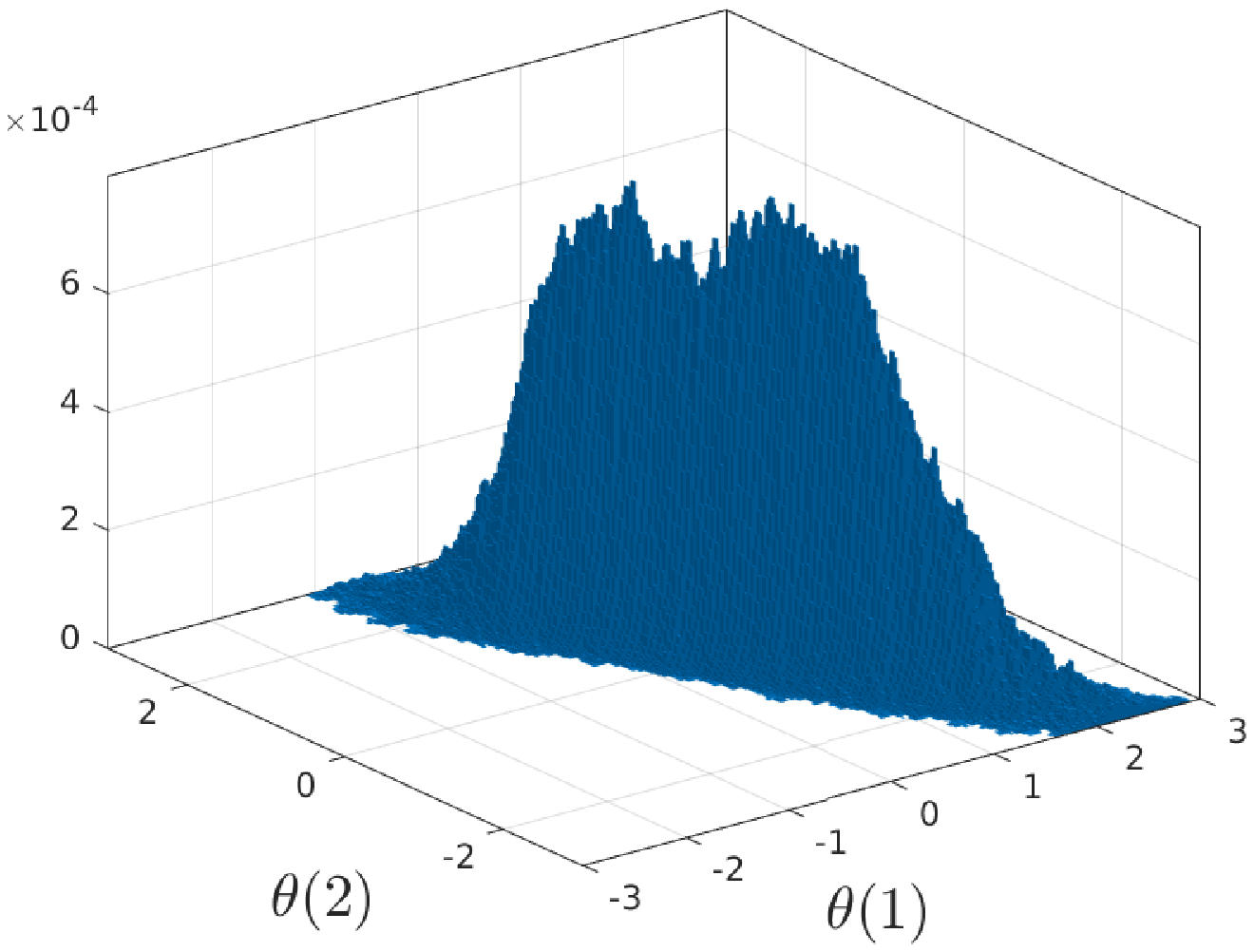}
  \end{subfigure}
  \begin{subfigure}{.45\textwidth}
    \includegraphics[scale=0.5]{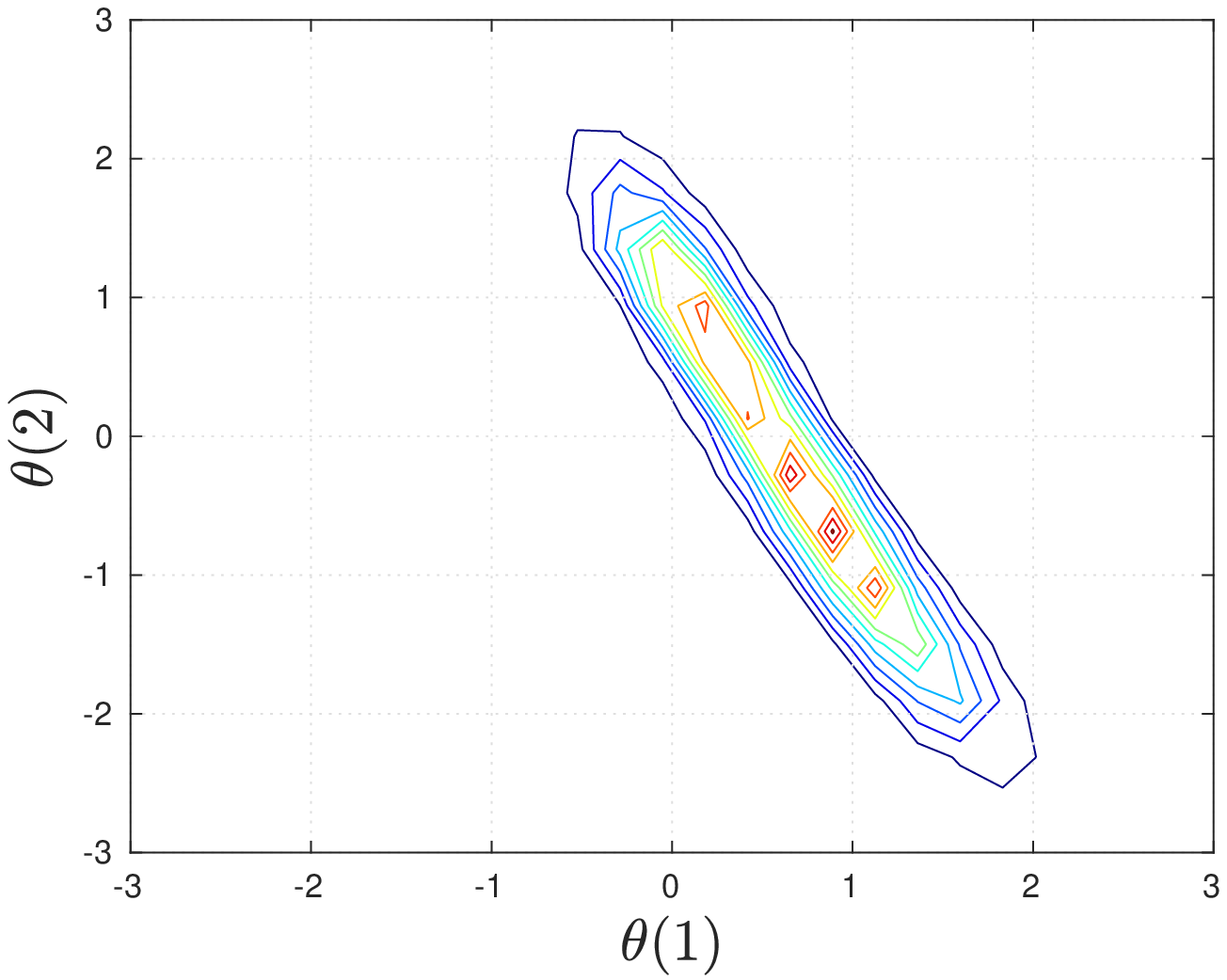}
  \end{subfigure}
  \caption{Two time scale multi-kernel IRL Algorithm (\ref{eq:mcmcirl})}
  \label{fig:mh2}
\end{figure}

\subsubsection{Passive IRL Algorithms}

We now illustrate the performance of our proposed passive IRL algorithms for the above model.
Recall that the framework comprises two parts: First, multiple RL agents run randomly initialized stochastic gradient algorithms to maximize $\Reward(\th)$.  Second, by observing these gradients, our passive IRL Langevin based algorithms construct a non-parametric estimate of the $\Reward(\th)$. We discuss these two parts below:

{\em 1. Multiple agent Stochastic Gradient Algorithm}.
Suppose  multiple RL agents aim to learn the optimal policy by estimating the optimal parameter $\th^*$. To do so, the  agents use the stochastic gradient algorithm (\ref{eq:rl}):
\begin{equation} \label{eq:rlsim}  \begin{split}
\th_{\dtime+1} &= \th_\dtime + \step \nabla_\th \reward_\dtime(\th_\dtime)
\\
\nabla_\th \reward_\dtime(\th_\dtime) &=   \nabla_\th \log  \pdf(\th_\dtime) + \horizon\, \nabla_\th \log \pdf(\obs_\dtime|\th_\dtime)
\end{split}
\end{equation}
with multiple random initializations, depicted by agents $\tslow=1,2,\ldots$.
For each agent $\tslow$, the initial estimate  was sampled randomly as $\th_{\stoptime_\tslow} \sim \belief(\cdot) =  \normal(0,I_{2\times 2})$. Each agent runs the gradient algorithm for 100 iterations with step size  $\step = 10^{-3}$ and the number of agents is  $10^5$.  Thus the sequence $\{\th_\dtime;\dtime=1,\ldots10^7\}$ is generated. \\

{\em 2. IRL algorithms and performance}.
 Given   the sequence of estimates $\{\th_\dtime\}$ generated by the RL agents above, and initialization density $\belief$,  the inverse learner aims to estimate $\Reward(\th)$  in (\ref{eq:jtheta}) by
 generating samples $\{\eth_k\}$ from
 $\exp(\temperature \Reward(\th))$. Note that the IRL algorithm
 has no knowledge of $\pdf(\th)$ or $\pdf(\obs|\th)$.
 Since the inverse learner has no control of where the reinforcement learner evaluates its gradients,  we are in passive IRL setting.
We ran
  the IRL algorithm (\ref{eq:irl}) with kernel
$ \kernel (\th,\eth)\propto \exp(- \frac{\|\eth- \th\|^2}{0.02})$, step size $\stepa = 5 \times 10^{-4}$, $\temperature=1$.
Figure \ref{fig:mh} displays both the empirical histogram and a contour plot.
Notice that the performance of our IRL is  very similar to classical Langevin dynamics (where the gradients are fully specified).

{\color{black}{We compared the performance of the classical Langevin with the passive Langevin IRL algorithm averaged over 100 independent runs. The comparison is with respect to the variational distance\footnote{Recall the variational distance is half the $L_1$ norm} $d(1)$ and $d(2)$ between the two marginals of the empirical
density $\pdf(\th)  \propto \exp(\Reward(\th))$. The values obtained from
our simulations
are
\beq d(1) = 0.0122, \quad d(2)= 0.0202 . \label{eq:l1dist}\eeq
}}
Finally,  we illustrate the performance of the two-time scale multikernel algorithm (\ref{eq:mcmcirl}). Recall this algorithm does not require knowledge of the initialization probabilities $\belief(\cdot)$. Figure \ref{fig:mh2} displays
both the empirical histogram and a contour plot. Again the performance of the IRL is  very similar to the classical Langevin dynamics performance.

\subsubsection{Multiple Inverse Learners}
We also considered the case where  multiple inverse learners act in parallel. Suppose each inverse learner $l \in \{1,2,\ldots, L\}$ deploys IRL algorithm (\ref{eq:irl})
with its own noise  sample path denoted by $\{\noise_k^{(l)}\}$, which is independent of
other inverse learners. Obviously, if the estimate $\eth_k^{(l)}$ of one of the  inverse learners
(say $l$) is close to
$\th_k$, then $\nabla_\th \reward_\dtime(\th_k)$ is  a more accurate gradient estimate
for $\nabla_\th \reward_\dtime(\eth_k^{(l)})$. However, for high dimensional problems, our numerical experiments (not presented here) show  very little benefit unless the number of inverse learners is chosen as $L = O(2^\thdim)$ which is intractable.

{\color{black}
\subsubsection{IRL for Adaptive   Bayesian Learning} \label{sec:irlabl}

Having discussed reconstructing the KL divergence via IRL, we now discuss how to   extend  the
Bayesian learning framework proposed in \citet{WT11}  to our IRL framework.

{\bf Bayesian Learning}. First a few words about the Bayesian learning framework in \citet{WT11}.
In comparison to the stochastic optimization problem (\ref{eq:relative}),
 They consider
 a \textit{fixed} sample path $\obs_{1:\horizon}$ and  the associated {\em deterministic} optimization problem of finding  global maximizers of
\beq  \label{eq:rllog1}  \Reward(\th) =  \log \pdf(\th|\obs_{1:\horizon}). \eeq
 \citet{WT11}  use the
 classical Langevin dynamics to generate samples from
 the posterior  $\pdf(\th|\obs_{1:\horizon})$ as follows:
First,
since $\obs_1,\ldots, \obs_\horizon$ are independent,
\beq  \nabla_\th \log \pdf(\th|\obs_{1:\horizon}) \propto   \nabla_\th \log  \pdf(\th) +  \sum_{\dtime=1}^\horizon \nabla_\th \log \pdf(\obs_{\dtime}|\th)
\label{eq:gradlang}
\eeq
Next it is straightforward to see that $\horizon$ iterations of the  classical Langevin algorithm (or a fixed step size deterministic gradient ascent algorithm) using the  gradient  $\nabla_\th \log  \pdf(\th) +
\horizon\, \nabla_\th \log \pdf(\obs_{k}|\th)$  is identical to running $\horizon$ sweeps
of the  algorithm through the sequence $\obs_{1:\horizon}$ with gradient (\ref{eq:gradlang}).
So \citet{WT11} run the classical Langevin algorithm using the  gradient  $$\nabla_\th \log  \pdf(\th) +
\horizon\, \nabla_\th \log \pdf(\obs_{k}|\th). $$
  Notice unlike the KL estimation framework (\ref{eq:relative}) which has an expectation $\E_\thtrue$ over the observations,   the underlying optimization of $\log \pdf(\th|\obs_{1:\horizon})$
 is {\em deterministic}
 since we have a  fixed sequence  $\obs_{1:\horizon}$.
Then clearly  the Langevin dynamics generates samples from the stationary distribution
 \beq \belief(\th) = \exp\big( \log (\pdf(\th| \obs_{1:\horizon}) ) \big) = \pdf(\th| \obs_{1:\horizon}) \label{eq:wt11}\eeq
  namely, the posterior distribution.\footnote{This is in contrast to our KL divergence estimation setup
    \eqref{eq:jtheta} where the stationary distribution is  $\belief(\th) = \exp\big( \E_\thtrue \{\log \pdf(\th| \obs_{1:\horizon}) \}\big) $ and
    $\E_\thtrue $ denotes expectation wrt $\pdf(\obs_{1:\horizon}|\thtrue)$.}
So  the classical Langevin algorithm which  sweeps repeatedly through the dataset $\obs_{1:\horizon}$ generates samples
  from the posterior distribution - this is the main idea of Bayesian learning in \citet{WT11}.

  {\bf IRL}.
 We now consider  IRL in this Bayesian learning framework to reconstruct the posterior density.
Given the sample path $\obs_{1:\horizon}$,
suppose multiple forward learners seek to estimate the maximum (mode) of the multimodal posterior $\log \pdf(\th|\obs_{1:\horizon})$.
The agents run the (deterministic) gradient ascent algorithm (\ref{eq:rl})
 with gradient
 $$  \nabla_\th \reward_\dtime(\th_\dtime) = \nabla_\th \log  \pdf(\th_\dtime) + \horizon\, \nabla_\th \log \pdf(\obs_\dtime|\th_\dtime)$$
The IRL problem we consider is:  By passively observing these gradients, how can the IRL algorithm reconstruct the posterior
 distribution  $\pdf(\th|\obs_{1:\horizon})$?
We use  our IRL algorithm (\ref{eq:irl}).
  The implementation of  IRL algorithm (\ref{eq:irl}) follows the   \citet{WT11} setup: The RL agents choose random initializations $\th_0 \sim \belief$ and  then run gradient algorithms  sweeping repeatedly through the dataset $\obs_{1:\horizon}$. The IRL algorithm (\ref{eq:irl}) passively views these estimates $\{\th_k\}$ and  reconstructs the posterior distribution  $p(\th| \obs_{1:\horizon})$  from these estimates.




\subsection{Example 2. IRL with Logistic Regression Classifier} \label{sec:logistic}
We now  consider a high dimensional IRL problem ($\thdim = 124)$ on the   benchmark adult {\tt a9a} dataset. Performing IRL, i.e., generating samples from a $124$ dimensional probability density that represents the utility, is challenging and requires use of the multi-kernel variance reduced IRL algorithm (\ref{eq:mcmcirl}).

\subsubsection*{Setup}
 In a logistic regression model parameterized by $\th \in \reals^\thdim$,
the observations (labels)
$y_\dtime \in \{0,1\}$  are assumed to be generated probabilistically  from
$$P(y_\dtime=1| \theta) = \sigma(\psi_\dtime^\p \th) = \frac{1}{1 + \exp(-\psi_\dtime^\p \th)} , \quad \th \in \reals^\thdim$$
Here $\psi_k \in \reals^\thdim$ is known input vector at time $k$ and is called the feature.

We consider a Bayesian setting where the prior of $\th$ is assumed to be an $\thdim$-variate  Laplacian density with independent components. So the prior is
$$ \pdf(\th)  \propto \exp(-\sum_{i=1}^\thdim |\th(i)|) .$$
As in the Bayesian learning setup (\ref{eq:rllog1})  above, given the fixed sequence $\obs_{1:\horizon}$, the  RL agents aim to find the global maximizer of
\beq   \Reward(\th) =  \log \pdf(\th | \obs_{1:\horizon}) 
\label{eq:rllog}
\eeq
To do so, the RL agents use the gradient algorithm
\beq  \th_{\dtime+1} = \th_\dtime + \step \bigl[  \nabla_\th \log  \pdf(\th_\dtime) + \horizon\, \nabla_\th \log \pdf(\obs_\dtime|\th_\dtime)\bigr].
\label{eq:forwardagents}
\eeq
with multiple sweeps over the dataset.
Note that for the logistic model,
$  \nabla_\th \log  \pdf(\th_\dtime) = -\sgn(\th)$ elementwise and $ \nabla_\th \log \pdf(\obs_\dtime|\th_\dtime) =  \psi_\dtime  \big(y_\dtime -  \sigma(\psi_\dtime^\p \theta_\dtime) \big)   $.

\subsubsection*{Dataset}  We consider the benchmark  adult {\tt a9a} dataset
which  can be downloaded from
\newline {\small
\url{https://www.csie.ntu.edu.tw/~cjlin/libsvmtools/datasets/binary.html}.}

The dataset consists of a time series of  binary valued (categorical) observations ${y}_k \in \{0,1\}$   and a time series of  regression vectors $\bar{\psi}_\dtime \in \reals^{123}$  for $k=1,\ldots, 32651$.
 To model the bias, we add one additional component; so the unknown parameter vector is   $\th\in \reals^{124}$ and
 the  augmented regression vectors are   $\psi_\dtime= \begin{bmatrix} 1 \\ \bar{\psi}_\dtime \end{bmatrix} \in \reals^{124}$, for $k=1,\ldots, 32651$.

 \subsubsection*{Performance of IRL Algorithm (\ref{eq:mcmcirl})}

 Suppose the inverse learner observes the estimates  $\{\th_k\}$ generated by the  RL agents according to (\ref{eq:forwardagents}). The inverse learner aims to reconstruct the posterior
 $\pdf(\th|\obs_{1:\horizon})$.
 Since $\thdim =124$, the  IRL algorithm  needs to explore and sample from
 a 124-variate distribution which is a formidable task. The vanilla IRL algorithm (\ref{eq:irl}) is not tractable since it would take a prohibitive number of iterations to converge.
 We illustrate the performance of the multi-kernel variance reduction IRL algorithm~(\ref{eq:mcmcirl}).

\begin{figure}[h] \centering
\begin{subfigure}{.45\textwidth}
\includegraphics[scale=0.5]{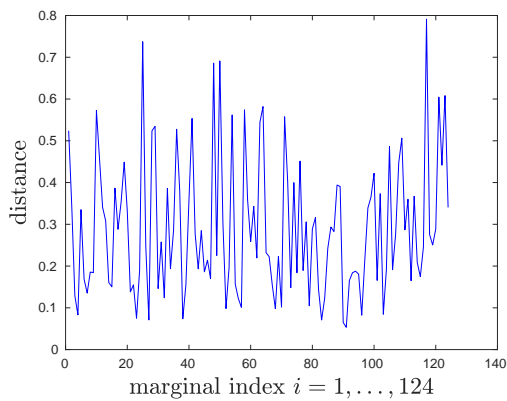}
\caption{Active  IRL Algorithm (\ref{eq:activeirl}) vs ground truth. Wasserstein 1 distance (\ref{eq:L1}) of each of the 124 marginals}
\end{subfigure} \hspace{0.4cm}
\begin{subfigure}{.45\textwidth}
\includegraphics[scale=0.5]{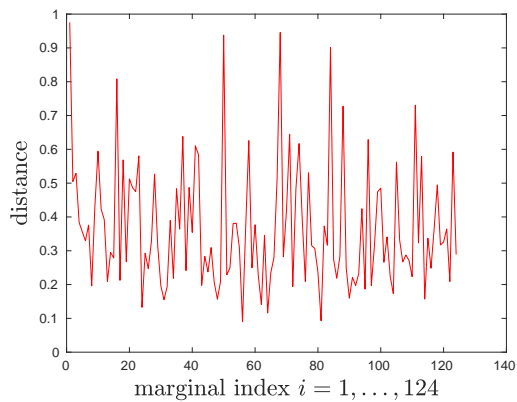}
\caption{Multikernel Algorithm (\ref{eq:mcmcirl}) vs ground truth. Wasserstein 1 distance (\ref{eq:L1}) of each of the 124 marginals}
\end{subfigure}
\begin{subfigure}{.6\textwidth}
\includegraphics[scale=0.6]{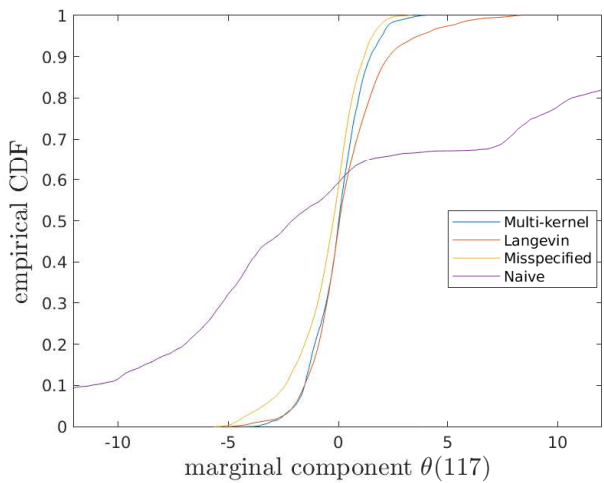}
\caption{Comparison of  117th marginal}
\end{subfigure}
\caption{Comparison of multi-kernel IRL Algorithm (\ref{eq:mcmcirl}) and  active IRL algorithm (\ref{eq:activeirl}) with classical Langevin (\ref{eq:standard_ld}) (ground truth)}
\label{fig:logistic}
\end{figure}

We ran  multi-kernel IRL algorithm  (\ref{eq:mcmcirl}) and active IRL algorithm (\ref{eq:activeirl}) on the {\tt a9a} dataset. As mentioned in Sec.\ref{sec:activeirl}, the active IRL (\ref{eq:activeirl}) is an idealization of the multikernel IRL algorithm  (\ref{eq:mcmcirl}) and so  forms a benchmark for it.
 The parameters were chosen as
 $ \stepa = 2.5\times 10^{-4}$, $ \belief(\th) = \normal(0,I)$, $\sigma=0.1 $, $\numparticles = 100$ in
 (\ref{eq:mcmcirl})
 and  $\horizon =10$ in  (\ref{eq:rllog}).
 As in \cite{WT11}, we ran 10 ``sweeps'' through the dataset. That is, we  appended 9  repetitions of the data set resulting in a single dataset of $10 \times 32651$ time points; and then  ran the IRL algorithms on this appended dataset.

To benchmark these algorithms, we also ran the classical Langevin dynamics algorithm:
\begin{equation}
   \eth_{\dtime+1} = \eth_\dtime + \stepa\,  \frac{\temperature}{2}  \nabla \reward_\dtime(\eth_\dtime)
   + \sqrt{\stepa}\,   \noise_\dtime,
    \quad \dtime =1,2,\ldots,
\label{eq:standard_ld}
\end{equation}
which corresponds to the ground truth (since  the gradients are evaluated at $\eth_k$).

IRL Algorithms (\ref{eq:mcmcirl}) and  (\ref{eq:activeirl}) generate samples $\{\eth_k\}$ from a 124-dimensional distribution. To visualize the performance,  we used the output sequence $\{\eth_k\}$   from these algorithms to compute the empirical cumulative distribution functions for each of the 124  marginal distributions, denoted by $\hat{F}_i(\eth(i))$, $i=1,\ldots,124$.
For each such marginal empirical distribution, we then computed the corresponding marginal from the classical Langevin dynamics  (\ref{eq:standard_ld}), denoted as ${F}_i(\eth(i))$; this can be viewed as  the ground truth.
Finally, we computed the $L_1$ distance  (Wasserstein 1-metric)
\beq
d(i)  = \int  | \hat{F}_i(\eth(i)) - F_i(\eth(i)) |\, d\eth(i), \quad i = 1,\ldots, 124.
\label{eq:L1}
\eeq
This $L_1$ distance is more appropriate for our purposes than the Kolmogorov-Smirnov distance since
typically the constant or proportionality $\temperature$ is not known and so the regions of support of the empirical cumulative distribution functions can vary substantially.

Figure \ref{fig:logistic}(a) and (b) display  the $L_1$ distance $d(i)$ vs $i=1,2,\ldots,124$ for the IRL Algorithms (\ref{eq:mcmcirl}) and  (\ref{eq:activeirl}). In a sense,
Algorithm (\ref{eq:activeirl}) can be viewed as an upper bound for the performance of Algorithm (\ref{eq:mcmcirl}) since the conditional density $\pdf(\th|\eth_k)$ used to generate
$\th$ in Algorithm (\ref{eq:activeirl})
is exactly the same kernel used in
Algorithm (\ref{eq:mcmcirl}).  As can be seen from Figure \ref{fig:logistic}(a) and (b), the two algorithms perform
similarly, despite the fact that
Algorithm (\ref{eq:mcmcirl}) has no control over where the derivative is evaluated.  This shows that the IRL algorithm is a viable method  for sampling  from the high-dimensional Bayesian posterior; or equivalently estimating $ J(\th,\thtrue)$ in (\ref{eq:rllog}).  Finally, Figure \ref{fig:logistic}(c) shows the marginal distribution for the 117-th component of $\th$ for the classical Langevin (ground truth), active IRL (mis-specified), multi-kernel IRL and a naive Langevin. By naive Langevin we mean the Langevin algorithm that uses the gradient $\nabla_\th \reward_\dtime(\th_k)$ instead of  $\nabla_\th \reward_\dtime(\eth_k)$ at the estimate $\eth_k$, without any kernel. We see that the multi-kernel and active IRL are close to the ground truth (Langevin) while the naive IRL performs very poorly (since it completely disregards the fact that the gradients evaluated at  $\eth_k$ and $\th_k$ are different).

\subsection{Example 3. IRL for Constrained Markov Decision Process (CMDP)}  \label{sec:mdp}
In this section we illustrate the performance of the IRL algorithms for reconstructing the cumulative reward of a constrained Markov decision process (CMDP) given gradient information from a RL algorithm.
This is in contrast to classical IRL \citep{NR00} where the transition matrices of the MDP are assumed known to the inverse learner.

Consider  a unichain\footnote{By {\em unichain} \cite[pp.~348]{Put94} we mean that every policy where $\action_n$  is a deterministic function of $\state_n$ consists of a
single recurrent class  plus possibly an empty set of transient states.} average reward CMDP  $\{\state_n\}$ with finite state space $\statespace
= \{1,\ldots,\statedim\}$  and  action space $\actionspace= \{1,2,\ldots,\actiondim\}$. The CMDP evolves with
transition probability matrix $\tp(\action)$ where
\begin{equation}
\label{Aij}
\tp_{ij}(\action) \defn \prob[\state_{n+1} = j | \state_n = i, \action_n = \action],  \quad u \in \actionspace.
\end{equation}
When the system is in state $\state_n\in \statespace$, an action $\action_n = \policy(\state_n) \in \actionspace$ is chosen, where $\policy$ denotes (a possible randomized)  stationary policy.
The reward incurred at stage $n$ is  $\mreward(\state_n,\action_n)\geq 0$.

Let $\admissible$ denote the class of stationary randomized Markovian policies.
For any stationary  policy $\policy \in \admissible$, let $ \esp_{\policy}$ denote the corresponding expectation and define the infinite horizon average reward
\begin{equation}
   \label{J}
J(\policy) = \lim_{\finaltime\to\infty} \inf \frac{1}{\finaltime} \esp_{\policy} \Big[\sum_{n=1}^\finaltime
\mreward(\state_n,\action_n) \mid \state_0 = x\Big].
\end{equation}
Motivated by modeling fairness constraints  in network optimization  \citep{NK10},
 we consider the reward  (\ref{J}), subject to the
average constraint:
\begin{equation}
 \label{costconstraint}
  \Cons(\policy) =  \lim_{\horizon\to\infty} \inf {1\over \horizon} \esp_{\policy}\Big[ \sum_{n=1}^\horizon\con(\state_n,\action_n)
\Big] \leq \rcon,
\end{equation}
(\ref{J}), (\ref{costconstraint}) constitute a CMDP. Solving a CMDP involves computing
the optimal policy $\policy^* \in \admissible$ that satisfies
\begin{equation}
J(\policy^*) = \sup_{\policy \in \admissible} J(\policy) \quad
\forall x_0 \in \statespace,  \text{ subject to } (\ref{costconstraint}) \label{eq:objective1}
\end{equation}

To solve a  CMDP,  it is sufficient to consider randomized stationary policies:
\begin{equation} \label{eq:randpol}
 \policy(\state)  =  \action \text{ with probability } \; \cond(\action|\state)= \frac{\statpi({\state,\action})}{\sum_{\tilde \action \in \actionspace} \statpi({\state,\tilde \action})} ,\end{equation}
where the conditional probabilities $\cond$ and joint probabilities $\statpi$ are defined as
\beq  \cond(\action|\state)  = \prob(\action_n =\action| \state_n = \state),  \quad \statpi(\state,\action) = \prob(\action,\state). \label{eq:conditionalprob} \eeq
Then the optimal policy $\optpolicy$ is obtained as the solution of a   linear programming problem in terms of the $\statedim \times \actiondim$ elements of $\statpi$; see \cite{Put94} for the precise equations.

Also  \citep{Alt99}, the optimal policy $\policy^*$ of the CMDP
is {\em randomized} for at most one of the states. That is,
\begin{equation}
\optpolicy(\state)  = \randmix \,\optpolicy_1(\state)  + (1 - \randmix)\, \optpolicy_2(\state)  \label{eq:randomizedt} \end{equation}
where $\randmix \in [0,1]$ denotes the randomization probability and $\optpolicy_1,\optpolicy_2$ are pure (non-randomized) policies.
Of course,
when there is no constraint   (\ref{costconstraint}),  the CMDP reduces to classical MDP and the optimal stationary  policy $\optpolicy(\state)$ is  a pure policy.  That is, for each state $\state \in \statespace$, there exists an action $\action$ such that $\cond(\action|\state)= 1$.

{\em Remarks}. (i) (\ref{costconstraint}) is a global constraint that applies to the entire sample path \citep{Alt99}. Since the optimal policy is randomized, classical value iteration based approaches and  Q-learning cannot be used to solve CMDPs as they yield deterministic policies. One can construct a Lagrangian dynamic programming formulation \citep{Alt99} and Lagrangian Q-learning algorithms \citep{DK07}. Below for brevity, we consider a  policy gradient RL algorithm.

\blue{(ii) {\em Discounted CMDPs}. Instead of an average cost CMDP,  a discounted cost CMDP can be considered. Discounted CMDPs are  less technical in the sense that an optimal policy always exists (providing  the constraint set is non-empty); whereas average cost CMDPs require a unichain assumption. It is easily shown \citep{Kri16}  that the dual linear program of a discounted CMDP can be expressed in terms of the conditional probabilities
  $\cond(\action|\state)$ and the optimal randomized policy is of the form
  (\ref{eq:randomizedt}).
  The final IRL algorithm is identical to
\eqref{eq:irl_mdp} below.}

 \subsubsection{Policy Gradient for RL of  CMDP}
Having specified the CMDP model, we next turn to the RL algorithm.
RL algorithms\footnote{In adaptive control, RL algorithms such as policy gradient are   viewed as simulation based  {\em implicit} adaptive control methods that bypass estimating the MDP parameters (transition probabilities) and  directly estimate the optimal policy.} are   used to estimate the optimal policy of an MDP    when the transition matrices are not known. Then the LP formulation in terms of joint probabilities $\statpi$ is not useful since the constraints depend on the transition  matrix.
In comparison, \textit{policy gradient RL algorithms} are stochastic gradient algorithms of the form (\ref{eq:rl}) that  operate on the  conditional action probabilities $ \cond(\action| \state)$ defined in  (\ref{eq:conditionalprob})  instead of the joint probabilities $\statpi(\state,\action)$. 

Note that
 (\ref{eq:objective1}) written as a minimization (in terms of $-J$), together with constraint (\ref{costconstraint}) is in general, no longer a convex optimization problem in the variables $\cond$;
 see Figure \ref{fig:mdp} for an illustration. So it is not possible  to guarantee that simple gradient descent schemes\footnote{Consider minimizing the negative of the objective function, namely  $-J$ without constraint (\ref{costconstraint}). Even though $-J$  is nonconvex in $\cond$, one can show (using Lyapunov function arguments)  that for this unconstrained MDP case, the gradient algorithm  will converge to a global optimum. However for the constrained MDP case this is not true; the nonconvex  objective and constraints  results in a duality gap.} can achieve the global optimal policy. This motivates  the setting of (\ref{eq:rl}) where multiple agents that are initialized randomly aim to estimate the optimal policy.

Since the problem is non-convex, and the inequality constraint is active (i.e.,\ achieves equality) at the global maximum, we assume that the RL agents use a quadratic penalty method:
For $\lambda \geq 0$, denote the quadratic penalized objective to be maximized  as
\beq
\Reward(\cond) = J(\cond) - \lambda \, (\Cons^2(\cond) - \rcon)
\label{eq:penalty_reward}
 \eeq
 Such quadratic penalty functions are used widely for equality constrained non-convex problems.

The RL agents aim to minimize the $\finaltime$-horizon sample path penalized objective which at batch $k$ is
\beq
\begin{split}
  \reward_k (\cond)  &\defn  J_{k,\horizon}(\cond) + \lambda \,\Big(\Cons^2_{k,\horizon}(\cond)-\rcon\Big),  \quad \lambda \in \reals_+\\
  J_{k,\horizon} &=  \frac{1}{\finaltime} \sum_{n=1}^\finaltime
  \mreward(\state_n, \policy_\cond(\action_n)) , \quad
  \Cons_{k,\horizon} = \frac{1}{\finaltime} \sum_{n=1}^\finaltime
  \con(\state_n, \policy_\cond(\action_n))
\end{split} \label{eq:penalized}
\eeq
There are several methods for estimating the  policy gradient $\nabla_\cond \reward_\dtime(\cond_\dtime)$  \citep{Pfl96} including the score function method, weak derivatives \citep{VK03} and finite difference methods. A useful finite difference gradient estimate is given by the SPSA algorithm \citep{Spa03}; useful because SPSA evaluates the gradient along a single random direction.

\subsubsection{IRL for CMDP}
Consider the CMDP (\ref{Aij}), (\ref{J}), (\ref{eq:randpol}). Assume we are given a sequence of  gradient estimates $\{\nabla_\cond \reward_\dtime(\cond_\dtime)\}$ of the sample path wrt to the parametrized policy  $\cond$ from (\ref{eq:penalized}).  The aim of the inverse learner is to reconstruct the reward $\Reward(\cond) $ in  (\ref{eq:penalty_reward}). Since by construction the constraint is active at the optimal policy, the aim of the inverse learner is to explore regions of $\cond$ in the vicinity where the constraint $\{\cond: \Cons(\cond) \approx \rcon\}$ is active in order  to estimate  $\Reward(\cond) $.


A naive application of Langevin IRL algorithm (\ref{eq:irl}) to update the conditional probabilities $\{\cond_\dtime\}$  will not work. This is because there is no guarantee that the  estimate sequence $\{\cond_\dtime\}$ generated by the algorithm are valid  probability vectors, namely
\beq \cond_\dtime(\action|\state) \in [0,1], \quad \sum_{\action\in \actionspace} \cond_\dtime(\action|\state) = 1, \quad x \in \statespace.
\label{eq:simplex}
\eeq
We will use spherical coordinates\footnote{Another parametrization  widely used in machine learning is  exponential coordinates: $ \cond(\action|\state)  =  \frac{\exp(\param({\state, \action}))}{\sum_{a\in \actionspace} \exp(\param({\state, a}))}$,  where $ \param({\state,\action}) \in \reals$ is unconstrained. However, as shown in \citet{Kri16,KV18}, spherical coordinates typically yield faster convergence. We also found this in numerical studies on IRL (not presented here).}
to ensure that  the conditional probability estimates $\cond_\dtime$  generated by the IRL algorithm satisfies  (\ref{eq:simplex}) at each iteration $\dtime$. The idea is to parametrize $\sqrt{\cond_\dtime(\action|\state)}$ to lie on the unit hyper-sphere in $\reals^\actiondim$.  Then all needed are the $\actiondim-1$ angles for each $\state$, denoted as $\param(i,1),\ldots \param(i,\actiondim-1)$.
Define the spherical coordinates in terms of the mapping:
\begin{equation} \cond = \logistic(\param), \quad \text{ where }
      \cond(\action|\state)  =  \begin{cases} \cos^2 \param(i,1)  & \text{ if } \action =1  \\
        \cos^2 \param(i,u) \prod_{p=1}^{u-1} \sin^2 \param(i,p)  & u \in \{2,\ldots, \actiondim-1\}           \\
        \sin^2 \param(i,\actiondim-1) \prod_{p=1}^{\actiondim-2} \sin^2\param(i,p) & u = \actiondim
                                                                        \end{cases}
                                                                        \label{eq:expo}
\end{equation}
Then clearly $\cond({\action|\state})$ in (\ref{eq:expo})  always satisfies feasibility
(\ref{eq:simplex})
for any real-valued (un-constrained)  $\param({\state,\action})$.
To summarize,  there are  $(\actiondim-1) \statedim$ unconstrained parameters in $\param$.
Also for $\param(i,u) \in [0,\pi/2]$,
the mapping $\logistic: \reals^{\actiondim\times \statedim}\rightarrow \reals^{\actiondim\times \statedim}$ in (\ref{eq:expo}) is  one-to-one and therefore invertible. We denote the inverse as $\logistic^{-1}$.

{\em Remark}: As an example, consider $\actiondim=2$. Then  in spherical coordinates $\cond(1|i) = \sin^2\param(i,1)$, $\cond(2|i) = \cos^2\param(i,1)$, where $\param(i,1)$ is un-constrained.; clearly
$\cond(1|i) + \cond(2|i) = 1$, $\cond(u|i) \geq 0$.

With the above re-parametrization,
we can run any of the  passive Langevin dynamics IRL algorithms proposed in this paper.
In the numerical example below,
we ran the two-time scale multi-kernel  IRL algorithm (\ref{eq:mcmcirl}). Recall this does not require knowledge of $\belief(\cdot)$ and also provides variances reduction:
Given the current IRL estimate $\eth_\dtime$, the RL gives us  a sequence
  $\{\cond_i, \nabla_\cond \reward_{\dtime}(\cond_i), i=1,\ldots,L\}$
 The IRL algorithm (\ref{eq:mcmcirl})  operating on the $(\actiondim-1) \statedim$ unconstrained parameters of $\param$ is:
\beq
\begin{split}
 \eth_{\dtime+1} &= \eth_\dtime + \stepa \,\frac{\temperature}{2} \, \frac{\sum_{i=1}^\numparticles   \pdf(\th_i|\eth_\dtime) \hnabla_\th \reward_\dtime(\th_i) }{\sum_{l=1}^\numparticles \pdf(\th_l|\eth_\dtime)}  
  + \sqrt{\stepa}  \noise_\dtime,\qquad \cond_i \sim \belief(\cdot)  \\
    \text{ where } &\quad \th_i = \logistic^{-1}(\cond_i), \quad \nabla_\th \reward_\dtime(\th_i) =
( \nabla_\cond \reward_{\dtime}(\cond_i) )^\p \, \nabla_\th \cond_i, \\
&\pdf(\th|\eth) = \pdf_\obsnoise(\th - \eth), \qquad  \pdf_\obsnoise(\cdot) = \normal(0,\sigma^2I_\thdim) 
\end{split} \label{eq:irl_mdp}
     \eeq
 In the
 second line of (\ref{eq:irl_mdp}), we transformed  $\nabla_\cond \reward_{\dtime}(\cond_\dtime) $ to
$ \nabla_\th \reward_\dtime(\th_\dtime)$  to use in the  IRL algorithm.

To summarize,  the IRL algorithm (\ref{eq:irl_mdp}) generates samples  $\eth_\dtime \sim \exp(\Reward(\logistic(\eth)))$.
Equivalently, $\cond_\dtime = \logistic(\eth_\dtime) \sim \exp(\Reward(\cond))$, where $\Reward(\cond)$ is defined in (\ref{eq:penalty_reward}). Thus given only gradient information from a RL algorithm,  we can reconstruct (sample from) the penalized reward $\Reward(\cdot)$ of the  CMDP without any knowledge of the CMDP parameters.

\subsubsection{Numerical Example}
We generated a CMDP with $\statedim=2$ (2 states),  $\actiondim=2$ (2 actions) and 1 constraint with
\beq
\tp(1) = \begin{bmatrix} 0.8 & 0.2 \\ 0.3 & 0.7
\end{bmatrix}, \quad \tp(2) =  \begin{bmatrix} 0.6 & 0.4 \\ 0.1 & 0.9
\end{bmatrix},  \mreward = \begin{bmatrix} 1 & 100 \\ 30 & 2
\end{bmatrix},  \con = \begin{bmatrix} 0.2 & 0.3 \\ 2 & 1
\end{bmatrix}, \rcon = 1, \lambda = 10^5
\eeq
Recall the transition matrices $\tp(\action)$ are defined in (\ref{Aij}),  the reward matrix $(\mreward(\state,\action))$ in  (\ref{J}),  constraint matrix $(\con(\state,\action))$ and $\rcon$ in (\ref{costconstraint}),  and penalty multiplier  $\lambda$  in (\ref{eq:penalized}).

The  randomized policy $\cond(\action|\state)$, $\action\in \{1,2\}$, $\state \in \{1,2\}$  is a $2\times 2$ matrix. It is completely determined  by $(\cond(1|1), \cond(1|2)) \in [0,1]\times [0,1]$; so it suffices to  estimate $\Reward(\cond)$ over $[0,1]\times[0,1]$.

Figure \ref{fig:mdp}(a) displays the cumulative reward $J(\cond)$; this constitutes the ground truth.
To obtain this figure, we computed  the average reward MDP value function $J(\cond)$ and constraint $\Cons(\cond)$ for each policy $\cond$ where $\cond$ sweeps over $[0,1]\times[0,1]$.
Given a policy $\cond$, $J(\cond)$ and $\Cons(\cond)$ are  computed by first evaluating the joint probability $\statpi$ as \citep[pp.101]{Ros83}
$$ \statpi(j,a) = \sum_i \sum_{\bar{a}} \statpi(i,\bar{a})\, \tp_{ij}(\bar{a})\, \cond(a|j), \quad
\sum_j \sum_a \statpi(j,a)= 1 $$
and then
$ J(\cond)  = \sum_\state\sum_ \action \statpi(\state,\action) \mreward(\state,\action)$,
$\Cons(\cond) =  \sum_\state \sum_\action \statpi(\state,\action) \con(\state,\action) $.

For values of $\cond$ that do not satisfy the constraint $\Cons(\cond) < \rcon$, we plot $J(\cond) = 0$.
Figure \ref{fig:mdp}(a) illustrates the non-convex nature of the constraint set.

Figure \ref{fig:mdp}(b) displays the penalized cumulative reward $\Reward(\cond) = J(\cond) - \lambda\, (\Cons(\cond)-\rcon)^2$ where the quadratic penalty function is $\lambda\, (\Cons(\cond)-\rcon)^2$. As mentioned earlier, since we know that the constraint is active at the optimal policy, we want the IRL  to explore the vicinity of the region of $\cond$ where the constraint is active.

We then ran the  IRL algorithm (\ref{eq:irl_mdp})
 using spherical coordinates with parameters  $\stepa = 5 \times 10^{-6}$, $\sigma =0.1$, $L=50$ for
 $\horizon = 10^5$ iterations.
Figure \ref{fig:mdp}(c)   displays a 3-dimensional stem plots of the log of the  empirical distribution of $\cond_k = \logistic(\eth_k)$.
 wrt coordinates $\cond(1|1)$ and $\cond(1|2)$.  As can be seen from the two plots, the IRL algorithm samples from the high probability regions $\{\cond: \Cons(\cond) \approx \rcon\}$ to reconstruct the penalized reward $\Reward(\cond)$. Specifically, the $C$-shaped curve profile generated by the IRL estimates match the $C$-shaped curve of the penalized cumulative reward Figure \ref{fig:mdp}(b).

\begin{figure}[p] \centering
  \begin{subfigure}{.6\textwidth}
\includegraphics[scale=0.45]{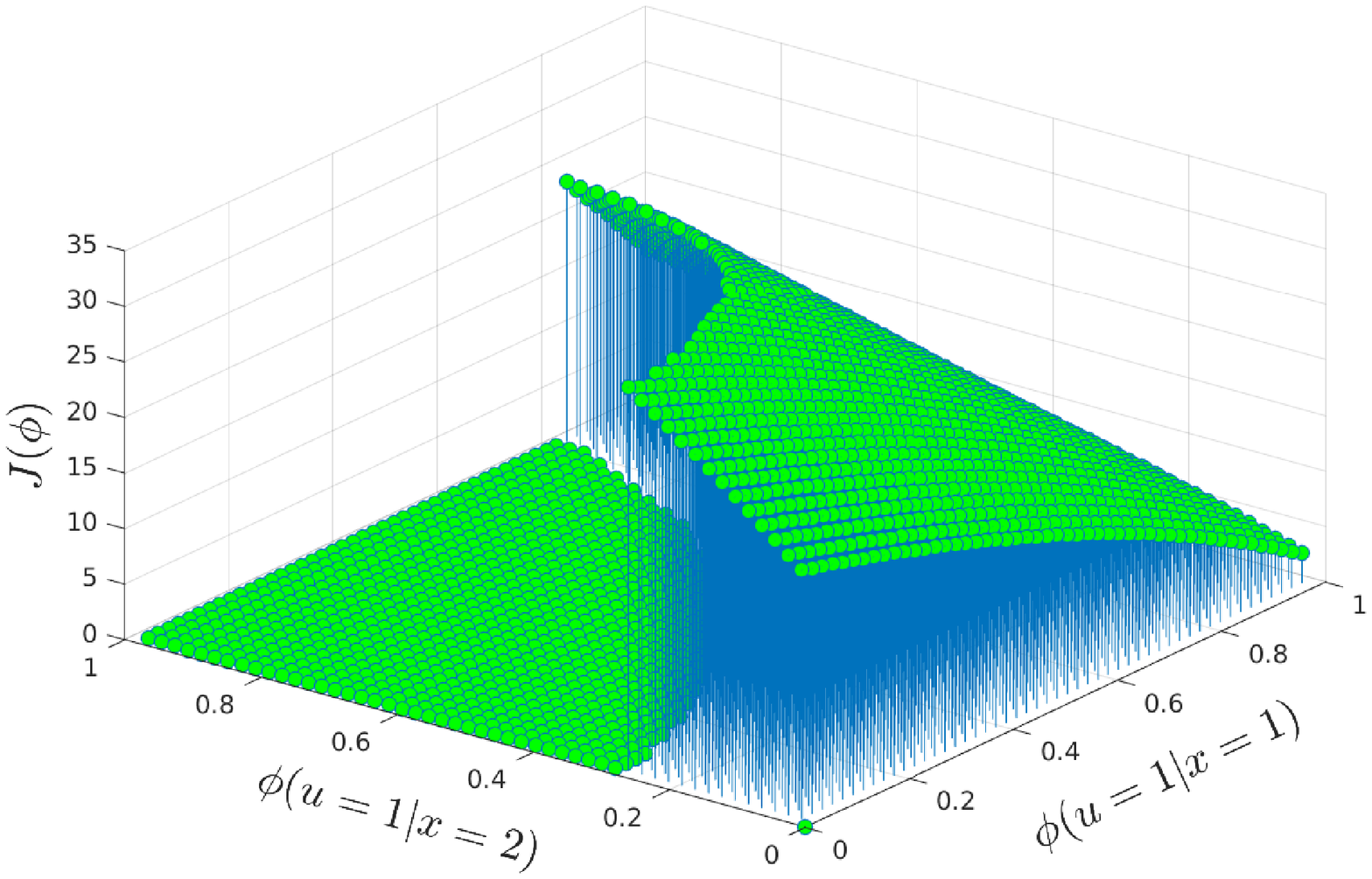}
\caption{Cumulative Reward $J(\cond)$ with active constraint $\Cons(\cond)\leq 1$. The non-convexity of the constraint set is clearly seen. }
\end{subfigure} \\
\begin{subfigure}{.6\textwidth}
\includegraphics[scale=0.45]{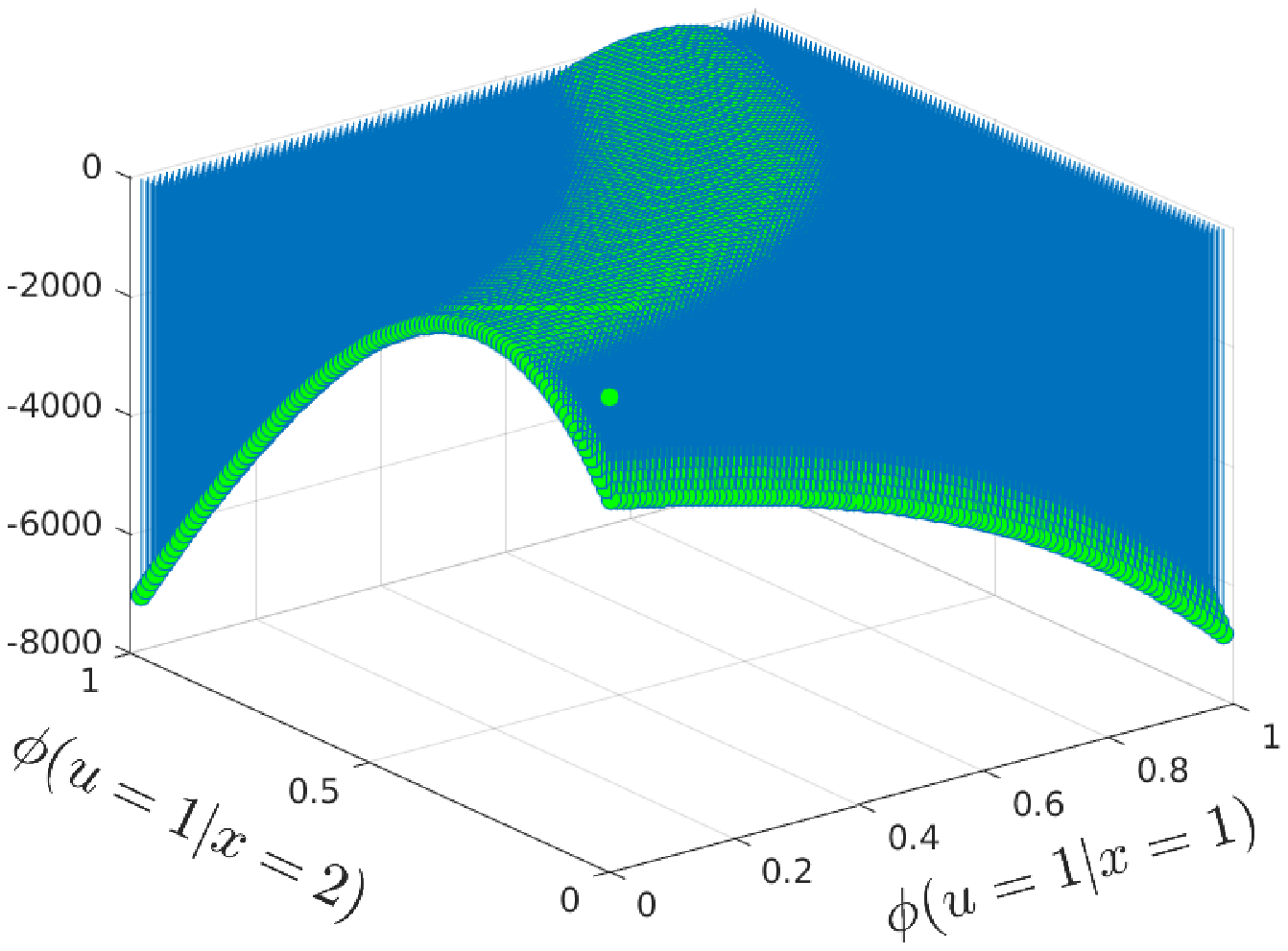}
\caption{Penalized Cumulative Reward with Quadratic Penalty  $\Reward(\cond) = J(\cond) - \lambda\, (\Cons(\cond)-1)^2$. The lighter green shade  on top shows the active constraint. This plot constitutes the ground truth}
\end{subfigure}\\
\begin{subfigure}{\textwidth}
\includegraphics[scale=0.4]{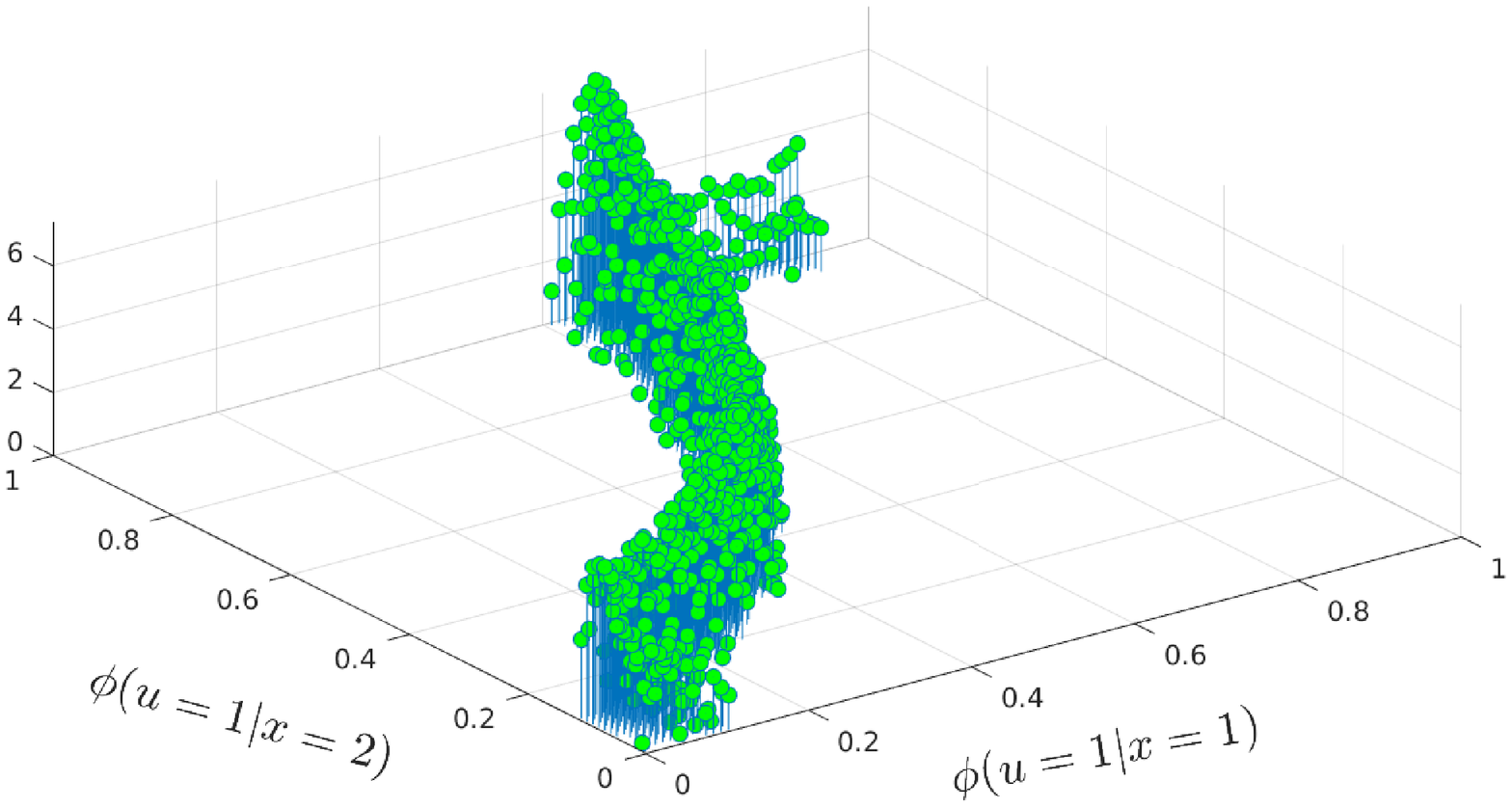}
  \includegraphics[scale=0.4]{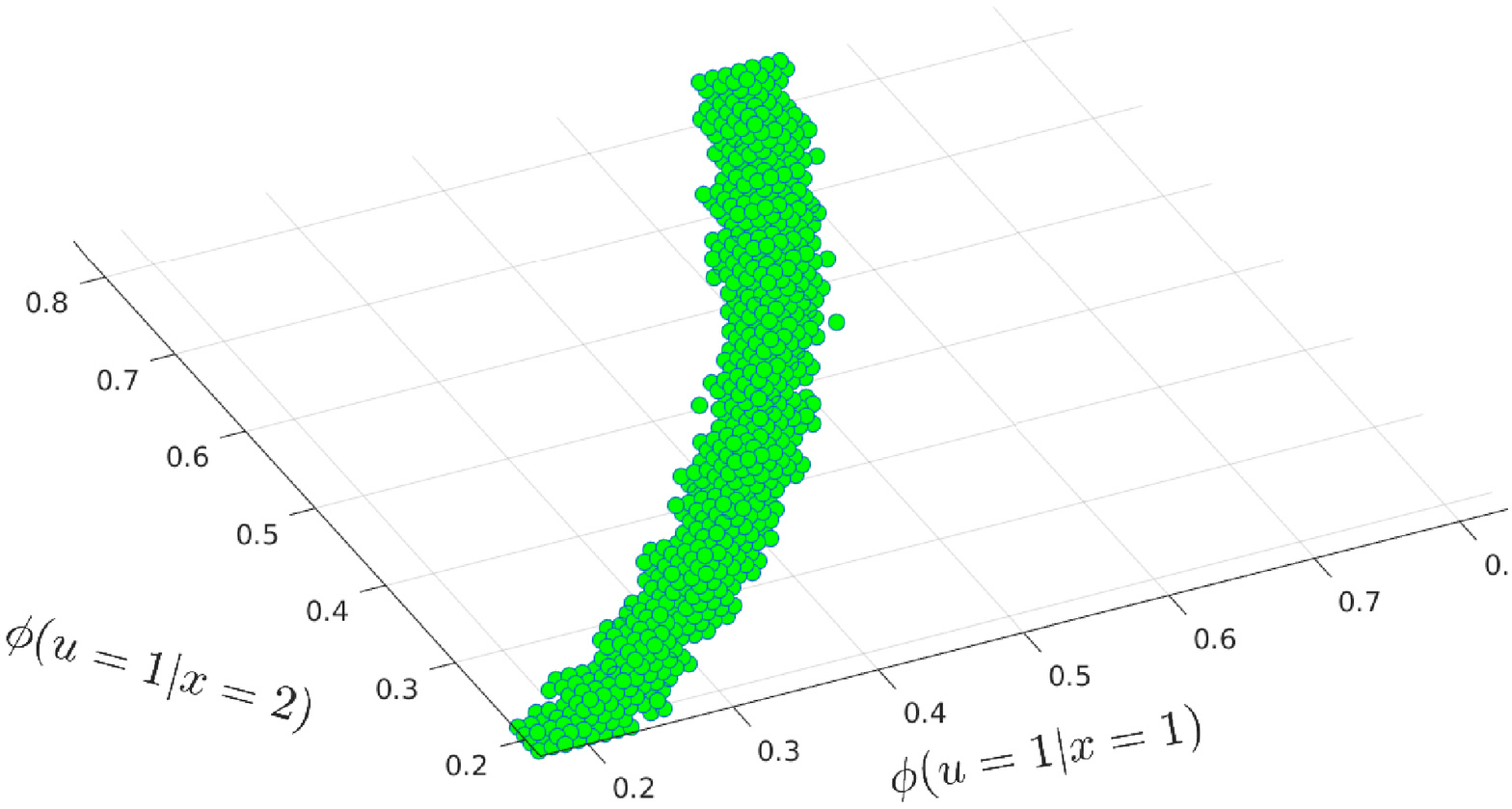}
  \caption{IRL algorithm estimate. Snapshot 1 shows that the IRL estimates $\Reward(\cond)$   in the vicinity of the active constraint..
Snapshot 2  shows that the IRL explores regions  in the vicinity of the active constraint. Specifically the curve is close to the lighter shade green in Fig (b)}
\end{subfigure}
\caption{IRL for Constrained MDP}
\label{fig:mdp}
\end{figure}


\section{Weak Convergence Analysis of IRL Algorithm}
\label{sec:weak}

This section discusses the main assumptions, weak convergence theorem and proof regarding IRL algorithm~(\ref{eq:irl}). (Recall the informal proof in Sec.\ref{sec:informal} for the motivation of weak convergence.)

\subsubsection*{Notation}
\begin{compactitem}
\item Since  $\nabla_\th \reward_k(\th_k)$ is a noise corrupted estimate of the gradient $\nabla_\th \Reward(\th)$, we write it in more explicit notation as
$\wdt \reward (\th_k,\xi_k)$, where $\{\xi_k\}$ is a sequence of random variables satisfying appropriate conditions specified below.
\item We  use $\belief_\al(\cdot)$ to denote $\nabla_\al \belief(\cdot)$. 
\item Finally,  $\E_m$  denotes the conditional expectation (conditioning up to time $m$), i.e.,
conditioning wrt the $\sigma$-algebra ${\mathcal F}_m=\sigma\{ \al_0,\th_j,\xi_j; \
j<m\} $.
\end{compactitem}

\subsubsection*{Algorithm}
There are two possible implementations of  IRL algorithm (\ref{eq:irl}).
The first implementation is (\ref{eq:irl}), namely,
\beql{eq:irl-1}\al_{k+1}=\al_k + {\stepa \over \kernelstep^\thdim} K\( {\th_k-\al_k \over \kernelstep} \) {\temperature \over 2} \wdt \reward(\th_k,\xi_k)\belief(\al_k)+ \stepa \belief_\al(\al_k) \belief(\al_k) +\sqrt \stepa \belief(\al_k) \noise_k,
\eeq
and the second implementation is
\beql{eq:irl-2}\al_{k+1}=\al_k + {\stepa \over \kernelstep^\thdim}  K\( {\th_k-\al_k \over \kernelstep} \)\Big[ {\temperature \over 2} \wdt \reward(\th_k,\xi_k)\belief(\th_k) + \belief_\al(\th_k)\Big]  +\sqrt {\stepa \over \kernelstep^\thdim}  K\( {\th_k-\al_k \over \kernelstep} \)\belief(\th_k) \noise_k,
\eeq
where $\stepa$ is the stepsize and $\kernelstep=\kernelstep(\stepa)$ is chosen so
$\stepa /\kernelstep^\thdim\to 0$ as $\stepa\to 0$.

Both the above  algorithms converge  to the same limit. The proof below is devoted to \eqref{eq:irl-1}, but
\eqref{eq:irl-2} can be handled similarly.
Also the proofs of the other two proposed IRL algorithms, namely   (\ref{eq:irl2}) and  (\ref{eq:irl3}) are similar.

Taking a continuous-time interpolation
\beq \al^\stepa(t)= \al_k
\ \hbox{ for } \ t \in [\stepa k, \stepa k+ \stepa),  \label{eq:interpolated}\eeq
we aim to show that the sequence $\al^\stepa(\cdot)$ converges weakly to $\al(\cdot)$, which give the desired limit.

\subsection{Assumptions}
We begin by stating the conditions needed.

\begin{enumerate}[label=(A{\arabic*})]

\item \label{H1} For each $\xi$, $\wdt \reward(\cdot,\xi)$ has continuous partial
derivatives up to the second order such that the second partial
$\wdt \reward_{\al\al}(\cdot, \xi)$ is bounded.
For each
$b<\infty$ and $T<\infty$,
$\{\wdt r(\al,\xi_j); |\al|\le b, j\stepa \le T\}$
is
uniformly integrable.

\item \label{H2}  The sequences $\{ \th_k\} $ is stationary and  independent of $\{\xi_k\}$.
For each
$k\ge n$, there exists a conditional density of $\th_k$
given ${\mathcal F}_n$, denoted by $\belief_k(\th| {\mathcal F}_n)$ such that
$\belief_k(\th|{\mathcal F}_n)>0$ for each $\th$ and that
$\belief_k(\cdot |{\mathcal F}_n)$ is continuous. The sequence
$\{\belief_k(\cdot|{\cal F}_n)\}_{k\ge n}$ is bounded uniformly.
The probability
density $\belief(\cdot)$ is continuous and bounded with
$\belief(\th)>0$ for each $\th$ such that
\beql{pi-ave}\lim_{k-n\to \infty}\E | \belief_k(\th|{\mathcal F}_n) - \belief(\th) | =
0 .\eeq

\item \label{H3}
The measurement noise $\{\xi_n\}$ is
exogenous, and bounded stationary mixing process with mixing measure $\ph_k$ such that
$\E \wdt r(\al,\xi_k)= R_\al(\al)$ for each $\al$ and $\sum_{k} \ph_k<\infty$. The $\{\noise_k\}$ is a sequence of $\reals^\thdim$-valued i.i.d. random variables with mean $0$ and covariance matrix $I$ (the identity matrix); $\{w_k\}$ and $\{\xi_k\}$ are independent.

\item \label{H4}
 The kernel $\kernel(\cdot)$ satisfies
 \beql{ker}\barray
 \ad K(u)\ge 0, \ K(u)=K(-u), \sup_u K(u) < \infty,\\
 \ad \int K(u) du =1, \  \int |u|^2 K(u)du < \infty.
 \earray\eeq

\end{enumerate}

{\em Remarks}:
We briefly  comment on the assumptions \ref{H1}-\ref{H4}.
\begin{compactitem}
\item Assumption \ref{H1} requires the smoothness of $\wdt r(\cdot, \xi)$, which is natural because we are using $\wdt r(\cdot,\xi_k)$ to approximate the smooth function $\nabla R$. We consider a general noise so the uniform integrability is used. If
    the noise is additive in that $\wdt r(\th,\xi)= \nabla R(\th) + \xi$, then we only need the finite $\wdt p$-moments of $\xi_k$ for $\wdt p>1$.
\item Assumption \ref{H3} requires the stochastic process $\{\xi_n\}$ to be
exogenous, and bounded stationary mixing. Thus
for each $\al$,  $\{\wdt r(\al,\xi_k)\}$ is also a mixing sequence. A mixing process is one in which remote past and distant future are asymptotically independent. It covers a wide range of random processes such as i.i.d. sequences, martingale difference sequences,  moving average sequences driving by a martingale difference sequence, and functions of stationary Markov processes with a finite state space \citep{Bil99}, etc. The case of $\{w_k\}$ and $\{\xi_k\}$
being dependent can be handled, but for us $\{w_k\}$ is the added  perturbation to get the desired Brownian motion so independence is sufficient.

\item By exogenous in \ref{H3}, we mean that
\bea \ad P(\xi_{n+1}\in A_1,\ldots, \xi_{n+k}\in
A_k|\al_0,\xi_j, x_j; \
j\le n)
\\
\aad \  =P(\xi_{n+1}\in A_1,\ldots,\xi_{n+k}\in A_k|\al_0, x_j,
\xi_j,\al _{j+1}; \
j\le n),\eea
for all Borel sets $A_i$, $i\le k$, and for all $k$ and $n$.

\item In view of the mixing condition \ref{H3} on $\{\xi_k\}$,
for each
$b<\infty$ and $T<\infty$,
$\{\wdt r(\al,\xi_j); |
\al|\le b, j\stepa \le T\}$ and $\{\wdt r_\al(\al,\xi_j); |\al|\le b,
j \stepa\le T\}$ are
uniformly integrable.

\item
Again, using the mixing condition,
for each $\al$, as $n\to \infty$,
\beql{ave}{1\over {n}}\sum^{m+n-1}_{j=m} \E_m \wdt r(\al,\xi_j) \to R_\al(\al)
\hbox{ in probability.}\eeq

\item For a Borel set $A$,  we have
$P(\th_k\in A| {\cal F}_n)=\int_{\th\in A} \belief_k(\th|{\cal F}_n) d\th$.
If $\{\th_n\}$ is itself a stationary $\phi$-mixing sequence with a
continuous density, and if $\E|\th_n|^2<\infty$,
then by virtue of a well-known mixing inequality, some $\wdt c_0>0$,
\cite[Corollary 2.4 in Chapter~7]{EK86},
$$\E \bigl\{ | \int  \th \belief_k(\th|{\cal F}_n)d\th -\int \th \belief( \th) d\th |\bigr\} \le
\wdt c_0
\ph^{1/2}_\th(k-n)\E^{1/2}|\th_k|^2
\to 0 \hbox{   as  }k-n\to \infty,$$
where $\ph_\th(\cdot)$ denotes the
mixing measure.

\item \textcolor{black}{Condition \ref{H4} is
  concerned with the properties of $K(\cdot)$.
 It assumes  that the kernel is  nonnegative, symmetric, bounded (similar to a probability density function),
 and square integrable. \ref{H4} is satisfied by a large class of kernels.
 For example, commonly used symmetric kernels with compact supports
 satisfy this condition (e.g., truncated Gaussian kernels). Moreover, it is also verifiable for kernels with {\em unbounded} support.
 A crucial point is  that the tails of $K(\cdot)$ are small (asymptotically negligible).
  For simplicity, we use \ref{H4} as a nicely packaged version.
   In fact,  \ref{H4} is a sufficient condition
for a much larger class of
kernels satisfying
\beql{ker-1}\barray
\ad \int K(u) du =1, \ \int |u|^l K(u) du < \infty \ \hbox{ some } l,\\
\ad \int K^2 (u)du <\infty, \ \int |u|^2 K(u) du < \infty, \\
\ad \int (u^1)^{m_1} (u^2)^{m_2} \cdots (u^N)^{m_N} K(u) du =0\ \hbox{ if } \ l > 1,
\earray\eeq
where $1 \le m_1+ m_2 + \cdots + m_N \le l-1$, where $u^1,\dots,u^N$ denote the components of $u\in {\mathbb R}^N$. The parameter $l$ is a smoothness indicator of the kernel and the last line of \eqref{ker-1} is often used in  nonparametric estimation in statistics.
Such a condition stems from a large class of kernels used
in the so-called
 $l$th-order averaging operator; see \cite{Kat76}. Thus, \ref{H4} can be replaced by this more general setup. However,
 we use the current form of \ref{H4} because it is easily verifiable (e.g., by Gaussian kernel).}
\end{compactitem}

\subsection{Main Result and Proof} \label{sec:proofmain}

As is well known
 \citep{KY03}, a classical fixed step size  stochastic gradient  algorithm
converges weakly  to a \textit{deterministic}  ordinary differential equation (ODE) limit; this is the basis of the so called ODE approach  for analyzing stochastic gradient algorithms.  In comparison, the discrete time IRL algorithm (\ref{eq:irl})  converges weakly to a \textit{stochastic} process limit $\al(\cdot)$.  In this section we prove weak convergence of the interpolated process $\{ \al^\stepa(\cdot)\}$ to the stochastic process limit $\al(\cdot)$ as $\stepa\rightarrow 0$.
Proving weak convergence
requires first that  the tightness of the sequence be verified and
then the limit be characterized via the so called
martingale problem formulation.  For a comprehensive treatment of the martingale problem of Stroock and Varadhan,  see \cite{EK86}.

\begin{theorem}\label{thm:weak-conv}
  Assume conditions {\rm \ref{H1}-\ref{H4}}. Then the interpolated process $\al^\stepa(\cdot)$ $($defined in \eqref{eq:interpolated}$)$ for IRL algorithm \eqref{eq:irl}  has the following properties:
  \begin{compactenum} \item
$\{ \al^\stepa(\cdot)\} $
is tight in $D^d[0,\infty)$.
\item Any weakly convergent subsequence
of $\{\al^\stepa(\cdot)\}$  has a
limit $\al(\cdot)$ that
satisfies
\beq \begin{split} d\al(t) &= \Big[ {\temperature \over 2} \belief^2(\al(t)) \Reward_\al(\al(t))+ \belief_\al(\al(t)) \belief(\al(t))\Big] dt+ \belief(\al(t) ) d\bm(t), \\
 \al(0) &=\al_0,
\end{split}
\label{eq:sde} \eeq
where $\bm(\cdot)$ is a standard Brownian motion with mean 0 and covariance being the identity matrix $I \in \reals^{\thdim\times \thdim}$,
provided~\eqref{eq:sde} has a
unique weak solution $($in a distributional sense$)$ for each initial condition.
\end{compactenum}
\end{theorem}

For sufficient conditions leading to  unique weak solutions of  stochastic differential equation and uniqueness of martingale problem, see \citet[p. 182]{EK86} or
\cite{KS91}.

{\bf Proof.}
The proof is divided into 4  steps.

{\underbar{Step 1. Use a truncation device.}}
Because the sequence $\{\eth_\dtime\}$  is not {\it a priori}  bounded, the main idea is to
 use a truncation device \cite[p.284]{KY03}. (Step 4 below deals with the un-truncated process.)
Let $M>0$ be a fixed but otherwise arbitrary constant.
Denote by $S_M=\{\al \in \reals^\thdim: |\al| \le M\}$ the $\thdim$-dimensional ball centered at the origin with radius $M$.
Consider the truncated algorithm
\beql{eq:irl-t}\al^M_{k+1}\!=\!\al^M_k + \stepa  \Big[{1\over \kernelstep^\thdim} \kernel\( {\th_k-\al^M_k \over \kernelstep} \) {\temperature \over 2} \wdt  \reward(\th_k,\xi_k)+ \belief_\al(\al^M_k)\Big]\belief(\al^M_k)  q_M(\al^M_k) +\sqrt \stepa \belief(\al^M_k)q_M(\al^M_k) \noise_k ,
\eeq
where $$q_M(\al)= \Bigg\{\barray 1, \   \al\in S_M;\\
0, \ \al \in \reals^\thdim-S_{M+1};\\
\text{smooth }  \ \text{ otherwise.}\earray$$
\textcolor{black}{By virtue of \ref{H4},
the integrability of the kernel
forces $\th_k$ to be in line with the iterates $\alpha^M_k$ in that only  asymptotically negligible tails can be added.}

{\em Remark}.
 Define $\al^{\stepa,M}(t)=\al^M_n \hbox{ on } [\stepa k,\stepa k+\stepa).$ Then
$\al^{\stepa,M}(\cdot)\in D^\thdim[0,\infty)$ and is an $M$-truncation for
$\al^\stepa(\cdot)$ \citep[p.284]{KY03}.  We proceed to prove the tightness and
weak convergence of the truncated sequence $\{\al^{\stepa,M}(\cdot)\}$ first
and then complete the proof by letting $M\to \infty$ in Step 4.

{\underbar{Step 2.
Prove the tightness of $\{\al^{\stepa,M}(\cdot)\}$.}}
Note that in view of \citet[Lemma 1]{NPT89}, by virtue of \ref{H4},
for a function $h\cd$ that is twice continuously differentiable with bounded second derivative, it follows that
\beql{eq:k-est} \Big| {1\over \kernelstep^\thdim} \int \kernel\Big( {\th -\al \over \kernelstep} \Big) h(\th) d\th - h(\al) \Big| = O(\kernelstep^2).\eeq
Using \eqref{eq:k-est}, \ref{H1}, and noting that $\{\noise_k\}$ is an i.i.d. sequence with mean $0$ and covariance matrix $I$, we can show that
 $\Big\{
 \big[\kernelstep^{-\thdim} \kernel\( {\th_k-\al^M_k \over \kernelstep} \) {\temperature \over 2} \wdt  r(\th_k,\xi_k)+\belief_\al(\al^M_k)\big]\belief(\al^M_k)  q_M(\al^M_k)\Big\}$ is uniformly integrable and also $\{\belief(\al^M_k)q_M(\al^M_k) \noise_k\}$ is uniformly integrable. Then using \citet[p.51, Lemma 7]{Kus84} (or use a perturbed test function methods as in \citet[Chapter 7]{KY03}), it can be shown that
 $\{\al^{\stepa,M}\cd\}$ is tight in $D([0,\infty), \reals^\thdim)$, the space of $\reals^\thdim$-valued functions that are right continuous, have left limits, endowed with the Skorohod topology.

{\underbar{Step 3. Characterize the limit process.}}
 Because $\{\al^{\stepa,M}\cd\}$ is tight, by virtue of Prohorov's theorem \citep{Bil99}, we can extract a weakly convergent subsequence. To simplify notation, still denote the subsequence by $\{\al^{\stepa,M}\cd\}$ whose limit is $\al^M\cd$. By Skorohod representation  \cite[p. 230]{KY03} with a slight abuse of notation, we may assume that $\al^{\stepa,M}\cd$ converges  to $\al^M\cd$ w.p.1.
To complete the proof, we need only  characterize the limit process by showing that the limit
$\al^M\cd$ is a solution of the martingale problem with backward operator
\beql{M-op}\op^M f(\alpha) = f_\al'(\al) \Big[ {\temperature \over 2} \belief^2(\al)R_\al(\al) + \belief_\al(\al) \belief(\al)\Big] q_M(\al)
+ {1\over 2} \belief^2(\al) \Tr[ f_{\al\al}(\al)] q_M(\al)\eeq
for any real-valued function $f\cd \in C^2_0$ (\citet[Lemma 8.1, p.225]{EK86}), where $f'$ denotes the transpose of $f$.

By Theorem 8.2 in \citet{EK86},
to verify the martingale property, we need to show
that for any bounded and continuous test function $g\cd$, any $t,s >0$, any positive integer $\kappa_1$, and any $t_\mi \le t$,
\beql{mg-M}\barray \ad \E \Big\{g(\al^M(t_\mi): \mi\le \kappa_1)  \Big[ f(\al^M(t+s))-f( \al^M(t))-
\int^{t+s}_ t \op^M f(\al^M(u)) du \Big]\Big\}=0 . \earray\eeq
Note that (\ref{mg-M}), namely, the solution of the martingale problem,
is a statement about the finite dimensional   distributions of $\eth^M(\cdot)$
at times $t_1,\ldots,t_{\kappa_1}$.

To verify \eqref{mg-M}, we work with the sequence indexed by $\stepa$.
By the continuity of $f\cd$, the weak convergence, and the Skorohod representation, we have
that as $\stepa\to 0$,
\beql{es1}\barray
\ad  \E g(\al^\mm(t_\mi): \mi\le \kappa_1) [ f(\al^\mm(t+s))- f(\al^\mm(t))]\\
\aad \ \to \E  g(\al^M(t_\mi): \mi\le \kappa_1) [ f(\al^M(t+s))- f(\al^M(t))].\earray\eeq

To  simplify notation, we denote $q^M_k = q_M(\al^M_k)$ in what follows whenever there is no confusion and retain the notation $q_M(\al^M_k)$ whenever it is needed.
Dividing the segment $$\lf t/\stepa \rf \le k\le \lf (t+s) /\stepa\rf$$ into sub-blocks of size $m_\stepa$ each so that
$m_\stepa\to \infty$ as $\stepa\to 0$ and $\delta_\stepa= \stepa m_\stepa  \to 0$.
Then
we obtain
\beql{es2}\barray \ad\!\!\! \E \rk
 [ f(\al^\mm(t+s))- f(\al^\mm(t))] \\
\aad  = \E \rk
 \(\Em \sum^{(t+s)/ \delta_\stepa}_{l= t/\delta_\stepa}[ f(\al^M_{lm_\stepa +m_\stepa})- f(\al^M_{lm_\stepa})]\) \\
\aad  = \E
 \rk \Big\{\Em \suml f'_\al(\al^M_{lm_\stepa}) \sumk \Big[
 {\stepa \over \kernelstep^\thdim}  \kernel\( {\th_k-\al^M_k \over \kernelstep} \)\\
 \aad \hspace*{1.8in}\times {\temperature \over 2} \wdt r(\th_k,\xi_k)\belief(\al^M_k) + \stepa\belief_\al(\al^M_k) \belief(\al^M_k) +\sqrt \stepa \belief(\al^M_k) \noise_k
 \Big]\qmk\\
\aad\hspace*{1.8in}  + {1\over 2}\Em   \stepa\suml \sumk  \belief^2(\al_k)
\Tr[ f_{\al\al}(\al^M_{lm_\stepa})\noise_k\noise^\p_k]\qmk\\
\aad\hspace*{1.8in} +\Em \suml  e^\stepa_l
 \Big\},
\earray\eeq
where $\Em$ denotes the conditional expectation with respect to the past information up to the time $lm_\stepa$
(i.e., the $\sigma$-algebra generated by $\{ \al_{k},\th_k, \xi_k:  k< lm_\stepa\}$), and
$e^\stepa_l$ is an error term. It can be shown that
\beql{es2-e} \E\rk \Big| \Em\suml  e^\stepa_l \Big|^2\to 0 \ \hbox{ as  } \ \stepa\to 0.\eeq

Noting that $\{\noise_k\}$ is an i.i.d. sequence with mean 0 and covariance $I$ (the identity matrix), using the continuity of $\belief\cd$, the limit of
\bea \ad  \E\rk
\Big\{{1\over 2} \Em   \stepa\suml \sumk  \belief^2(\al^M_k)
\Tr[ f_{\al\al}(\al^M_{lm_\stepa})\noise_k\noise^\p_k]\qmk \Big\}, \eea
is the same as that of
\bea \ad  \E\rk \Big\{{1\over 2}
 \suml \belief^2(\al^M_{lm_\stepa}) \Tr [f_{\al\al}(\al^M_{lm_\stepa})]  \delta_\stepa q_M(\al_{lm_\stepa})\Big\}\\
 \aad \ =
  \E\rk \Big\{{1\over 2}
 \suml \belief^2(\al^{\stepa,M}(l\delta_\stepa)) \Tr [f_{\al\al}(\al^{\stepa,M}(l\delta_\stepa))]  \delta_\stepa q_M(\belief(\al^{\stepa,M}(l\delta_\stepa))) \Big\}
 . \eea It then follows from weak convergence of $\al^{\stepa,M}\cd$ to $\al^M\cd$ and the Skorohod representation,
\beql{es-iid-c}\barray
\ad  \E\rk
\Big\{{1\over 2} \Em   \stepa\suml \sumk  \belief^2(\al^M_k)
\Tr[ f_{\al\al}(\al^M_{lm_\stepa})\noise_k\noise^\p_k] \qmk\Big\}\\
\aad \ \to \E \rkm \Big\{{1\over 2} \int^{t+s}_t \belief^2 (\al^M(u)) \Tr[ f_{\al\al} (\al^M(u))] q_M(\al^M(u)) du  \Big\}\ \hbox{ as } \ \stepa\to 0.\earray\eeq

Using the condition on the i.i.d. noise $\{\noise_k\}$,
it is readily seen that
\beql{iid-0} \barray
\ad  \E\rk \Big\{ \Em \sqrt \stepa \suml f'_\al(\al^M_{lm_\stepa})\sumk  \belief(\al^M_k) \noise_k \qmk\Big\}
\\
\aad \ = \E\rk \Big\{ \sqrt \stepa \suml f'_\al(\al^M_{lm_\stepa})\sumk \Em \belief(\al^M_k)\Em \noise_k \qmk\Big\}\\
\aad \ \to 0 \ \hbox{ as } \ \stepa \to 0.\earray\eeq
Next, using the continuity of $\belief\cd$, $\belief_\al\cd$,
$f_\al\cd$,
together with the weak convergence of $\al^{\stepa,M}\cd$ to $\al^M\cd$, the Skorohod representation, the notation $\qmk$ defined before, and the notation convention $\qM= q_M(\al^M_{lm_\stepa})$ and $\qMl= q_M(\al^M(l \delta_\stepa))$,
we have
\beql{ave-pi}
\barray \ad \lim_{\stepa\to 0} \E \Big[ \rk \Big\{ \Em \stepa \suml f'_\al(\al^M_{lm_\stepa}) \sumk  \belief_\al(\al^M_k) \belief(\al^M_k)\qmk\Big\}\Big]\\
\aad \ = \lim_{\stepa\to 0} \E\Big[\rk \Big\{\Em \stepa \suml f'_\al (\al^M_{lm_\stepa}) \sumk
\belief_\al(\al^M_{lm_\stepa}) \belief(\al^M_{lm_\stepa})\qM \Big\} \Big]\\
\aad \ = \lim_{\stepa\to 0} \E\Big[\rk \Big\{\Em  \suml f'_\al(\al^{\stepa,M}(l\delta_\stepa)) \belief_\al(\al^{\stepa,M}( l \delta_\stepa)) \belief(\al^{\stepa,M}( l\delta_\stepa)) \qMl \delta_\stepa \Big\} \Big]\\
\aad \ = \E\Big[\rkm \Big\{ \int^{t+s}_t f'_\al(\al^M(u)) \belief_\al(\al^M(u)) \belief(\al^M(u)) q_M(\al^M(u)) du \Big\} \Big].\earray\eeq

Note that
\beql{drift}
\barray
\ad\!\!\!
\E\Big[
 \rk \Big\{\Em \!\!\suml f'_\al(\al^M_{lm_\stepa}) \sumk
 {\stepa \over \kernelstep^\thdim}  \kernel\( {\th_k-\al^M_k \over \kernelstep} \)
 {\temperature \over 2} \wdt \reward(\th_k,\xi_k)\belief(\al^M_k)\qmk \Big\}\Big]\\
 \aad =  \E\Big[
 \rk \Big\{{\temperature\over 2}\suml \delta_\stepa f'_\al(\al^M_{lm_\stepa})\belief(\al^M_{lm_\stepa})
 \\
 \aad \hspace*{1.8in} \times {1\over \kernelstep^\thdim} {1\over m_\stepa}\sumk\Em
 \kernel\( {\th_k-\al^M_k \over \kernelstep} \)
 \wdt \reward(\th_k,\xi_k)\qmk \Big\}\Big]+ o(1),
 \earray\eeq
 where $o(1)\to 0$ as $\stepa\to 0$ uniformly in $t$.
 By the continuity of $\belief\cd$ and $\wdt \reward(\cdot,\xi)$ for each $\xi$,
\bea  \psi_\stepa \ad = {1\over \kernelstep^\thdim} {1\over m_\stepa}\sumk \Em
 \kernel\( {\th_k-\al^M_k \over \kernelstep} \)
 \wdt \reward(\th_k,\xi_k) \qmk\\
 \ad  = {1\over \kernelstep^\thdim} {1\over m_\stepa}\sumk
 \Em  \Big[ \int
 \kernel\( {\th-\al^M_k \over \kernelstep} \)
 \wdt \reward(\th,\xi_k) \belief(\th) d\th \Big]_{\th=\th_k}\qmk\\
 \aad \quad + {1\over \kernelstep^\thdim} {1\over m_\stepa}\sumk
 \Em \Big[\int
 \kernel\( {\th-\al^M_k \over \kernelstep} \)
 \wdt \reward(\th,\xi_k)[ \belief(\th|{\mathcal F}_{l m_\stepa})- \belief(\th)] d\th \Big]_{\th=\th_k} \qmk\\
\eea
In view of \ref{H2}, the last term above contributes nothing to the limit.
By virtue of \eqref{eq:k-est},
\bea \ad {1\over \kernelstep^\thdim} {1\over m_\stepa}\sumk
 \Em  \Big[ \int
 \kernel\( {\th-\al^M_k \over \kernelstep} \)
 \wdt \reward(\th,\xi_k) \belief(\th) d\th \Big]_{\th=\th_k}\qmk\\
 \aad \  ={1\over m_\stepa}\sumk
 \Em
 \wdt \reward(\al^M_k,\xi_k) \belief(\al^M_k)\qmk +o(1),\eea
 where $o(1)\to 0$ in probability.
Thus we have
\beql{psi-ave}
\barray
 \psi_\stepa \ad= {1\over m_\stepa}\sumk \Em \wdt \reward(\al^M_k, \xi_k)
 \belief(\al^M_k)\qmk +o(1) \\
 \ad={1\over m_\stepa}\sumk \Em \wdt \reward(\al^M_{lm_\stepa}, \xi_k)
 \belief(\al^M_{lm_\stepa})\qmk +o(1)\\
 \ad={1\over m_\stepa}\sumk \Em \wdt \reward(\al^\mm(l\delta_{\stepa}), \xi_k)
 \belief(\al^\mm({l \delta_\stepa}))\qMl +o(1),
  \earray\eeq
where $o(1)\to 0$ in probability as $\stepa\to 0$,
because of the continuity of $\belief\cd$ and $\wdt \reward(\cdot,\xi)$
for each $\xi$.
Letting $l \delta_\stepa \to u$ as $\stepa\to 0$,
then for any $l m_\stepa \le k \le m_\stepa + m_\stepa$,  $\stepa k\to u$.
Using the weak convergence of $\al^\mm\cd$ to $\al^M\cd$ and the Skorohod representation,
we  can approximate $\wdt \reward(\al^\mm(l\delta_{\stepa}), \xi_k)
 \belief(\al^\mm({l \delta_\stepa}))q_M(\al^\mm({l \delta_\stepa}))$ by
 $\wdt \reward(\al^M(u),\xi_k) \belief(\al^M(u)) q_M(\al^M(u))$ with an error going to 0. Because $\al^M\cd$ is bounded, for each $ \gamma >0$, we can choose $\{O^\gamma_i: i\le i_\gamma\}$ as a finite collection of disjoint sets of diameter no larger than $\gamma$ whose union covers the range of $\al^M\cd$. Thus, $\al^M\cd$ can be approximated by $\sum_{i=1}^{i_\gamma}\al^\gamma_i 1_{\{ \al^M(u) \in O^\gamma_i\}}$. Consequently,
 \beql{psi-ave-1}
\barray
 \psi_\stepa \ad= {1\over m_\stepa} \sum_{i=1}^{i_\gamma} \sumk \Em
\wdt \reward(\al^\gamma_i, \xi_k)
 \belief(\al^\gamma_i) 1_{\{ \al^M(u) \in O^\gamma_i\}}q_M(\al^M(u)) +o(1),
  \earray\eeq
  where $o(1)\to 0$ in probability.
  Now it is clear that condition \ref{H3} and hence \eqref{ave} can be used. Using \eqref{psi-ave-1} and \eqref{ave}
  together with \eqref{drift} and detailed calculation yields that
  \beql{drift-lim}
\barray
\ad\!\!\!
\E\Big[
 \rk \Big\{\Em \suml f'_\al(\al^M_{lm_\stepa}) \sumk
 {\stepa \over \kernelstep^\thdim}  \kernel\( {\th_k-\al^M_k \over \kernelstep} \)
 {\temperature \over 2} \wdt \reward(\th_k,\xi_k)\belief(\al^M_k)\qmk \Big\}\Big]\\
\aad \to \E\Big[ \rkm \Big\{ {\temperature \over 2}\int^{t+s}_t \Reward_\al(\al^M(u)) \belief^2(\al^M(u)) q_M(u) du \Big\}\Big].\earray\eeq

Using \eqref{es1} and \eqref{es2},  and combining the estimates and calculation in \eqref{es2-e}-\eqref{drift-lim} lead to  \eqref{mg-M}. Therefore,
we arrive at that $\al^M\cd$ is the solution of the martingale problem with operator $\op^M$ given in \eqref{M-op}.

{\underbar{Step 4.
Let the truncation level $M\to \infty$.}}
In the last step, we let $M\to \infty$ to obtain the convergence of the un-truncated process $\al^\stepa\cd$. The details are as in \citet[pp. 44-46]{Kus84}. The verbatim argument is thus omitted.

Now, our arguments in Steps 1-4 yield the desired result Theorem \ref{thm:weak-conv}. The proof of the theorem is concluded.

\subsection{Comments}
We make two remarks  below.
\begin{compactitem}
\item We proved Theorem \ref{thm:weak-conv} above for  algorithm \eqref{eq:irl-1}; equivalently (\ref{eq:irl}). The proof of convergence of \eqref{eq:irl-2} can be carried out similarly. The main difference is that we are utilizing the kernel $\kernel\cd$ to incorporate $\th_k$ used in the algorithm. 
  There is no additional technical difficulty.
\item
Note that in a way, \eqref{eq:irl-1} can be considered to be more efficient than \eqref{eq:irl-2}. First, because $\belief(\al)$ is available,
 \eqref{eq:irl-1} is more direct. Second, using  $\belief(\al)$ and $\belief_\al(\al)$
 in lieu of using $\belief(\th)$ and $\belief_\al(\th)$ together with the kernel $K\cd$ avoids an additional averaging and the involvement of a Dirac $\delta$-like function.

\end{compactitem}

\section{Tracking Analysis of IRL in Non-Stationary Environment} \label{sec:markov}
An important feature of the IRL algorithm (\ref{eq:irl}) is its constant step size $\stepa$  (as opposed to a decreasing step size).  This facilities  estimating (tracking) time evolving reward functions. This section analyzes the ability of IRL algorithm  to track  a time-varying reward  function.

Since we are  estimating  a time evolving  reward, we first  give a model for the evolution of the reward
$\Reward(\th)$ over time.
Below, the Markov chain $\{\mc_\dtime\}$ will be used as a \textit{ hyper-parameter} to model the evolution
of the time varying  reward, which we will denote as $\Reward(\th,\mc_\dtime)$.  By hyper-parameter we mean that the Markov chain model is not known or used  by the IRL  algorithm (\ref{eq:irl}). The Markov chain assumption is used  only in  our convergence analysis
to determine how well does the IRL  algorithm estimates (tracks) the  reward $\Reward(\th,\mc_\dtime)$  that jump changes
(evolves) according to an unknown Markov chain $\mc_\dtime$

We assume that the RL agents perform gradient algorithm (\ref{eq:rl}) by
evaluating the sequence of gradients  $\{\nabla_\th \reward_\dtime(\th_\dtime, \mc_\dtime)\}$. Note that both   the RL and IRL do not know the sample path $\{\mc_\dtime\}$.
We will use similar   notation to  Sec.\ref{sec:weak}:
\begin{compactitem} \item
  Denote $ \nabla_\th\reward_\dtime(\th_\dtime, \mc_\dtime) $ as $\wdt \reward (\th_k,\xi_k,\mc_\dtime)$,
  \item We use $\belief_\al(\cdot)$ to denote $\nabla_\al \belief(\cdot)$.
\end{compactitem}

\subsection{Assumptions}
We focus on the following algorithm
\beql{eq:irl-1a}\al_{k+1}=\al_k + {\stepa \over \kernelstep^\thdim} K\( {\th_k-\al_k \over \kernelstep} \) {\temperature \over 2} \wdt \reward(\th_k,\xi_k,x_k)\belief(\al_k)+ \stepa \belief_\al(\al_k) \belief(\al_k) +\sqrt \stepa \belief(\al_k) \noise_k,
\eeq
The  main assumptions are as follows.
\begin{enumerate}[label=(M{\arabic*})]
  \item  \label{M1} (Markovian hyper-parameter)   Let $\{\mc_\dtime, \dtime \geq 0\}$ be a Markov chain with finite state space  $\statespace=\{1,\dots,X\}$ and
  transition probability matrix $I +\mcstep Q$, where $\mcstep>0$ is a small parameter and $Q=(q_{ij})$ is an
  $\statedim \times \statedim$ irreducible generator (matrix) \cite[p.23]{YZ13} with $$q_{ij}\ge 0, \quad i\not =j, \qquad
  \sum_j q_{ij}=0 , \quad   i \in \statespace ,$$ also $\{x_k\}$ is independent of $\{\th_k\}$ and $\{w_k\}$.
\item \label{M2} Assumption \ref{H1} holds on   $ \wdt \reward(\cdot,\xi, i)$ for each fixed  state $i \in \statespace$.
    Also  \ref{H2}, \ref{H3}, \ref{H4} hold.
\end{enumerate}

\subsection{Main Result}
Recall that $\stepa$ is the step size of the IRL algorithm while $\mcstep$ reflects the rate at which the hyper-parameter Markov chain $\mc_\dtime$ evolves.   In the following tracking analysis of IRL algorithm (\ref{eq:irl}) , we will consider three cases,
 $\stepa = O(\mcstep)$, $\stepa \ll \mcstep$, and $\stepa \gg \mcstep$.
 The three cases represent three different types of asymptotic behavior.
 If $\stepa \gg \mcstep$, the frequency of changes of the Markov chain is very slow. Thus, we are treating a case similar to a
 constant parameter, or
 we essentially deal with a ``single'' objective function.
 If $\stepa \ll \mcstep$, then the Markov chain jump changes frequently.
 So what we are optimizing is a function
 $\sum^X_{i=1} R(\al,i) \nu_i$,
 where $\nu_i$ is the stationary distribution  associated with the generator $Q$.
 If $\stepa =O(\mcstep)$, then the Markov chain changes in line with the optimization recursion. In this case, we obtain switching limit
 Langevin diffusion.

 In  Theorem \ref{thm:gld1} below, for brevity we use  $\stepa =\stepmc$ for $\stepa = O(\stepmc)$,
 $\mcstep = \stepa^{1+\wdt\Delta}$ for $\mcstep = o(\stepa)$ and   $\mcstep= \stepa ^{\wdt\Delta}$
 for $\stepa = o(\mcstep)$, respectively. These cover all three possible cases of the rate at which the hyper-parameter evolves compared to the dynamics of the Langevin IRL algorithm.

 \begin{theorem} \label{thm:gld1}
   Consider the
   IRL algorithm \eqref{eq:irl-1a}.
Under Assumptions {\rm\ref{M1}} and {\rm\ref{M2}}, assuming that
\eqref{eq:sde_track}, or \eqref{eq:sde_track2}, or \eqref{eq:sde_track3} has a unique solution in the sense in distribution. Then
   the following results hold.
    \begin{itemize}
    \item[{\rm 1.}]
Assume $\stepa =\stepmc$.
Then as $\stepa \downarrow 0$,
the interpolated process $(\eth^\stepa\cd,\mc^\stepa\cd)$  converges weakly to the switching diffusion $(\eth\cd, \mc\cd)$ satisfying
\beq  d\al(t) = \Big[ {\temperature \over 2} \belief^2(\al(t)) \Reward_\al(\al(t),\mc(t))+ \belief_\al(\al(t)) \belief(\al(t))\Big] dt+ \belief(\al(t) ) d\bm(t), \\
 \label{eq:sde_track} \eeq
where $\bm(\cdot)$ is a standard Brownian motion with mean 0 and covariance being the identity matrix $I \in \reals^{\thdim\times \thdim}$,
 and  $x\cd$ is a continuous-time Markov chain with generator $Q$.
    \item[{\rm 2.}]
    Suppose $\mcstep = \stepa^{1+\wdt\Delta}$ with $\wdt\Delta > 0$ and denote the initial distribution of $x^\mcstep(0)$ by $p_\iota$ $($independent of $\mcstep)$ for each $\iota \in {\cal X}$.
      Then  as $\stepa \downarrow 0$,
    the interpolated process $(\eth^\stepa\cd)$  converges weakly to the following
    diffusion process
    \beq d\al(t) = \Big[ {\temperature \over 2} \belief^2(\al(t)) \sum_{\iota \in \statespace} \Reward_\al(\al(t),\iota )\, p_\iota+ \belief_\al(\al(t)) \belief(\al(t))\Big] dt+ \belief(\al(t) ) d\bm(t), \\
 \label{eq:sde_track2} \eeq
\item[{\rm 3.}]  Suppose that $\mcstep= \stepa ^{\wdt\Delta}$ with $0<\wdt\Delta<1$ and denote the stationary distribution associated with the continuous-time Markov chain with generator $Q$ by $\nu =(\nu_1,\dots, \nu_{X})$. Then  as $\stepa \downarrow 0$,
    the interpolated process $(\eth^\mu\cd)$  converges weakly to the following
    diffusion process
     \beq  d\al(t) = \Big[ {\temperature \over 2} \belief^2(\al(t)) \sum_{\iota \in \statespace} \Reward_\al(\al(t),\iota )\, \nu_\iota+ \belief_\al(\al(t)) \belief(\al(t))\Big] dt+ \belief(\al(t) ) d\bm(t).
     \label{eq:sde_track3} \eeq
   \end{itemize}
 \end{theorem}

{\em Remark}.
Theorem \ref{thm:gld1} presented the asymptotic behavior of the IRL  algorithm (\ref{eq:irl-1a})  with Markovian switching.  In accordance with the rates of variations of the adaptation rates (represented by the stepsize $\stepa$)
and the switching rate (represented by the stepsize $\mcstep$),
three cases are considered.  Case~1 indicates that when $\stepa$ is in line with $\stepmc$, the limit differential equation is a switching diffusion. Case~2 concentrates on the case that the switching is much slower than the stochastic approximation generated by the recursion. Thus, the limit Langevin equation is one in which the drift and diffusion coefficients are averaged out with respect to the initial distribution of the limit Markov chain. Roughly, it reveals that the ``jump change'' parameter $\mc(t)$ is more or less as a constant in the sense the coefficients are averages w.r.t. the initial distribution.
Case~3 is the one that the Markov chain is changing much faster than the stochastic approximation rate. As a result, the ``jump change'' behavior is replaced by an average with respect to the stationary distribution of the Markov chain.  Then we derive the associated limit Langevin equation. Again, the limit has no switching in it.

\subsection{Proof of Theorem \ref{thm:gld1}}

We will prove  Statement 1 for the case $\mu=\eta$.
Consider $(\alpha^\mu(\cdot),x^\mu(\cdot))$, the pair of interpolated processes. We shall show that
this  pair of processes converges weakly to $(\alpha(\cdot), x(\cdot))$ such that the limit is a solution of \eqref{eq:sde_track}
or equivalently, $(\alpha(\cdot), x(\cdot))$ is a solution of the martingale problem with an operator redefined by
\beql{eq:5-op}
{\mathcal L} f(\alpha, i)
= f_\al'(\al,i) \Big[ {\temperature \over 2} \belief^2(\al,i) R_\al(\al) + \belief_\al(\al)\belief(\al) \Big]
+ {1\over 2} \belief^2(\al) \Tr[ f_{\al\al}(\al,i)] +Qf(\al,\cdot)(i), \eeq
where $$ Qf(\al,\cdot)(i)= \sum_{j \in \statespace} q_{ij} f(\al, j), \ \hbox{ for each }\  i \in \statespace.$$
We still need to use an $M$ truncation device (truncation on $\al$). However, for notation simplicity, we suppress the $M$ truncation.
From \eqref{eq:irl-1a}, it is easily seen that
\beql{eq:irl-1ab}
\al_{k+1}=\al_k + {\stepa \over \kernelstep^\thdim} K\( {\th_k-\al_k \over \kernelstep} \) {\temperature \over 2}\sum_{i\in  \statespace}\wdt \reward(\th_k,\xi_k,i)\belief(\al_k)1_{\{ x_k =i\}}+ \stepa \belief_\al(\al_k) \belief(\al_k) +\sqrt \stepa \belief(\al_k) \noise_k.
\eeq
To prove the tightness of
 $(\alpha^\mu(\cdot),x^\mu(\cdot))$, we prove the tightness $\{x^\mu(\cdot)\}$ first.
 This can be done by considering $\chi_k =(1_{\{x_k=1\}}, \dots, 1_{\{x_k=X\}}) \in \reals^{1\times X},$  and defining
 $\chi^\mu(t)= \chi_k$ for $t\in [\mu k, \mu k+\mu)$.
 Denote by ${\cal F}_t^\mu$, the $\sigma$-algebra generated by $\{\xi_k,\th_k, x_k, \al_0: k \le t/\mu\}$, and denote the corresponding conditional expectation by $\E^\mu_t$.
 Because $\chi_k$ is a Markov chain and because of the independence of $x_k$ with $\xi_k$ and $\th_k$, we can show for any $\delta>0$,
 $t>0$, $s>0$ with $s\le \delta$, for some random variable $\wdh \gamma^\mu(\delta)>0$,
 $$\sup_{0\le s\le \delta} \E_{t}^\mu [| \chi^\mu(t+s)-\chi^\mu(t)|^2\big| {\cal F}^\mu_t]\le \E^\mu_t\wdh\gamma^\mu(\delta).$$
 Furthermore, $$\lim_{\delta\to 0}\limsup_{\mu\to 0} \E \wdh \gamma^\mu(\delta)=0,$$
 which implies the tightness of $\{\chi^\mu(\cdot)\}$ (see \cite[p. 47, Theorem 3]{Kus84} and hence the tightness of $\{x^\mu(\cdot)\}$. We can also prove the tightness of $\{\al^\mu(\cdot)\}$. Then the tightness of $\{\al^\mu\cd, x^\mu\cd\}$ can be proved. The rest of the averaging procedure is similar to that of the proof of Theorem \ref{thm:weak-conv}.

 For the proofs of Statement 2, the  case $\mcstep = \stepa^{1+\wdt\Delta}$, and Statement 3, the case $\mcstep= \stepa ^{\wdt\Delta}$, the arguments are similar to \cite{YYW13}
 Section 4.1 and Section 4.2, respectively.
 We thus omit the details.

 \section{Proof of Convergence of IRL Algorithm \eqref{eq:mcmcirl}} \label{sec:proofmcmcirl}
 \blue{Here we prove weak convergence of the multi-kernel variance reduction IRL algorithm
 \eqref{eq:mcmcirl}. The proof involves a novel application of the Bernstein von-Mises
 theorem (which in simple terms is a central limit theorem for a Bayesian estimator); see
\eqref{p-app} below.}

Recall that we write $\nabla r_k(\theta)$  as $\wdt r(\theta, \xi_k)$ as in the proof of Theorem \ref{thm:weak-conv}. The algorithm \eqref{eq:mcmcirl} is
\beq
\begin{split}
  \eth_{\dtime+1} &= \eth_\dtime + \stepa \,\frac{\temperature}{2} \, \sum_{i=1}^{L_\mu}  \frac{  \pdf(\eth_\dtime|\th_i) \wdt r(\th_i,\xi_k) }{\sum_{l=1}^{L_\mu} \pdf (\eth_\dtime|\th_l)}
  + \sqrt{\stepa}  \noise_\dtime ,
     \end{split} \label{eq:mcmcirl-a}
     \eeq
where $L_\mu$ is so chosen that $L_\mu\to \infty$ as $\mu\to 0$.

We start  with the  following assumptions:

\begin{enumerate}[label=(B{\arabic*})]
\item \label{B1} \ref{H1} holds and
the reward $R\cd$ has continuous partial derivatives up to the second order and the second-order partial of $R$ is uniformly bounded.
\item  \label{B2} The $\{\th_l\}$ is a stationary sequence
$\th_l \sim \belief(\cdot)$; $\{\th_l\}$ is
independent of
 $\{\xi_k\}$ and $\{w_k\}$, where $\{\xi_k\}$ and $\{w_k\}$  satisfy \ref{H3}.
\item  \label{B3} For each fixed $\al$, and each $i=1,\dots, L_\mu$, define
$$\gamma_i(\al) = {\pdf(\al|\th_i)\over \sum^{L_\mu}_{l=1} \pdf(\al|\th_l )}.$$    For each $\xi$ and each $\al$, as $\mu\to 0$, $L_\mu\to \infty$ and
\beql{eq:ave-p} \sum^{L_\mu}_{i=1} \gamma_i(\al)  \wdt r(\th_i,\xi) \to \E \wdt r(\th,\xi| \al) \ \text{ w.p.1.}
\eeq
\item \label{B4} $\E |\wdt r(\th_i,\xi_k)|^2 < \infty$ for each $i$ and each $k$, and $ \int (1+ |\nabla R(\th)|^2) (p(\th/\al)/ \belief(\th))^2 \belief(\th) d\th < \infty.$
\item \label{B5} (a)  The conditional probability density function
        \beql{eq:cond-al2}p(\eth|\th) =  p_v(\th - \al)\eeq
        where $ p_v(\cdot)$  is
a symmetric density with zero mean and covariance $O(\kernelstep^2) I$.
where $I$ denotes the identity matrix.
\\
(b) The Fisher information matrix $I_{\th}= \int_{\reals^\thdim} 
\nabla
\log \pdf(\eth| \th)\,
\pdf(\eth|\th) \, d\eth $ is invertible for all $\th
\in \reals^\thdim$.
\end{enumerate}

\noindent {\it Remarks.}
We briefly comment on the assumptions.
\ref{B1}  is a  smoothness assumption on  $\wdt \reward(\cdot,\xi)$ and
$\Reward(\cdot)$. The second order differentiability of $\Reward(\cdot)$ is used in a Taylor series expansion in  Proposition~\ref{thm:conv1} to obtain the final stochastic diffusion limit. Note that \ref{B1} is a stronger assumption than \ref{H1}.

Condition \ref{B2} specifies the distribution of $\th_l$.  We also assume that this sequence is independent of the $\xi_k$ and $w_k$. The assumption builds on \ref{H3}.

\ref{B3} is an averaging condition;
  i.i.d. samples $\{\th_i\}$
  is a special case. In fact,
  we only need the convergence to be in the sense of convergence in probability.

\ref{B4} is a
 classical square integrability assumption for asymptotic normality.

Finally,  \ref{B5} is used in the Bernstein von-Mises theorem to show that the posterior $\pdf(\th|\eth)$ is asymptotically normal and behaves as a Dirac delta as $\kernelstep \downarrow 0$; see Proposition .\ref{thm:conv1} below.


{\color{black}
As in our previous proofs,
we define the  interpolated process $\al^\mu(t)= \al_k$, $ t \in [\stepa k, \stepa k+ \stepa)$.
For convenience, the proof proceeds in two steps: In the first step, Proposition \ref{prop:conv} below shows  that $\al^\mu\cd$ converges weakly to $\al\cd$ such that
$\eth(t)$ satisfies the stochastic differential equation    (\ref{eq:avgmcmc})  w.r.t.  conditional expectation $\pdf(\th|\eth(t))$.

\begin{proposition}\label{prop:conv}
Assume conditions {\rm \ref{B1}--\ref{B5}}
hold
and that the stochastic differential
equation
 \beq \label{eq:avgmcmc2}
  d \eth(t) = \int_{\reals^\thdim} \frac{\temperature}{2}\,\nabla \Reward(\th) \, \pdf\big(\th| \eth(t)\big) \, d\th \, dt + d\bm(t)  , \qquad \eth(0) = \eth_0
  \eeq
 has a unique
 solution in the sense in distribution
 for each initial condition. Then the interpolated process
$\al^\mu \cd$ converges weakly to $\al\cd$ such that
$\al\cd$ is the solution of \eqref{eq:avgmcmc2}.
\end{proposition}

Note that in the above, we used the uniqueness solution in the weak or distribution sense. Such a uniqueness
is equivalent to  the uniqueness of the associated martingale problem; see \citet[p. 182]{EK86}
or \citet{KS91}.

In the second step,
we use the Bernstein von-Mises theorem below to characterize the posterior as a normal distribution when the parameter $\kernelstep$ in the likelihood   density goes to zero.
The Bernstein-von Mises theorem
\citep{Vaa00}
implies that for small parameter  $\kernelstep$ in the likelihood  (\ref{eq:cond-al}),
the posterior converges to the Gaussian
density
  $\normal(\th;\eth, \kernelstep^2 I_{\bar\th})$.   More precisely,
\beql{p-app}\int |\pdf(\th|\eth)- \normald(\th; \al, \kernelstep^2 I_{\bar\th} )
| d\th \rightarrow 0
\text{ in probability under } P_{\bar \th}  \hbox{ as } as \kernelstep\to 0.\eeq
Here
$I_{\bar \th}= \int_{\reals^\thdim} \nabla
\log \pdf(\eth| \th)\,
\pdf(\eth|\th) \, d\eth \vert_{\th = \bar \th}$
is  the  Fisher information matrix  evaluated at the
parameter value\footnote{It suffices to  choose any  $\bar \th$ such that
$\eth \sim \pdf(\cdot|\bar \th)$. The precise value of $\bar \th$ need not be known
 and is irrelevant to our analysis.} $\bar \th$
and
\beq \label{Sig}
 \normald(\th; \al, \kernelstep^2 I_{\bar\th} )
 =
{2 \pi}^{-\thdim/2}
\exp\Big[ -\frac{1}{2} (\th-\eth)^\p   |\kernelstep^2 I_{\bar\th}^{-1}|^{-1}
(\th-\eth) \Big].
\eeq
In view of the parametrization by $\Delta$ above, $\al^\mu\cd$ should be written as $\al^{\mu, \Delta}\cd$.

 \begin{proposition}\label{thm:conv1}
 Assume conditions {\rm\ref{B1}} to {\rm\ref{B5}}, and \eqref{p-app} hold.
 Then the limit in Proposition {\rm\ref{prop:conv}} can be written as $\al^{\Delta}(t)$. As $\Delta\to 0$, $\al^{\Delta}(t)$ has the limit
 $\al(t)$ satisfying
 \beql{ode-r} d \al (t)=  \frac{\temperature}{2}\, \nabla \Reward(\al(t)) \, dt + d\bm(t), \qquad \eth(0) = \eth_0 . \eeq
\end{proposition}

\subsection{Proof Outline of Proposition \ref{prop:conv}}

We present the main ideas of the proof and the underlying
intuition.
Define $\al^\mu(t)= \al_k,$ for $t\in [\mu k, \mu k +\mu)$. As in the proof
of Theorem \ref{thm:weak-conv} in Sec.\ref{sec:proofmain},  we should still use a truncation device and use
the martingale problem formulation. However, to present  the main idea without
overburdening with technical details, we will use simpler and intuitive ideas. Thus we simply assume that the iterates are bounded.
 For example, we should use a smooth function with compact support $f\cd$ as in the proof of Theorem \ref{thm:weak-conv}. However, for simplicity of argument, we will illustrate the idea without using this function $f\cd$; we will also suppress the truncation notation.

 Denote by ${\cal F}^\mu_t$, the $\sigma$-algebra generated by $\{\th_k,\xi_k, \al_0: k \le \lfloor t/\mu \rfloor\}$, where $\lfloor s \rfloor$ denotes the integer part of $s$. In what follows, we shall suppress the floor function notation.
Denote by $\E_t^\mu$, the conditional expectation with respect to  ${\cal F}^\mu_t$. We also use $\E_{\xi_k}$ to denote the conditioning on $\{\xi_j: j\le k\}$.
For any $\delta>0$,  $t>0$, $s>0$ and
$s\le \delta$, by the boundedness of the iterates, conditions \ref{B1}, \ref{B2}, the form of $\gamma_i(\al)$ in \ref{B3}, and \ref{B4},
we have
\bea \ad \E^\mu_t | \al^\mu(t+s)- \al^\mu(t)|^2\\
\aad \ \le K \Big[ \E^\mu_t \Big|\mu \sum^{(t+s)/\mu -1}_{k=t/\mu}  \sum_{i=1}^{L_\mu}
\gamma_i(\al_k)  \wdt r(\th_i,\xi_k) \Big|^2
  + \E^\mu_t \Big|\sqrt \mu\sum^{(t+s)/\mu -1}_{k=t/\mu}  w_k\Big|^2
  \Big]\\
\aad \ \le K s \mu\sum^{(t+s)/\mu-1}_{k=t/\mu}\E^\mu_t \Big| \sum_{i=1}^{L_\mu}
\gamma_i(\al_k)  \wdt r(\th_i,\xi_k)\Big|^2  + K \mu \E^\mu_t \sum^{(t+s)/\mu-1}_{k=t/\mu} w'_k w_k
 \le \E^\mu_t \wdh\gamma^\mu(\delta),
 \eea
where $\wdh\gamma^\mu(\delta) $ is
a random variable. Moreover,
$$\lim_{\delta\to 0}\limsup_{\mu \to 0} \E \wdh \gamma^\mu(\delta) =0.$$
Thus the tightness of $\{\al^\mu\cd\}$ is obtained; see \citet[p. 47]{Kus84}.

By Prohorov's theorem, we can extract a weakly convergent subsequence.
Select such a sequence and still use $\mu$ as its index (for notional simplicity) with limit
$\al\cd$. By Skorohod representation (without changing notation), $\al^\mu\cd$ converges w.p.1 to $\al\cd$.
Now
for any $t>0$ and $s>0$,
\beql{eq:pre-l}\barray  \al^\mu(t+s)- \al^\mu(t) \ad = {\beta\over 2}\mu\sum^{(t+s)/\mu-1}_{k=t/\mu}\sum_{i=1}^{L_\mu}
\gamma_i(\al_k)  \wdt r(\th_i,\xi_k) + \sqrt \mu \sum^{(t+s)/\mu-1}_{k=t/\mu} w_k.\earray\eeq
Define
$$ \bm^\mu (t)=\sqrt \mu \sum^{(t/\mu)-1}_{k=0} w_k .$$
By using the classical functional invariance theorem, clearly
 $\bm^\mu\cd$ converges weakly to $\bm\cd$ a standard Brownian motion. As a consequence,
\bea \bm^\mu(t+s)-\bm^\mu(t)\ad=\sqrt \mu \sum^{(t+s)/\mu-1}_{k=t/\mu} w_k\\
\ad \to \bm(t+s)- \bm(t)\eea
by the weak convergence and the Skorohod representation.
To determine the limit of the drift term, we proceed similarly to  the proof of Theorem \ref{thm:weak-conv}.
In view of \eqref{eq:cond-al}, $\gamma_i(\al)$ is continuous (and in fact smooth) w.r.t. $\al$.
We use the finite value approximation
 argument as just above \eqref{psi-ave-1} together with the averaging condition in \ref{B3}.   That is, for each $\tilde\eta>0$,
we can choose $\{O^{\tilde\eta}_j: j\le j_{\tilde\eta}\}$
 as a
finite collection of disjoint sets of diameter no larger than $\tilde\eta$ whose union covers the range of $\al^\mu(u)$ so $\al^\mu(u)$ can be approximated by $\sum^{j_{\tilde\eta}}_{j=1}\al^{\tilde\eta}_j1_{\{ \al^{\tilde\eta}(u) \in
O^{\tilde\eta}_j\}}$.
Thus using the notation as in the proof of Theorem \ref{thm:weak-conv}, and choosing any positive integer $\kappa_1$ and $t_\iota \le t$ with $\iota \le \kappa_1$,
\beql{eq:lim0}\barray \ad
\lim_{\mu\to 0}
\E g(\al^\mu(t_\iota): \iota \le \kappa_1) \Big[
\mu\sum^{(t+s)/\mu-1}_{k=t/\mu}\sum_{i=1}^{L_\mu}
\gamma_i(\al_k)  \wdt r(\th_i,\xi_k)\Big] \\
\aad \ = \lim_{\mu\to 0}
\E g(\al^\mu(t_\iota): \iota \le \kappa_1) \Big[\sum^{t+s}_{l \delta_\mu =t} \delta_\mu
{1\over m_\mu}\sum^{l m_\mu +m_\mu -1}_{k=l m_\mu} \sum_{i=1}^{L_\mu}
\gamma_i(\al_k)  \wdt r(\th_i,\xi_k)\Big]\\
\aad \ = \lim_{\mu\to 0}
\E g(\al^\mu(t_\iota): \iota \le \kappa_1) \Big[\sum^{t+s}_{l \delta_\mu =t} \delta_\mu
{1\over m_\mu}\sum^{l m_\mu +m_\mu -1}_{k=l m_\mu}\E_{lm_\mu} \sum_{i=1}^{L_\mu}
\gamma_i(\al_{lm_\mu})  \wdt r(\th_i,\xi_k)\Big]\\
\aad \ =\lim_{\mu\to 0}
\E g(\al^\mu(t_\iota): \iota \le \kappa_1) \Big[\sum^{t+s}_{l \delta_\mu =t} \delta_\mu
{1\over m_\mu}\sum^{l m_\mu +m_\mu -1}_{k=l m_\mu} \E_{lm_\mu} \sum^{j_{\tilde\eta}}_{j=1}\sum_{i=1}^{L_\mu}
\gamma_i(\al^{\tilde\eta}_j)  \wdt r(\th_i,\xi_k)1_{\{ \al^{\tilde\eta}(u) \in
O^{\tilde\eta}_j\}}\Big]\\
\aad \ = \lim_{\mu\to 0}
\E g(\al^\mu(t_\iota): \iota \le \kappa_1) \Big[ \sum^{t+s}_{l \delta_\mu =t} \delta_\mu {1\over m_\mu}\sum^{l m_\mu +m_\mu -1}_{k=l m_\mu}
\E_{lm_\mu}\sum^{j_{\tilde\eta}}_{j=1}
 \E_{\xi_k}[ \wdt r(\th,\xi_k)|\al^{\tilde\eta}_j]1_{\{ \al^{\tilde\eta}(u) \in
O^{\tilde\eta}_j\}}\Big]\\
\aad \ =\E g(\al(t_\iota): \iota \le \kappa_1) \Big[
\int^{t+s}_t \int_{\reals^N}\nabla R_\theta(\theta) p(\theta|\al(u)) d\th du\Big],
\earray\eeq
where
$\E_{\xi_k}$ denotes the conditioning on $\{\xi_j: j\le k\}$. In the above, we used \eqref{eq:ave-p}, and noted that
letting $\mu l m_\mu\to u$ yields $\mu k\to u$ for $l m_\mu \le k \le l m_\mu + m_\mu$. We also used \ref{B5}.
Putting the estimates together, we obtain the desired limit. 

\subsection{Proof Outline of Proposition \ref{thm:conv1}}
To prove Proposition \ref{thm:conv1},
using \ref{B5}, by virtue of \eqref{p-app},  $p(\th| \al)$
can be approximated by $ \normald(\th; \al, \kernelstep^2 I_{\bar\th} )$, the normal density given by \eqref{Sig}.
For notational convenience denote $\wdt p(\th,\al) \defn  \normald(\th; \al, \kernelstep^2 I_{\bar\th} )$ below.
  Now, we work with $\kernelstep\to 0$.
By  Taylor expansion,
$$ \nabla \Reward(\th) = \nabla \Reward(\al ) + \nabla ^2 \Reward (\al_+) [ \th -\al],$$ where  $\nabla^2 \Reward$ is the Hessian (the second partial derivatives) of $\Reward$, and $\al_+$ is on the line segment joining $\th$ and $\al$. Choose
$v$ so that
$\e l m_\e \to v$.  As a result, for any $k$ satisfying $ lm_\e \le k \le lm_\e +m_\e$,
$\e k \to v$.
It follows that
\beql{nabC}
\begin{split}  \int_{\reals^\thdim} \nabla \Reward(\th) p(\th | \al(v)) d\th
&  =  \int_{\reals^\thdim} \nabla \Reward(\th)
\wdt p(\th, \al(v)) d\th + o_\kernelstep(1) \\
&
=  \int_{\reals^\thdim}\nabla \Reward(\al) \wdt p(\th, \al(v)) d\th \\
&\qquad + \int_{\reals^\thdim}\nabla^2 \Reward(\al_+(v)) [\th -\al(v)]\wdt p(\th, \al(v)) d\th +o_\kernelstep(1) \\
& = \nabla \Reward(\al(v)) +o_\kernelstep(1)\\
&  \to \nabla \Reward(\al(v)) \ \hbox{ as } \ \kernelstep \to 0,
\end{split}\eeq
where
$o_\kernelstep(1) \to 0$ in probability as $\kernelstep\to 0$.
 Note that in the above,
 the form of the density implies
$$\int_{\reals^\thdim}\nabla^2 \Reward(\al_+(v)) [\th -\al(v)]\wdt p(\th, \al(v)) d\th =0.$$

}

\section{Conclusions and Extensions}

This paper has presented and analyzed the convergence of passive Langevin dynamics algorithms for adaptive inverse reinforcement learning (IRL). Given noisy gradient estimates of a possibly time evolving reward function $\Reward$, the Langevin dynamics algorithm generates samples $\{\eth_k\}  $ from the Gibbs measure $\stat(\eth)   \propto  \exp \bigl(  \temperature \Reward(\eth ) \bigr)$; so the log of the empirical distribution of $\{\eth_k\}$ serves as a non-parametric estimator for  $\Reward(\eth)$.
The proposed algorithm is  a {\em passive} learning algorithm since the gradients are not evaluated at $\eth_\dtime$ by the inverse learner; instead the gradients are evaluated at the random points  $\th_\dtime$ chosen by the gradient (RL) algorithm. This passive framework  is natural in an IRL where the inverse learner
passively observes  forward learners.

Apart from the main IRL algorithm (\ref{eq:irl}), we presented
a two-time scale
IRL algorithm for variance reduction, an active IRL algorithm which deals with mis-specified gradients, and a non-reversible diffusion IRL algorithm with larger spectral gap and therefore faster convergence to the stationary distribution.
We presented three detailed numerical  examples: inverse Bayesian learning, a large dimensional IRL problem in logistic learning involving a real dataset, and IRL for a constrained Markov decision process.
Finally, we presented a complete weak convergence proof of the IRL algorithm using martingale averaging methods. We also analyzed the tracking capabilities of the IRL algorithm when the utility function jump changes according to a slow (but unknown) Markov chain.

{\bf Extensions.} A detailed proof of the two-time scale variance reduction algorithm involves Bayesian asymptotics, namely, the Bernstein von Mises theorem. Since the submission of the current paper, in a recent work \citep{KY20a}, we have developed a complete convergence proof.  It is important to note that the IRL algorithms proposed in this paper are adaptive: given the estimates from an adaptive gradient algorithm, the IRL algorithm learns the utility function. In other words, we have a  gradient algorithm operating in series with a Langevin dynamics algorithm. In future work it is of interest to study the convergence properties of multiple such cascaded Langevin dynamics and gradient algorithms.
Finally,   the recent paper by \citet{KHH20}  shows that classical Langevin dynamics   yields more robust RL algorithms compared to classic stochastic gradient. In analogy to \cite{KHH20}, in future work it is worthwhile exploring how our passive Langevin dynamics framework can be viewed as a robust version of classical passive stochastic gradient algorithms.

\bibliographystyle{abbrvnat}
\bibliography{$HOME/texstuff/styles/bib/vkm}

\newpage
\appendix

\section{Matlab Source Code for IRL Algorithm (\ref{eq:irl}) in inverse Bayesian learning  of Sec.\ref{sec:bayesian}. (Generates
 Figure \ref{fig:mh})}

\lstset{language=Matlab,%
  breaklines=true,%
   basicstyle=\footnotesize,
    morekeywords={matlab2tikz},
    keywordstyle=\color{blue},%
    morekeywords=[2]{1}, keywordstyle=[2]{\color{black}},
    identifierstyle=\color{black},%
    stringstyle=\color{mylilas},
    commentstyle=\color{mygreen},%
    showstringspaces=false,
    numbers=left,%
    numberstyle={\tiny \color{black}},
    numbersep=9pt, 
    emph=[1]{for,end,break},emphstyle=[1]\color{red}, 
}
\begin{lstlisting}
%  IRL algorithm for mutimodal mixture
mixture_weight = 0.5; nsamples=100; sigp1=sqrt(10); sigp2 = 1;  sigl1 = sqrt(2); sigl2 = sqrt(2); mixture_mean1 = 0; mixture_mean2 = 1;
T = 80000000; th=[0;0];  alfa = randn(2,1); 
step =1e-3;  lang_step1 = 1e-5; lang_step2 =  sqrt(lang_step1);  kernelstep=0.2;
Cker = 1/(2*pi*kernelstep^2); kernelstepsq = 2*kernelstep^2;
alphaIRL = zeros(2,T); est=zeros(2,T);

for iter = 1: T
   th = randn(2,1);  
  
  % simulate data
      if rand <  mixture_weight
          y = sigl1 * randn + mixture_mean1;
     else
   	 y = sigl2 * randn + mixture_mean1 + mixture_mean2;
     end;

   t1 = th(1); t2 = th(2);

% evaluate gradients
    grad1 = nsamples*((mixture_weight*exp(-(t1 - y)^2/(2*sigl1^2))*(2*t1 - 2*y))/(2*sigl1^3) - (exp(-(t1 + t2 - y)^2/(2*sigl2^2))*(mixture_weight - 1)*(2*t1 + 2*t2 - 2*y))/(2*sigl2^3))/((exp(-(t1 + t2 - y)^2/(2*sigl2^2))*(mixture_weight - 1))/sigl2 - (mixture_weight*exp(-(t1 - y)^2/(2*sigl1^2)))/sigl1) - t1/(sigp1^2); 

 grad2 = - t2/(sigp2^2) - nsamples*(exp(-(t1 + t2 - y)^2/(2*sigl2^2))*(mixture_weight - 1)*(2*t1 + 2*t2 - 2*y))/(2*sigl2^3*((exp(-(t1 + t2 - y)^2/(2*sigl2^2))*(mixture_weight - 1))/sigl2 - (mixture_weight*exp(-(t1 - y)^2/(2*sigl1^2)))/sigl1)) ;

% passive Langevin dynamics
gaus =  exp(-alfa(1)^2/2) *  exp(-alfa(2)^2/2) / ( 2 * pi);
alfa = alfa + Cker*exp( -(norm(th-alfa))^2/kernelstepsq)* (lang_step1/2  * [grad1;grad2]/gaus) +  lang_step2  * randn(2,1);
alphaIRL(:,iter) = alfa;

% classical stochastic gradient
th = th + step * [grad1; grad2];
est(:,iter) = th;
end;
% Plotting
figure(4);  histogram2(alphaIRL(1,5000:end),alphaIRL(2,5000:end), 'Normalization','probability');
axis([-3,3,-3,3]);  xlabel('$\theta(1)$','Interpreter','latex','FontSize',18);   ylabel('$\theta(2)$','Interpreter','latex','FontSize',18);
 figure(5);  [M1,N1] = hist3(alphaIRL(:,5000:end)',[20,20]);
contour(N1{1}, N1{2}, M1'); axis([-3,3,-3,3]);
 xlabel('$\theta(1)$','Interpreter','latex','FontSize',18);
  ylabel('$\theta(2)$','Interpreter','latex','FontSize',18); grid on; colormap(jet);
\end{lstlisting}

\section{Matlab Source Code for multi-kernel IRL (\ref{eq:mcmcirl}) for Logistic Regression in Sec. \ref{sec:logistic}}

\begin{lstlisting}
%  Multi-kernel IRL  for logistic  regression a9a dataset

%Algorithm parameters
 lang_step1 = 0.25e-3; lang_step2 = sqrt(lang_step1);  sigma_theta=1; 
 sigma_kernel = 0.1;  L =100; incrementsig = 0.1;
 
% Read a9a Dataset
 load datasetFeatures.mat;  load datasetLabels.mat;

 thdim=size(features,1) + 1;  T = 32400; ; Nsweep =10; nsamples=10;
 labels(labels<0) = 0;    % set all -1 to 0

est  = zeros(thdim,T*Nsweep); th = zeros(thdim,1); alfa = th;

for sweep = 1: Nsweep
 
    for iter=1:T
				
	  th =   sigma_theta * randn(thdim,L);  % RL chooses th randomly
	   d = vecnorm(th-alfa);
	   weight =10^(2*thdim)*exp(- d.^2/(2*sigma_kernel));
	   	 
	   if   (sum(weight) < 1e-60)
		     alfa = 0.1*randn(thdim,1);  %reset alfa if stuck
	   end;		
	   nweight = weight/sum(weight);

	    psi =  [1;features(:,iter)]; y = labels(iter);
	    sigmoidy = 1./(1 + exp(-psi'*th));
	    wgrad =  (nsamples*psi .* (y - sigmoidy) - sign(th))*nweight';

	    k = (sweep-1) * T +iter;  k,

% logistic regression step for IRL algorithm
	    alfa = alfa +0.5*lang_step1 * wgrad +  lang_step2  * randn(thdim,1);

	    est(:,k) = alfa ;         
      end;
end;

figure(4); histogram(est(1,:),'Normalization','probability');
title('Multi-kernel algorithm')
\end{lstlisting}

Remarks:  Out of 10 sweeps, where each sweep has 32000 iterations, only 14 resets  (line 25) were required for $L=100$ in IRL algorithm (\ref{eq:mcmcirl}).

\newpage

\section{Matlab Source Code for multi-kernel IRL (\ref{eq:irl_mdp}) to solve Constrained MDP  in Sec.\ref{sec:mdp}. (Generates Fig.\ref{fig:mdp}(b) and (c))}

\begin{lstlisting}
%  Multi-kernel IRL for MDP

T = 150000;  gridpoints = 100; 
tp(:,:,1) = [0.8 0.2; 0.3 0.7];  tp(:,:,2) = [0.6 0.4; 0.1 0.9];
statedim=2; actiondim = 2;
 lang_step1 =5e-6;     lang_step2 =sqrt(lang_step1);
  l_step = 1/2*lang_step1;

cost = [1 100; 30 2];  constraint  = [0.2 0.3; 2 1];   lambda=1e5;
inversestep = gridpoints/2;

  pol=zeros(statedim,gridpoints^2);  Penalty_Reward = zeros(gridpoints^2,1);
   pol_con=zeros(gridpoints^2,1);
  pol_eval= zeros(gridpoints,gridpoints); alfa = zeros(2,T); cond_prob =zeros(2,T);  

% Solve avg cost MDP exactly, over a grid of 100x100 possible randomized policies
for i=1:gridpoints-1,
  for j=1:gridpoints-1,

    k =gridpoints*(i-1)+j;
    
    policy = [i/gridpoints , 1 - i/gridpoints; j/gridpoints, 1-j/gridpoints];
    pol(:,k) = [policy(1,1); policy(2,1)];

   [ polval,polcon]  = mdp_barrier(cost,constraint,policy,tp);  % external function
   pol_con(k) = polcon;
    Penalty_Reward(k) = polval  - lambda * (( 1 - polcon)^2);
    pol_eval(i,j) = Penalty_Reward(k);
 end; 
end

figure(7); stem3(pol(1,:),pol(2,:),Penalty_Reward,'MarkerFaceColor','g')
xlabel('pol1'); ylabel('pol2');

% Evaluate and store finite difference gradients of MDP over a 100x100 grid
for i=2:gridpoints-1,
  for j=2:gridpoints-1,
       grad(i,j,1) = (pol_eval(i+1,j) - pol_eval(i-1,j)) * inversestep;
       grad(i,j,2) = (pol_eval(i,j+1) - pol_eval(i,j-1)) * inversestep;
    end
end
%%%%%%%%%%%%%%%%%%%%%%%%%%
% IRL algorithm
kernelstep=0.1;
 alfabar = [1;1]; Cker = 1/(2*pi*kernelstep^2);   L=50;

 for k=1:T,
    thbar =  pi/2* rand(2,L);
    d = vecnorm(alfabar-thbar);
    weight =10*exp(- d.^2/(2*kernelstep^2));
       if   (sum(weight) < 1e-6)
		     alfabar = 0.1*randn(2,1);  %reset alfa if stuck
    	   end;		
	   nweight = weight/sum(weight);
        p = (sin(thbar)).^2;
    pindex = min(max(round(gridpoints* p), [1;1]),[gridpoints-1;gridpoints-1]);
     ghatp = zeros(2,1);
    for i=1:L
       ghatp  = [grad(pindex(1,i),pindex(2,i),1); grad(pindex(1,i),pindex(2,i),2)] .* sin(thbar(:,i)).*cos(thbar(:,i)) *  nweight(i) + ghatp ;
    end;
   ghatm = 2* ghatp;  % weighted gradient
  % Passive Langevin dynamics
   alfabar = alfabar +  (l_step * ghatm) +  lang_step2  * randn(2,1);  alfabar = abs(alfabar); 
   alfa(:,k) = alfabar;
   cond_prob(:,k) = (sin(alfabar)).^2;   %policy conditional probabilities
  end;
% plot log of empirical density 
[M,N] =  hist3(cond_prob(:,T/2:end)',[100,100]);
figure(3); stem3(flip(N{1}),flip(N{2}),(log(M)),'MarkerFaceColor','g');
xlabel('pol1');ylabel('pol2');

\end{lstlisting}

\subsection*{External Function used in above program}
\begin{lstlisting}
function [avg_cost,avg_constraint] = mdp_barrier(cost,constraint,pol,tp)
% evaluate MDP policy for average cost MDP

statedim=size(cost,1);   actiondim=size(cost,2);

%tp(:,:,1) = [0.8 0.2; 0.3 0.7];  tp(:,:,2) = [0.6 0.4; 0.1 0.9];
%pol = [0.8, 0.2; 0.3 0.7];

%cost = [1 10; 3 2];

for i=1:statedim
   for a=1:actiondim
      l= a + (i-1)*actiondim;
         for j = 1:statedim
	    for abar = 1:actiondim
    	       m = abar + (j-1)*actiondim;
               tp_composite(l,m) = tp(i,j,a) * pol(j,abar);
	     end;
         end;
	 cost_vector(l) = cost(i,a);
	 constraint_vector(l) = constraint(i,a);
    end;
end;

[aa,bb] =  eig(tp_composite');
joint_prob = aa(:,1)/sum(aa(:,1));

avg_constraint = constraint_vector * joint_prob;

 avg_cost = cost_vector * joint_prob;
\end{lstlisting}

\end{document}

%% file: mycommands.tex
\newcommand{\obsnoise}{v}
\newcommand{\weight}{\gamma}

\newcommand{\step}{\varepsilon}
\newcommand{\e}{\step}
\newcommand{\stepa}{\mu}
\newcommand{\kernelstep}{\Delta}
\newcommand{\kernel}{K}
\newcommand{\kerneln}{\frac{1}{\kernelstep^\thdim}\,\kernel}
\newcommand{\hnabla} {\nabla}

\renewcommand{\th}{\theta}
\renewcommand{\eth}{\alpha}

\renewcommand{\bm}{W}

\newcommand{\Reward}{R}

\newcommand{\mreward}{\rho}
\newcommand{\reward}{r}
\newcommand{\state}{x}

\newcommand{\obs}{y}
\newcommand{\tslow}{n}
\newcommand{\dtime}{k}
\newcommand{\stoptime}{\tau}
\newcommand{\temperature}{\beta}
\newcommand{\thdim}{N}
\newcommand{\reals}{{\mathbb R}}

\newcommand{\noise}{w}

\newcommand{\beq}{\begin{equation}}
\newcommand{\eeq}{\end{equation}}

\newcommand{\pdf}{p}

\newcommand{\prob}{\mathbb{P}}
\newcommand{\E}                 {\Bbb{E}}

\newcommand{\horizon}{T}
\newcommand{\belief}{\pi}
\newcommand{\stat}{\pdf}
\newcommand{\argmax}{\operatornamewithlimits{arg\,max}}
\newcommand{\argmin}{\operatornamewithlimits{arg\,min}}

\newcommand{\policy}{\mathbf{u}}
\newcommand{\normal}{\mathbf{N}}
\newcommand{\p}{\prime}
\newcommand{\diag}{\operatorname{diag}}
\newcommand{\thtrue}{{\th^o}}
\newcommand{\forward}{\mathcal L^*}
\newcommand{\kalmancov}{\Sigma}
\newcommand{\Tr}{\operatorname{Tr}}

\renewcommand{\div}{\operatorname{div}}
\newcommand{\statem}{f}
\newcommand{\snoisecov}{\sigma}
\renewcommand{\skew}{S}

\newcommand{\al}{\eth}
\newcommand{\lf}{\lfloor}
\newcommand{\rf}{\rfloor}
\newcommand{\wdt}{\widetilde}
\newcommand{\wdh}{\widehat}

\def\mi{\imath}
\def\mm{{\mu,M}}
\def\suml{\sum^{(t+s)/ \delta_\mu}_{l= t/\delta_\mu}}
\def\sumk{\sum^{lm_\mu + m_\mu-1}_{k=lm_\mu}}
\def\rk{{g (\al^\mm(t_\mi): \mi\le \kappa_1)}}
\def\rkm{{g (\al^M(t_\mi): \mi\le \kappa_1)}}
\def\Em{\E_{lm_\mu}}
\def\qmk{{q^M_k}}
\def\qM{{q^M_{lm_\mu}}}
\def\qMl{{q^M_l}}

\newcommand{\beql}[1]{\begin{equation} \label{#1}}

  \newcommand{\bed}{\begin{displaymath}}
\newcommand{\eed}{\end{displaymath}}
  \newcommand{\bea}{\bed\begin{array}{rl}}
\newcommand{\eea}{\end{array}\eed}
\newcommand{\ad}{&\!\!\!\disp}
\newcommand{\aad}{&\disp}
\def\disp{\displaystyle}
\newcommand{\barray}{\begin{array}{ll}}
                       \newcommand{\earray}{\end{array}}

                     \def\({\Big(}
                     \def\){\Big)}
                     \newcommand{\ph}{\varphi}
\newcommand{\cd}{(\cdot)}
\def\op{{\cal L}}
\newcommand{\george}{\footnote}
\def\george#1 {\fbox {\footnote {\ }}\ \footnotetext { From George: #1}}
\def\vikram#1 {\fbox {\footnote {\ }}\ \footnotetext { From Vikram: #1}}

\newcommand{\statespace}{\mathcal{X}}
\newcommand{\statedim}{X}

\newcommand{\action}{u}

\newcommand{\actionspace}{\mathcal{U}}
\newcommand{\actiondim}{U}

\newcommand{\finaltime}{\horizon}

\def \tp {{P}}
\def \defn {\stackrel{\triangle}{=}}

\def \tp {{P}}
\def \admissible{{\cal D}}

\def\esp{{\mathbb E}}

\def \con {{\beta}}
\newcommand{\Cons}{B}
\def \rcon {{\gamma}}
\newcommand{\optpolicy}{\policy^*}
\newcommand{\statpi}{\bar{\phi}}

\newcommand{\randmix}{p}

\newcommand{\sgn}{\operatorname{sgn}}
\newcommand{\numparticles}{L}

\definecolor{mygreen}{RGB}{28,172,0} 
\definecolor{mylilas}{RGB}{170,55,241}

\newcommand{\param}{\th}
\newcommand{\logistic}{{\mathcal E}}
\newcommand{\cond}{\phi}

\newcommand{\mc}{x}
\newcommand{\mcstep}{\eta}
\newcommand{\stepmc}{\mcstep}
\newcommand{\normald}{\mathbf{N}}

\newtheorem{theorem}            {Theorem}

\newtheorem{proposition}        [theorem]{Proposition}